\documentclass[manuscript,screen]{acmart}  
\AtBeginDocument{%
  }

\setcopyright{acmlicensed}
\copyrightyear{2018}
\acmYear{2018}
\acmDOI{XXXXXXX.XXXXXXX}
\acmConference[Conference acronym 'XX]{Make sure to enter the correct
  conference title from your rights confirmation email}{June 03--05,
  2018}{Woodstock, NY}
\acmISBN{978-1-4503-XXXX-X/2018/06}

\usepackage{algorithmic}
\usepackage{algorithm}
\usepackage{textcomp}
\usepackage{stfloats}
\usepackage{url}
\usepackage{verbatim}
\usepackage{bm}

\usepackage{color}
\usepackage{xcolor}
\usepackage{colortbl}
\usepackage{makecell}
\usepackage{soul}
\usepackage{url}
\usepackage{amsthm}
\usepackage{microtype}
\usepackage{subfigure}
\usepackage{subcaption}
\usepackage{xspace}
\usepackage{pifont}
\usepackage{amsfonts}
\usepackage{multirow}
\usepackage{tikz}
\usepackage[edges]{forest}
\definecolor{hidden-draw}{RGB}{205, 44, 36}
\definecolor{hidden-blue}{RGB}{194,232,247}
\definecolor{hidden-orange}{RGB}{243,202,120}
\definecolor{hidden-yellow}{RGB}{242,244,193}
\definecolor{tree-level-1}{RGB}{245,20,85}
\definecolor{tree-level-2}{RGB}{246,86,118}
\definecolor{tree-level-3}{RGB}{248,177,193}
\definecolor{tree-leaf}{RGB}{176,230,198}
\usepackage{lscape}
\usetikzlibrary{positioning,shapes,shadows,arrows}

\newcolumntype{L}[1]{>{\raggedright\let\newline\\\arraybackslash\hspace{0pt}}m{#1}}
\newcolumntype{C}[1]{>{\centering\let\newline\\\arraybackslash\hspace{0pt}}m{#1}}
\newcolumntype{R}[1]{>{\raggedleft\let\newline\\\arraybackslash\hspace{0pt}}m{#1}}

\newcolumntype{x}[1]{
>{\centering\hspace{0pt}}p{#1}}%

\newcommand{\ignore}[1]{}

\definecolor{Gray}{gray}{0.9}
\definecolor{LightCyan}{rgb}{0.88,1,1}

\begin{document}

\title{A Survey of Automatic Evaluation Methods on Text, Visual and Speech Generations}


\author{Tian Lan}
\authornote{Equal Contribution}
\email{lantiangmftby@gmail.com}
\orcid{1234-5678-9012}
\affiliation{%
  \institution{Beijing Institute of Technology}
  \streetaddress{5 South Zhongguancun Street, Haidian District}
  \city{Beijing}
  \state{Beijing}
  \country{China}
  \postcode{100081}
}

\author{Yang-Hao Zhou}
\authornotemark[1]
\email{zhouyh77@bit.edu.cn}
\orcid{0009-0001-9401-4432}
\affiliation{%
  \institution{Beijing Institute of Technology}
  \streetaddress{5 South Zhongguancun Street, Haidian District}
  \city{Beijing}
  \state{Beijing}
  \country{China}
  \postcode{100081}
}

\author{Zi-Ao Ma}
\authornotemark[1]
\email{maziaoylwt@gmail.com}
\orcid{0009-0004-3552-3191}
\affiliation{%
  \institution{Beijing Institute of Technology}
  \streetaddress{5 South Zhongguancun Street, Haidian District}
  \city{Beijing}
  \state{Beijing}
  \country{China}
  \postcode{100081}
}

\author{Fanshu Sun}
\authornotemark[1]
\email{sunfs@bit.edu.cn}
\orcid{0000-0001-7969-855X}
\affiliation{%
  \institution{Beijing Institute of Technology}
  \streetaddress{5 South Zhongguancun Street, Haidian District}
  \city{Beijing}
  \state{Beijing}
  \country{China}
  \postcode{100081}
}

\author{Rui-Qing Sun}
\authornotemark[1]
\email{2325557558@qq.com}
\orcid{0009-0002-5625-2955}
\affiliation{%
  \institution{Beijing Institute of Technology}
  \streetaddress{5 South Zhongguancun Street, Haidian District}
  \city{Beijing}
  \state{Beijing}
  \country{China}
  \postcode{100081}
}

\author{Junyu Luo}
\email{luojunyu@stu.pku.edu.cn}
\orcid{0009-0001-6894-1144}
\affiliation{%
  \institution{Peking University}
  \streetaddress{5 Yiheyuan Road, Haidian District}
  \city{Beijing}
  \state{Beijing}
  \country{China}
  \postcode{100871}
}

\author{Rong-Cheng Tu}
\email{rongcheng.tu@ntu.edu.sg}
\orcid{0000-0002-9567-159X}
\affiliation{%
  \institution{Nanyang Technological University}
  \streetaddress{50 Nanyang Ave}
  \city{Singapore}
  \state{Singapore}
  \country{Singapore}
  \postcode{639798}
}

\author{Heyan Huang}
\affiliation{%
 \institution{Beijing Institute of Technology}
 \streetaddress{5 South Zhongguancun Street, Haidian District}
 \city{Beijing}
 \state{Beijing}
 \country{China}}
 \email{hhy63@bit.edu.cn}

\author{Chen Xu}
\affiliation{%
  \institution{Beijing Institute of Technology}
  \streetaddress{5 South Zhongguancun Street, Haidian District}
  \city{Beijing}
  \country{China}}
\email{chenxu05037@bit.edu.cn}
\orcid{0000-0002-3495-4238}

\author{Zhijing Wu}
\affiliation{%
  \institution{Beijing Institute of Technology}
  \streetaddress{5 South Zhongguancun Street, Haidian District}
  \city{Beijing}
  \country{China}}
\email{zhijingwu@bit.edu.cn}

\author{Xian-Ling Mao}
\authornote{Xian-Ling Mao is the corresponding author of this paper.}
\affiliation{%
  \institution{Beijing Institute of Technology}
  \streetaddress{5 South Zhongguancun Street, Haidian District}
  \city{Beijing}
  \country{China}}
\email{maoxl@bit.edu.cn}

\renewcommand{\shortauthors}{Lan et al.}


\begin{abstract}

Recent advances in deep learning have significantly enhanced generative AI capabilities across text, images, and audio. However, automatically evaluating the quality of these generated outputs presents ongoing challenges. Although numerous automatic evaluation methods exist, current research lacks a systematic framework that comprehensively organizes these methods across text, visual, and audio modalities.
To address this issue, we present a comprehensive review and a unified taxonomy of automatic evaluation methods for generated content across all three modalities;
We identify five fundamental paradigms that characterize existing evaluation approaches across these domains.
Our analysis begins by examining evaluation methods for text generation, where techniques are most mature. We then extend this framework to image and audio generation, demonstrating its broad applicability. Finally, we discuss promising directions for future research in cross-modal evaluation methodologies.
\end{abstract}


\keywords{Do, Not, Us, This, Code, Put, the, Correct, Terms, for,Your, Paper}

\received{20 February 2007}
\received[revised]{12 March 2009}
\received[accepted]{5 June 2009}

\maketitle

\section{Introduction}

Deep learning technologies have advanced significantly in recent years, driving major progress in generative models across various domains and tasks~\cite{openai2024gpt4technicalreport,tu2025multimodal,liu2024sorareviewbackgroundtechnology,rombach2021highresolution,zhang2023surveyaudiodiffusionmodels,tu2025mllm}.
Large language models (LLMs) such as GPT-4 and Claude now generate remarkably human-like conversations~\cite{openai2024gpt4technicalreport}, while diffusion models \cite{ho2020denoisingdiffusionprobabilisticmodels,sun2024diffusion,rombach2021highresolution,sun2025vorta,podell2023sdxlimprovinglatentdiffusion,sun2024asymrnr} like DALL-E and Stable Diffusion~\cite{rombach2021highresolution} have transformed image and video synthesis.
This rapid development raises a critical research question: how can we achieve reliable and accurate automatic evaluation of model-generated content?

\begin{figure}[t]
    \center{\includegraphics[width=0.75\textwidth]{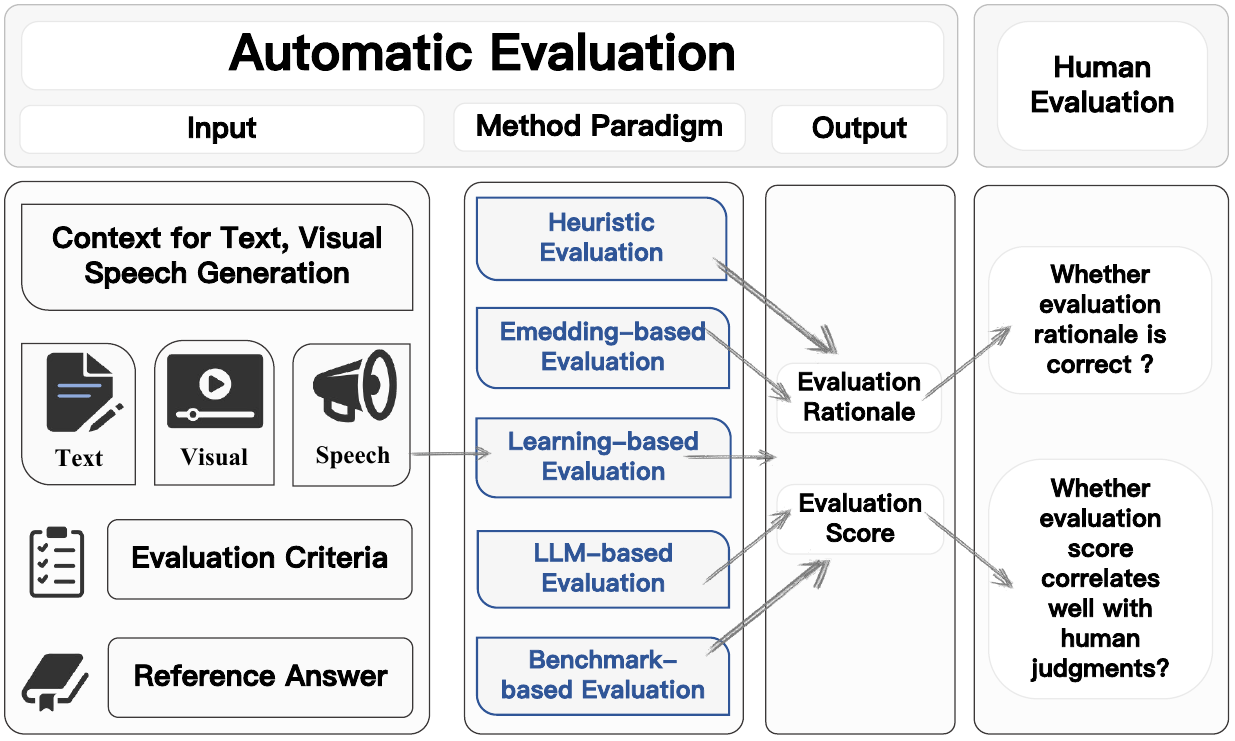}}
    \caption{Illustration of automatic evaluation for text, visual and speech generation.}\label{img:teaser}
\end{figure}

Although human evaluation remains the gold standard for assessing content quality, its high cost and inherent non-reproducibility limit its scalability for large-scale applications~\cite{lan2020ponenovelautomaticevaluation,lan2024criticevalevaluatinglargelanguage}. This has prompted researchers to develop automatic evaluation metrics that correlate strongly with human judgments~\cite{li-etal-2024-leveraging-large,li2024generationjudgmentopportunitieschallenges}.
As shown in Figure~\ref{img:teaser}, automatic evaluation aims to assess model-generated content quality based on specific evaluation criteria, reference answers, and context using appropriate evaluation methods.

Despite this progress, the field lacks a systematic survey of current developments in automatic evaluation across different tasks and modalities.
To address this gap, in this paper, we provide a comprehensive review and a unified taxonomy of automatic evaluation methods across text, visual, and speech modalities, offering insights into this evolving field.
We begin with an in-depth analysis of automatic evaluation techniques in Natural Language Generation (NLG), which has seen the most significant advances~\cite{tian-etal-2024-sheng,li-etal-2024-leveraging-large,li2024generationjudgmentopportunitieschallenges}.
Specifically, we describe and analyze the attributations of existing automatic evaluation methods, and conduct a systematic meta-evaluation of existing automatic evaluation methods.
Building on this analysis, we extend our review to other two important generative AI tasks: visual generation and audio generation. For each, we summarize the current development of automatic evaluation methods and outline promising directions for future research.

\paragraph{Difference from Related Surveys}
Existing surveys on automatic evaluation techniques primarily focus on specific methodological approaches within NLG tasks, like LLM-based evaluation methods~\cite{li-etal-2024-leveraging-large,tian-etal-2024-sheng,li2024generationjudgmentopportunitieschallenges}.
In contrast, our work provides a unified framework for automatic evaluation across three key modalities: text, visual, and speech. We cover the complete evolution of evaluation methods, from traditional heuristic approaches to modern LLM-based techniques. This cross-modal perspective offers a more comprehensive understanding of evaluation methodologies across generative AI systems.
\section{Preliminary}

In this section, we introduce the fundamental concepts of automatic evaluation methods (Section~\ref{sec:pre_concepts_for_eval}), and then describe three mainstream evaluation protocols in Section~\ref{sec:pre_eval_protocol}: single-wise, pair-wise, and corpus-wise evaluation.

\subsection{Concepts for Evaluation}
\label{sec:pre_concepts_for_eval}
As shown in Figure~\ref{img:teaser}, automatic evaluations require four key components:
(1) \textbf{Context}; (2) \textbf{Evaluation criteria}; (3) \textbf{Reference answer}; and (4) \textbf{Model-generated content} to be evaluated.

\paragraph{\textbf{Context} ($\boldsymbol{c}$)} refers to the input information that models use to generate content. Examples include conversation history in dialogue generation tasks~\cite{10.1145/3632750} and text prompts in text-to-image or text-to-video applications~\cite{tu2024automaticevaluationtexttoimagegeneration,yang2024diffusionmodelscomprehensivesurvey}.

\paragraph{\textbf{Evaluation criteria} ($\boldsymbol{cri.}$)} comprise task-specific dimensions designed for assessment, such as fluency for open-domain text generation~\cite{su2022contrastiveframeworkneuraltext,fu2023gptscoreevaluatedesire} and coherence for text summarization~\cite{fabbri2021summeval}.

\paragraph{\textbf{Reference answer} ($\boldsymbol{r}$)} is widely used for robust evaluation in typical NLG tasks such as translation and summarization~\cite{papineni-etal-2002-bleu,lin-2004-rouge}.
However, relying on a single or limited number of reference answers makes it difficult to effectively cover the vast space of possible outputs in open-ended generation tasks~\cite{lan2020ponenovelautomaticevaluation}, leading to suboptimal evaluation. Consequently, reference-free evaluation methods have gained significant attention~\cite{fu2023gptscoreevaluatedesire}, as they do not require reference answers during evaluation. In contrast, methods that depend on reference answers are termed reference-based evaluation methods.

\paragraph{\textbf{Generation} ($\boldsymbol{g}$)} refers to the model-generated content being evaluated, including text, images, videos, and audio.

\subsection{Evaluation Protocols}
\label{sec:pre_eval_protocol}
Unlike previous works~\cite{li-etal-2024-leveraging-large}, we categorize existing automatic evaluation methods into three mainstream evaluation protocols: single-wise, pair-wise, and corpus-wise evaluation. These protocols are described below.

\paragraph{Single-Wise Evaluation} directly assesses the quality of one specific generation ($g$):
\begin{equation}
    (r^*,s)=M_{\text{AE}}(c^*,cri.^*,r^*,g)
\end{equation}
where $M_{\text{AE}}$ represents any automatic evaluation method. 
$r^*$ is the textual rationale analyzing and describing the quality of $g$, typically generated by LLM-based evaluation methods~\cite{zheng2023judging} (Section~\ref{sec:llm_based_evaluation}).
$s$ is a quality score reflecting the generation quality, usually expressed as a Likert Score~\cite{zheng2023judgingllmasajudgemtbenchchatbot} within a constrained range, where higher scores indicate better quality.
The context $c$, criteria $cri.$, and reference answer $r$ are optional depending on evaluation settings and tasks, hence marked with $*$.

\paragraph{Pair-wise Evaluation} is another important and widely-used protocol~\cite{ligenerative} that determines preference between two generations ($g_A,g_B$):
\begin{equation}
    (r^*,p)=M_{\text{AE}}(c^*,cri.^*,r^*,g_A,g_B)
\end{equation}
Here, $p$ is the preference label indicating which generation is better.
Compared to single-wise evaluation, pair-wise evaluation is more robust and objective~\cite{li-etal-2024-dissecting}.

\paragraph{Corpus-level Evaluation} assesses the quality of all generations in a test set at the corpus level:
\begin{equation}
    s=M_{\text{AE}}(c^*,cri.^*,r^*,R,G)
\end{equation}
where $R=\{r_i\}_{i=1}^N$ and $G=\{g_i\}_{i=1}^N$ represent all $N$ references and model-generated samples in the corpus. Unlike single-wise and pair-wise protocols, most corpus-level evaluation methods do not generate explanatory rationales~\cite{NEURIPS2021_260c2432}.
Since corpus-level evaluation does not assess individual samples, it is considerably more coarse-grained than single-wise and pair-wise evaluation protocols.

As demonstrated in the right part of the Figure~\ref{img:teaser}, a good automatic evaluation methods should correlates well with human judgments and the generated evaluation rationale should be effective, accurate and helpful.
\section{Automatic Evaluation for Neural Text Generation (NLG)}\label{sec:nlg_evaluation}

\tikzstyle{my-box}=[
    rectangle,
    draw=hidden-draw,
    rounded corners,
    text opacity=1,
    minimum height=1.5em,
    minimum width=5em,
    inner sep=2pt,
    align=center,
    fill opacity=.5,
]
\tikzstyle{leaf}=[my-box, minimum height=1.5em,
    fill=hidden-orange!60, text=black, align=left,font=\scriptsize,
    inner xsep=2pt,
    inner ysep=4pt,
]
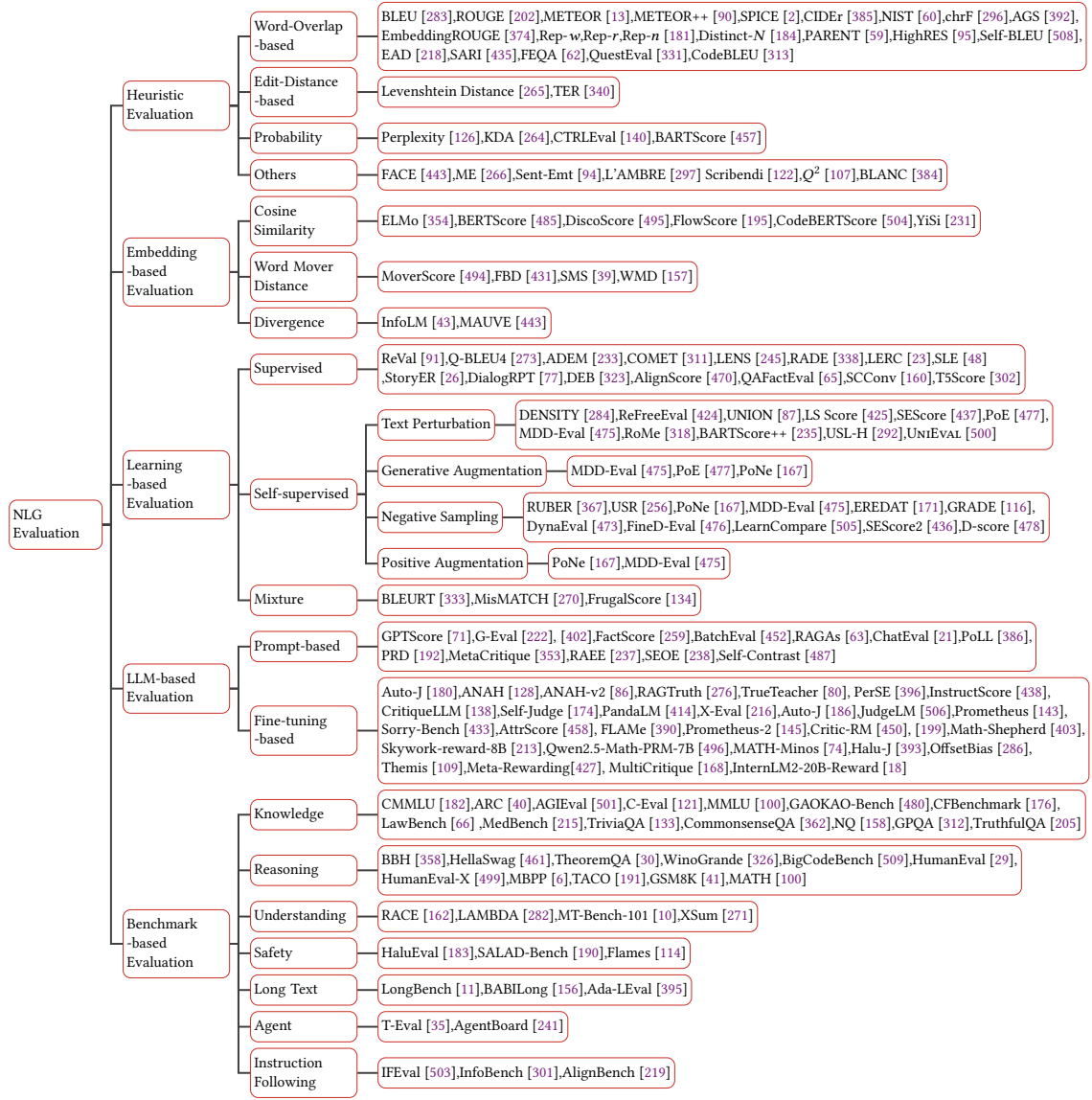
\begin{figure*}[t]
    \centering
    \resizebox{\textwidth}{!}{
        \begin{forest}
            forked edges,
            for tree={
                grow=east,
                reversed=true,
                anchor=base west,
                parent anchor=east,
                child anchor=west,
                base=left,
                font=\small,
                rectangle,
                draw=hidden-draw,
                rounded corners,
                align=left,
                minimum width=3em,
                edge+={darkgray, line width=1pt},
                s sep=3pt,
                inner xsep=2pt,
                inner ysep=3pt,
                ver/.style={rotate=90, child anchor=north, parent anchor=south, anchor=center},
            },
            where level=0{text width=5em,font=\small,}{},
            where level=1{text width=5.8em,font=\small,}{},
            where level=2{text width=5.8em,font=\small,}{},
            [NLG\\ Evaluation
                [
                    Heuristic\\ Evaluation
                    [Word-Overlap\\-based
                        [
                            BLEU~\cite{papineni-etal-2002-bleu}{,}ROUGE~\cite{lin-2004-rouge}{,}METEOR~\cite{banerjee-lavie-2005-meteor}{,}METEOR++~\cite{guo-etal-2018-meteor}{,}SPICE~\cite{anderson2016spice}{,}CIDEr~\cite{vedantam2015ciderconsensusbasedimagedescription}{,}NIST~\cite{10.5555/1289189.1289273}{,}chrF~\cite{popovic-2015-chrf}{,}AGS~\cite{wang-etal-2020-asking}{,}\\EmbeddingROUGE~\cite{10007404}{,}Rep-$w${,}Rep-$r${,}Rep-$n$~\cite{li2023repetition}{,}Distinct-$N$~\cite{li-etal-2016-diversity}{,}PARENT~\cite{dhingra-etal-2019-handling}{,}HighRES~\cite{hardy-etal-2019-highres}{,}Self-BLEU~\cite{zhu2018texygenbenchmarkingplatformtext}{,}\\EAD~\cite{liu-etal-2022-rethinking}{,}SARI~\cite{xu-etal-2016-optimizing}{,}FEQA~\cite{durmus-etal-2020-feqa}{,}QuestEval~\cite{scialom-etal-2021-questeval}{,}CodeBLEU~\cite{ren2020codebleu}
                        ]
                    ]
                    [Edit-Distance\\-based
                        [
                            Levenshtein Distance~\cite{moramarco-etal-2022-human}{,}TER~\cite{snover-etal-2006-study}
                        ]
                    ]
                    [Probability
                        [
                            Perplexity~\cite{Jelinek1977PerplexityaMO}{,}KDA~\cite{moon-etal-2022-evaluating}{,}CTRLEval~\cite{ke-etal-2022-ctrleval}{,}BARTScore~\cite{yuan2021bartscoreevaluatinggeneratedtext}
                        ]
                    ]
                    [Others
                        [
                            FACE~\cite{NEURIPS2023_37094fdc}{,}ME~\cite{mordido-meinel-2020-mark}{,}Sent-Emt~\cite{han-etal-2022-measuring}{,}L'AMBRE~\cite{pratapa-etal-2021-evaluating}
                            Scribendi~\cite{islam-magnani-2021-end}{,}$Q^2$~\cite{honovich-etal-2021-q2}{,}BLANC~\cite{vasilyev-etal-2020-fill} 
                        ]
                    ]
                ]
                [Embedding\\-based\\ Evaluation
                    [Cosine\\Similarity
                        [
                            ELMo~\cite{sun-nenkova-2019-feasibility}{,}BERTScore~\cite{zhang2020bertscoreevaluatingtextgeneration}{,}DiscoScore~\cite{zhao-etal-2023-discoscore}{,}FlowScore~\cite{li-etal-2021-conversations}{,}CodeBERTScore~\cite{zhou-etal-2023-codebertscore}{,}YiSi~\cite{lo-2019-yisi}
                        ]
                    ]
                    [Word Mover\\ Distance
                        [
                            MoverScore~\cite{zhao-etal-2019-moverscore}{,}FBD~\cite{xiang-etal-2021-assessing}{,}SMS~\cite{clark-etal-2019-sentence}{,}WMD~\cite{pmlr-v37-kusnerb15}
                        ]
                    ]
                    [Divergence
                        [
                            InfoLM~\cite{Colombo2021InfoLMAN}{,}MAUVE~\cite{NEURIPS2023_37094fdc}
                        ]
                    ]
                ]
                [Learning\\-based\\ Evaluation
                    [Supervised\\
                        [
                            ReVal~\cite{gupta-etal-2015-reval}{,}Q-BLEU4~\cite{nema-khapra-2018-towards}{,}ADEM~\cite{lowe-etal-2017-towards}{,}COMET~\cite{rei-etal-2020-comet}{,}LENS~\cite{maddela-etal-2023-lens}{,}RADE~\cite{shi-etal-2023-rade}{,}LERC~\cite{chen-etal-2020-mocha}{,}SLE~\cite{cripwell-etal-2023-simplicity}\\{,}StoryER~\cite{chen-etal-2022-storyer}{,}DialogRPT~\cite{gao2020dialogrpt}{,}DEB~\cite{sai2020improvingdialogevaluationmultireference}{,}AlignScore~\cite{zha-etal-2023-alignscore}{,}QAFactEval~\cite{fabbri2022qafacteval}{,}SCConv~\cite{Laban2022SummaCRN}{,}T5Score~\cite{qin2022t5scorediscriminativefinetuninggenerative}
                        ]
                    ]
                    [
                        Self-supervised
                        [
                            Text Perturbation
                            [
                                DENSITY~\cite{park-etal-2023-density}{,}ReFreeEval~\cite{wu-etal-2023-holistic}{,}UNION~\cite{guan-huang-2020-union}{,}LS Score~\cite{wu-etal-2020-unsupervised}{,}SEScore~\cite{xu-etal-2022-errors}{,}PoE~\cite{10056996}{,}\\MDD-Eval~\cite{zhang2022mddevalselftrainingaugmenteddata}{,}RoMe~\cite{rony-etal-2022-rome}{,}BARTScore++~\cite{lu2022humanlikeevaluationnaturallanguage}{,}USL-H~\cite{phy-etal-2020-deconstruct}{,}\textsc{UniEval}~\cite{zhong-etal-2022-towards}
                            ]
                        ] 
                        [
                            Generative Augmentation
                            [
                                MDD-Eval~\cite{zhang2022mddevalselftrainingaugmenteddata}{,}PoE~\cite{10056996}{,}PoNe~\cite{lan2020ponenovelautomaticevaluation}
                            ]
                        ]
                        [
                            Negative Sampling
                            [
                                RUBER~\cite{tao2017ruberunsupervisedmethodautomatic}{,}USR~\cite{mehri-eskenazi-2020-usr}{,}PoNe~\cite{lan2020ponenovelautomaticevaluation}{,}MDD-Eval~\cite{zhang2022mddevalselftrainingaugmenteddata}{,}\textsc{EREDAT}~\cite{le-scao-gardent-2023-joint}{,}GRADE~\cite{huang-etal-2020-grade}{,}\\DynaEval~\cite{zhang-etal-2021-dynaeval}{,}FineD-Eval~\cite{zhang-etal-2022-fined}{,}LearnCompare~\cite{zhou2020learningcomparebettertraining}{,}SEScore2~\cite{xu-etal-2023-sescore2}{,}D-score~\cite{10.1109/TASLP.2021.3074012}
                            ]
                        ]
                        [
                            Positive Augmentation
                            [
                                PoNe~\cite{lan2020ponenovelautomaticevaluation}{,}MDD-Eval~\cite{zhang2022mddevalselftrainingaugmenteddata}
                            ]
                        ]
                    ]
                    [
                        Mixture
                        [
                            BLEURT~\cite{sellam-etal-2020-bleurt}{,}MisMATCH~\cite{murugesan-etal-2023-mismatch}{,}FrugalScore~\cite{kamal-eddine-etal-2022-frugalscore}
                        ]
                    ]
                ]
                [
                    LLM-based\\ Evaluation
                    [
                        Prompt-based
                        [
                            GPTScore~\cite{fu2023gptscoreevaluatedesire}{,}G-Eval~\cite{liu2023gevalnlgevaluationusing}{,}~\cite{wang2023chatgptgoodnlgevaluator}{,}FactScore~\cite{min2023factscore}{,}BatchEval~\cite{yuan-etal-2024-batcheval}{,}RAGAs~\cite{es-etal-2024-ragas}{,}ChatEval~\cite{chan2023chateval}{,}PoLL~\cite{verga2024replacingjudgesjuriesevaluating}{,}\\PRD~\cite{li2024prdpeerrankdiscussion}{,}MetaCritique~\cite{sun2024critiquecritique}{,}RAEE~\cite{lu2024exactmatchsemanticallyreassessing}{,}SEOE~\cite{lu2025seoescalablereliablesemantic}{,}Self-Contrast~\cite{zhang2024selfcontrastbetterreflectioninconsistent}
                        ]
                    ]
                    [
                        Fine-tuning\\-based
                        [
                            Auto-J~\cite{li2024generationjudgmentopportunitieschallenges}{,}ANAH~\cite{ji-etal-2024-anah}{,}ANAH-v2~\cite{gu2024anah}{,}RAGTruth~\cite{niu-etal-2024-ragtruth}{,}TrueTeacher~\cite{gekhman2023trueteacher}{,}
                            PerSE~\cite{wang-etal-2024-learning-personalized}{,}InstructScore~\cite{xu-etal-2023-instructscore}{,}\\
                            CritiqueLLM~\cite{ke2024critiquellm}{,}Self-Judge~\cite{lee-etal-2024-aligning}{,}PandaLM~\cite{wangpandalm}{,}X-Eval~\cite{liu2024x}{,}Auto-J~\cite{ligenerative}{,}JudgeLM~\cite{zhu2023judgelm}{,}Prometheus~\cite{kim2023prometheus}{,}\\
                            Sorry-Bench~\cite{xie2024sorry}{,}AttrScore~\cite{yue2023automatic}{,} FLAMe~\cite{vu2024foundational}{,}Prometheus-2~\cite{kim-etal-2024-prometheus}{,}Critic-RM~\cite{yu2024self}{,}~\cite{lightman2023let}{,}Math-Shepherd~\cite{wang-etal-2024-math}{,}\\
                            Skywork-reward-8B~\cite{liu2024skywork}{,}Qwen2.5-Math-PRM-7B~\cite{zheng2024processbench}{,}MATH-Minos~\cite{gao2024llm}{,}Halu-J~\cite{wang2024halujcritiquebasedhallucinationjudge}{,}OffsetBias~\cite{park-etal-2024-offsetbias}{,}\\
                            Themis~\cite{hu-etal-2024-themis}{,}Meta-Rewarding\cite{wu2024meta}{,} MultiCritique~\cite{lan2024training}{,}InternLM2-20B-Reward~\cite{cai2024internlm2technicalreport}
                        ]
                    ]
                ]
                [
                    Benchmark\\-based\\ Evaluation \\
                    [Knowledge\\
                        [
                            CMMLU~\cite{li2023cmmlu}{,}ARC~\cite{clark2018think}{,}AGIEval~\cite{zhong2023agieval}{,}C-Eval~\cite{huang2023ceval}{,}MMLU~\cite{hendrycksmath2021}{,}GAOKAO-Bench~\cite{zhang-etal-2023-sac3}{,}CFBenchmark~\cite{lei2024cfbenchmarkchinesefinancialassistant}{,}\\LawBench~\cite{fei2023lawbenchbenchmarkinglegalknowledge}
                            {,}MedBench~\cite{liu2024medbenchcomprehensivestandardizedreliable}{,}TriviaQA~\cite{JoshiTriviaQA2017}{,}CommonsenseQA~\cite{talmor2018commonsenseqa}{,}NQ~\cite{kwiatkowski2019natural}{,}GPQA~\cite{rein2024gpqa}{,}TruthfulQA~\cite{lin2022truthfulqa}
                        ]
                    ]
                    [
                        Reasoning\\
                        [
                            BBH~\cite{suzgun2022challenging}{,}HellaSwag~\cite{zellers2019hellaswag}{,}TheoremQA~\cite{chen2023theoremqa}{,}WinoGrande~\cite{sakaguchi2019winogrande}{,}BigCodeBench~\cite{zhuo2024bigcodebench}{,}HumanEval~\cite{chen-etal-2021-training}{,}\\HumanEval-X~\cite{zheng2023codegeex}{,}MBPP~\cite{austin2021program}{,}TACO~\cite{li2023taco}{,}GSM8K~\cite{cobbe2021gsm8k}{,}MATH~\cite{hendrycksmath2021}
                        ]
                    ]
                    [
                        Understanding\\
                        [
                            RACE~\cite{lai2017race}{,}LAMBDA~\cite{paperno2016lambada}{,}MT-Bench-101~\cite{bai2024mt}{,}XSum~\cite{narayan2018dont}
                        ]
                    ]
                    [
                        Safety\\
                        [
                            HaluEval~\cite{li2023halueval}{,}SALAD-Bench~\cite{li-etal-2024-salad}{,}Flames~\cite{huang2023flames}
                        ]
                    ]
                    [
                        Long Text\\
                        [
                            LongBench~\cite{bai2023longbench}{,}BABILong~\cite{kuratov2024babilong}{,}Ada-LEval~\cite{wang2024adaleval}
                        ]
                    ]
                    [
                        Agent\\
                        [
                            T-Eval~\cite{chen2023t}{,}AgentBoard~\cite{ma2024agentboard}
                        ]
                    ]
                    [
                        Instruction\\Following\\
                        [
                            IFEval~\cite{zhou2023instruction}{,}InfoBench~\cite{qin2024infobench}{,}AlignBench~\cite{liu2024alignbenchbenchmarkingchinesealignment}
                        ]
                    ]
                ]
        ]
        \end{forest}
    }
    \caption{Taxonomy of representative automatic evaluation methods in NLG. The detailed classifications about LLM-based evaluation methods could be found in Section~\ref{sec:nlg_evaluation}.}
    \label{img:nlg_taxonomy}
\end{figure*}

In this section, we comprehensively review the progress of automatic evaluation techniques for Neural Language Generation (NLG) tasks, including open-ended text generation, summarization, translation, etc.
Figure 2 presents a comprehensive taxonomy of automatic evaluation methods for NLG tasks, which can be broadly classified into five main categories:
(1) \textbf{\textit{Heuristic evaluation}} uses rule-based and heuristic features in natural language for assessment, such as word-overlap metrics or generation probabilities;
(2) \textbf{\textit{Embedding-based evaluation}} measures semantic similarity between references and generated texts~\cite{zhang2020bertscoreevaluatingtextgeneration};
(3) \textbf{\textit{Learning-based evaluation}} trains neural networks on human-annotated data to assess text quality;
(4) \textbf{\textit{LLM-based evaluation}} employs large language models to perform evaluations through carefully designed prompts and chain-of-thought reasoning, also known as LLM-as-a-judge~\cite{li2024generationjudgmentopportunitieschallenges};
and (5) \textbf{\textit{Benchmark-based evaluation}} tests specific capabilities using human-annotated benchmarks, such as mathematical reasoning and code generation.
The following subsections provide detailed explorations of each category with their representative works.
Finally, in Section~\ref{sec:comparing_automatic_evaluation}, we systematically compare and analyze the characteristics of these automated evaluation paradigms.

\subsection{Heuristic Evaluation}

Heuristic evaluation relies on rule-based or heuristic features for evaluation. Existing heuristic methods could be classified into three categories: (1) \textbf{Word-Overlap}; (2) \textbf{Edit-Distance}; and (3) \textbf{Generation Probability}. Table~\ref{tab:heuristic_based_evaluation} provides a complete overview of these methods.

\subsubsection{Word-overlap}
Word-overlap-based heuristic metrics evaluate text generation through three main criteria:

\paragraph{(1) Text Similarity:} These metrics assume that high-quality generated texts should closely match ground-truth texts~\cite{ren2020codebleu,hardy-etal-2019-highres,dhingra-etal-2019-handling}. Common examples include BLEU~\cite{papineni-etal-2002-bleu}, ROUGE~\cite{lin-2004-rouge}, and METEOR~\cite{banerjee-lavie-2005-meteor}, which measure machine translation and summarization quality using $n$-gram overlap (precision and recall). Additionally, chrF~\cite{popovic-2015-chrf} and METEOR~\cite{banerjee-lavie-2005-meteor} combine precision and recall in $n$-gram matching to provide more accurate similarity assessments. NIST~\cite{10.5555/1289189.1289273} and CIDEr~\cite{vedantam2015ciderconsensusbasedimagedescription} assign weights to different $n$-grams, focusing on more important words and phrases.

\paragraph{(2) Text Diversity:} Diversity is crucial for open-domain text generation. Metrics such as Rep-$n$~\cite{su2023contrastive} and Distinct-$N$~\cite{li-etal-2016-diversity} measure generation quality by calculating the proportion of unique $n$-grams in the output. Self-BLEU~\cite{zhu2018texygenbenchmarkingplatformtext} calculates the BLEU score for each generated sentence and uses the average score to measure diversity. To address Distinct-$N$'s bias toward longer texts, Expectation-Adjusted Distinct (EAD)~\cite{liu-etal-2022-rethinking} adjusts the number of distinct tokens based on statistical expectations.

\paragraph{(3) Factual Consistency:} Factual consistency or faithfulness is essential for text summarization. Most automatic evaluation methods use question answering (QA) models~\cite{rajpurkar-etal-2018-know} to generate answers from both the summary and source document. Word-overlap metrics then measure the similarity between these answers~\cite{durmus-etal-2020-feqa,scialom-etal-2021-questeval,wang-etal-2020-asking}.

Most word-overlap-based metrics require references (reference-based), as they measure differences between generated text and reference text. 
Although HighRES~\cite{hardy-etal-2019-highres} is reference-free, it still needs human-annotated source sentences from documents to serve as reference proxies for evaluating generated summaries.

\subsubsection{Edit-Distance} 
Unlike word-overlap-based methods that compute $n$-gram overlap, edit-distance-based methods evaluate text quality by counting the number of character or word-level transformations needed to convert generated text into reference text~\cite{moramarco-etal-2022-human}.
TER~\cite{snover-etal-2006-study} and WER~\cite{negri-etal-2014-quality} are two representative methods that evaluate the quality of model-generated summaries and translations by computing edit distance from ground-truth.

\subsubsection{Probability-based}
While word-overlap and edit-distance methods use $n$-gram lexical features, many approaches leverage probability features for evaluation~\cite{moon-etal-2022-evaluating,ke-etal-2022-ctrleval,su2023contrastive}, such as perplexity (PPL) \cite{Jelinek1977PerplexityaMO}. These methods assume that generations with high probability have better quality.
Pre-trained Language Models (PLMs) such as BERT~\cite{devlin-etal-2019-bert} and BART~\cite{lewis-etal-2020-bart} are commonly used to compute generation probability.
For example, BARTScore~\cite{yuan2021bartscoreevaluatinggeneratedtext} measures generation probability using pre-trained language models given the input or reference.
Additionally, CTRLEval \cite{ke-etal-2022-ctrleval} evaluates text generation across multiple dimensions—coherence, consistency, and attribute relevance—based on context (prefix and attribute label) by computing generation probability in designed text infilling tasks.
An increasing number of studies now use large language models (LLMs) to evaluate generated text probability~\cite{li-etal-2024-leveraging-large}, such as GPTScore~\cite{fu2023gptscoreevaluatedesire}. 
Though these approaches share similarities with other probability-based methods, we discuss them separately under the LLM-based evaluation paradigm (Section~\ref{sec:llm_based_evaluation}) due to their specific use of LLMs.

\subsubsection{Other Heuristic Attributions}
Beyond word-overlap, edit-distance, and generation probability, several approaches leverage alternative heuristic features for evaluation. For example, FACE~\cite{NEURIPS2023_37094fdc} measures text similarity based on Fourier Analysis of estimated Cross-Entropy of text.
Some methods focus on corpus-level evaluation. Mark-Evaluate~\cite{mordido-meinel-2020-mark} introduces evaluation metrics inspired by population size estimators widely used in ecology. Zipf~\cite{holtzman2020curiouscaseneuraltext} measures the exponential relationship between word rank and frequency in text, analyzing how well generated text adheres to natural language distribution patterns.
For faithfulness evaluation, Q$^2$~\cite{honovich-etal-2021-q2} uses Natural Language Inference (NLI) models to compare answers generated with knowledge sources to answers generated from dialogue responses, quantifying information consistency between sources and generations.

\begin{table}[htbp]
    \centering
    \resizebox{0.9\textwidth}{!}{
    \begin{tabular}{cccccc}
        \toprule
         \textbf{Heuristic Methods} &\textbf{NLG Task}&\textbf{Category} & \textbf{\begin{tabular}[c]{@{}c@{}}Need\\Reference\end{tabular}} & \textbf{\begin{tabular}[c]{@{}c@{}}Need\\Context\end{tabular}} & \textbf{\begin{tabular}[c]{@{}c@{}}Evaluation\\Protocols\end{tabular}}\\
         \midrule
         BLEU~\cite{papineni-etal-2002-bleu}&Machine Translation&Word-overlap&Yes&No&Single\\
         ROUGE~\cite{lin-2004-rouge}&Text Summarization&Word-overlap&Yes&No&Single\\
         METEOR~\cite{banerjee-lavie-2005-meteor}&Machine Translation&Word-overlap&Yes&No&Single\\
         METEOR++~\cite{guo-etal-2018-meteor}&Machine Translation&Word-overlap&Yes&No&Single\\
         SPICE~\cite{10.1007/978-3-319-46454-1_24}&Image Caption&Word-overlap&Yes&No&Single\\
         CIDEr~\cite{vedantam2015ciderconsensusbasedimagedescription}&Image Caption&Word-overlap&Yes&No&Single\\
         NIST~\cite{10.5555/1289189.1289273}&Machine Translation&Word-overlap&Yes&No&Single\\
         chrF~\cite{popovic-2015-chrf}&Machine Translation&Word-overlap&Yes&No&Single\\
         EmbeddingROUGE~\cite{10007404}&Text Summarization&Word-overlap&Yes&No&Single\\
         MEANT2.0~\cite{lo-2017-meant}&Machine Translation&Word-overlap&Yes&No&Single\\ 
         CodeBLEU~\cite{ren2020codebleu}&Code Generation&Word-overlap&Yes&No&Single\\
         PARENT~\cite{dhingra-etal-2019-handling}&Data-to-Text&Word-overlap&Yes&No&Single\\
         HighRES~\cite{hardy-etal-2019-highres}&Text Summarization&Word-overlap&No&Yes&Single\\
         EAD~\cite{liu-etal-2022-rethinking}&General Text Generation&Word-overlap&Yes&No&Single\\
         Rep-n~\cite{su2023contrastive}&General Text Generation&Word-overlap&Yes&No&Single\\
         Rep-w~\cite{li2023repetition}&General Text Generation&Word-overlap&Yes&No&Single\\
         Rep-r~\cite{li2023repetition}&General Text Generation&Word-overlap&Yes&No&Single\\
         Distinct-$N$~\cite{li-etal-2016-diversity}&General Text Generation&Word-overlap&Yes&No&Single\\
         Self-BLEU~\cite{zhu2018texygenbenchmarkingplatformtext}&General Text Generation&Word-overlap&Yes&No&Single\\
         QAGS~\cite{wang-etal-2020-asking}&Text Summarization&Word-overlap&No&Yes&Single\\
         FEQA~\cite{durmus-etal-2020-feqa}&Abstractive Summarization&Word-overlap&Yes&Yes&Single\\
         QuestEval~\cite{scialom-etal-2021-questeval}&Text Summarization&Word-overlap&No&Yes&Single\\
         SARI~\cite{xu-etal-2016-optimizing}&Text Simplification&Word-overlap&Yes&Yes&Single\\
         TER~\cite{snover-etal-2006-study}&Machine Translation&Edit-Distance&Yes&No&Single\\
         Levenshtein~\cite{moramarco-etal-2022-human}&Consultation Note Generation&Edit-Distance&Yes&No&Single\\
         BARTScore~\cite{yuan2021bartscoreevaluatinggeneratedtext}&General Text Generation&Probability&Yes&No&Single\\
         Perpleity~\cite{Jelinek1977PerplexityaMO}&General Text Generation&Probability&Yes&No&Single\\
         KDA~\cite{moon-etal-2022-evaluating}&Multiple Choice Questions &Probability&Yes&No&Single\\
         CRTLEval~\cite{ke-etal-2022-ctrleval}&Controlled Text Generation&Probability&No&Yes&Single\\
         FACE~\cite{NEURIPS2023_37094fdc}&General Text Generation&Other&Yes&No&Single\\
         ME~\cite{mordido-meinel-2020-mark}&General Text Generation&Other&Yes&No&Corpus\\
         Zipf~\cite{holtzman2020curiouscaseneuraltext}&General Text Generation&Other&No&No&Corpus\\
         Sent-Emt~\cite{han-etal-2022-measuring}&Dialogue Generation&Other&No&Yes&Corpus\\
         L'AMBRE~\cite{pratapa-etal-2021-evaluating}&Machine Translation&Other&No&No&Single\\
         Scibendi~\cite{islam-magnani-2021-end}&Grammatical Error Correction&Other&No&Yes&Single\\
         Q$^2$~\cite{honovich-etal-2021-q2}&Knowledge-Grounded Dialogue&Other&No&Yes&Single\\
         BLANC~\cite{vasilyev-etal-2020-fill}&Text Summarization&Other&No&Yes&Single\\
         \bottomrule
    \end{tabular}
    }
    \caption{The representative evaluation methods for heuristic methods. Based on whether references are required, the methods can be further divided into reference-based and reference-free. The same principle applies to the context dimension (context-free and context-based).}
    \label{tab:heuristic_based_evaluation}
\end{table}

\subsection{Embedding-based Evaluation}
Heuristic evaluation methods often overlook semantic features in text generation, which leads to significant discrepancies between automated and human evaluation results.
With the rapid development of distributed word embeddings and representation learning techniques~\cite{10007404,pennington-etal-2014-glove}, \textbf{\textit{Embedding-based Evaluation}} has emerged as an alternative approach that measures the semantic similarity between generated text~\cite{zhang2020bertscoreevaluatingtextgeneration} and the context or reference.
Based on the methods used to measure embedding similarity, existing embedding-based evaluation approaches can be categorized into three main types: (1) Cosine Similarity; (2) Embedding Distance; and (3) Divergence.
Table~\ref{tab:embeding_based_metrics} summarizes the current embedding-based evaluation metrics, which we discuss in detail below.

\begin{table*}[htbp]
    \centering
    \resizebox{0.8\textwidth}{!}{
    \begin{tabular}{cccccc}
        \toprule
         \textbf{\begin{tabular}[c]{@{}c@{}}Embedding-based\\Methods\end{tabular}} &\textbf{NLG Task}&\textbf{Category} & \textbf{\begin{tabular}[c]{@{}c@{}}Need\\Reference\end{tabular}} & \textbf{\begin{tabular}[c]{@{}c@{}}Need\\Context\end{tabular}} & \textbf{\begin{tabular}[c]{@{}c@{}}Evaluation\\Protocols\end{tabular}}\\        \midrule
         ELMO~\cite{sun-nenkova-2019-feasibility}&Text Summarization&Cosine Similarity&Yes&No&Single\\
         FlowScore~\cite{li-etal-2021-conversations}&Dialogue Generation&Cosine Similarity&No&Yes&Single\\
         DiscoScore~\cite{zhao-etal-2023-discoscore}&General Text Generation&Cosine Similarity&Yes&No&Single\\
         YiSi~\cite{lo-2019-yisi}&Machine Translation&Cosine Similarity&Yes&No&Single\\
         CodeBERTScore~\cite{zhou-etal-2023-codebertscore}&Code Generation&Cosine Similarity&Yes&Yes&Single\\
         BERTScore~\cite{zhang2020bertscoreevaluatingtextgeneration}&General Text Generation&Cosine Similarity&Yes&No&Single\\
         FBD~\cite{xiang-etal-2021-assessing}&Dialogue Generation&WMD&Yes&No&Corpus\\
         SMS~\cite{clark-etal-2019-sentence}&General Text Generation&WMD&Yes&No&Single\\
         WMD~\cite{pmlr-v37-kusnerb15}&General Text Generation&WMD&Yes&No&Single\\
         MoverScore~\cite{zhao-etal-2019-moverscore}&General Text Generation&WMD&Yes&No&Single\\
         MAUVE~\cite{NEURIPS2021_260c2432}&General Text Generation&Divergence&Yes&No&Corpus\\
         InforLM~\cite{colombo-etal-2021-automatic}&General Text Generation&Divergence&Yes&No&Single\\
         \bottomrule
    \end{tabular}
    }
    \caption{The complete list of the embedding-based evaluation methods.}
    \label{tab:embeding_based_metrics}
\end{table*}

\subsubsection{Cosine similarity}
Cosine similarity is the most common approach for measuring semantic similarity between two sentence embeddings~\cite{zhao-etal-2023-discoscore,lo-2019-yisi,mathur-etal-2019-putting,zhan-etal-2021-difficulty,gao-etal-2020-supert,chen-guo-2015-representation}.
Before the advent of pre-trained language models like BERT~\cite{devlin-etal-2019-bert}, researchers typically used pre-trained word embeddings such as Word2Vec~\cite{mikolov2013efficientestimationwordrepresentations} and GloVe~\cite{pennington-etal-2014-glove} to construct semantic vectors for text~\cite{ng-abrecht-2015-better,sun-nenkova-2019-feasibility}.
For example, Vector Extrema, Embedding Average, and Greedy Matching \cite{liu-etal-2016-evaluate} apply cosine similarity to evaluate dialog generation quality based on the pre-trained word embeddings.
With the emergence of pre-trained language models (PLMs) such as BERT~\cite{devlin-etal-2019-bert} and RoBERTa~\cite{liu2020roberta}, semantic embeddings from these models have demonstrated superior ability to capture semantic information, leading to more robust evaluation methods~\cite{chen-etal-2021-training,zhao-etal-2020-limitations,li-etal-2021-conversations,zhu-etal-2023-imagine,zhou-etal-2023-codebertscore}. A notable example is BERTScore~\cite{Zhang2020BERTScore:}, a representative embedding-based metric that extracts semantic vectors using BERT and computes the IDF-weighted maximum cosine similarity between each token pair from reference and generated text.

\subsubsection{Word Mover's Distance (WMD)}
WMD calculates the minimum transport cost (distance) between two sets of word embeddings\cite{johnson-2022-binary}, with Euclidean distance and $\mathcal{L}_p$ distance (where $p\in {1,2,+\infty}$) being the most frequently used measures~\cite{colombo-etal-2021-automatic}.
MoverScore \cite{zhao-etal-2019-moverscore}, for instance, uses Euclidean distance as the transport cost to compute WMD between generated and reference texts using BERT word embeddings.
Similarly, WMD~\cite{pmlr-v37-kusnerb15} and Sentence Mover's Similarity (SMS) \cite{clark-etal-2019-sentence} measure the cost of transforming generated text into reference text using semantic embeddings. Besides, \citet{xiang-etal-2021-assessing} introduced Frechet Bert Distance (FBD) to compute the distance between references and generated texts.

\subsubsection{Divergence}
Divergence represents another category of embedding-based evaluation metrics. InfoLM \cite{Colombo2021InfoLMAN} computes three families of divergences on discrete probability distributions over the vocabulary generated by PLMs.
MAUVE~\cite{NEURIPS2021_260c2432} measures the KL-divergence between reference and generated texts by analyzing their mixture distributions.
\subsection{Learning-based Evaluation}\label{sec:learning_based_methods}

Embedding-based evaluation methods often show limited correlation with human judgments due to two key limitations: (1) they operate at a coarse-grained level, failing to capture critical error patterns, especially in grammar and diversity; and (2) they rely on finite reference sets that inadequately represent the full spectrum of high-quality responses, potentially penalizing valid generations that differ from references.
Inspired by data-driven machine learning approaches, \textbf{\textit{Learning-based Evaluation}} uses deep neural networks to mimic human evaluation patterns~\cite{lowe-etal-2017-towards} by training on specially constructed evaluation datasets. 
The training dataset forms the core of learning-based evaluation methods. Current approaches can be classified into three categories based on their dataset construction methods:
(1) \textbf{Supervised methods} train evaluation models on human-annotated datasets;
(2) \textbf{Self-supervised methods} train evaluation models on synthetic datasets generated through heuristic strategies;
(3) \textbf{Mixture methods} employ a two-stage training approach, typically pre-training on large-scale synthetic data followed by fine-tuning on human-annotated datasets.

\begin{table*}[htbp]
    \centering
    \resizebox{0.7\textwidth}{!}{
        \begin{tabular}{ccccc}
        \toprule
        \textbf{Metric} &\textbf{NLG Task}& 
        \textbf{\begin{tabular}[c]{@{}c@{}}Need\\Reference\end{tabular}} & \textbf{\begin{tabular}[c]{@{}c@{}}Need\\Context\end{tabular}} & \textbf{\begin{tabular}[c]{@{}c@{}}Evaluation\\Protocols\end{tabular}}\\\midrule
        T5Score~\cite{qin2022t5scorediscriminativefinetuninggenerative}&Text Generation&Both&Yes&Single\\
        ReVal~\cite{gupta-etal-2015-reval}&Machine Translation&Yes&No&Single\\
         Q-BLEU4~\cite{nema-khapra-2018-towards}&Question Generation&Yes&No&Single\\
         LENS~\cite{maddela-etal-2023-lens}&Text Simplification&Yes&Yes&Single\\
         RADE~\cite{shi-etal-2023-rade}&Dialogue Generation&Yes&Yes&Single\\
         LERC~\cite{chen-etal-2020-mocha}&Read Comprehensive&Yes&Yes&Single\\
         SLE~\cite{cripwell-etal-2023-simplicity}&Text Simplification&No&Yes&Single\\
         StoryER~\cite{chen-etal-2022-storyer}&Story Generation&No&No&Single\\
         DEB~\cite{sai-etal-2020-improving}&Dialogue Generation&No&Yes&Single\\
         BEER~\cite{stanojevic-simaan-2014-beer}&Machine Translation&Yes&No&Single\\
         LEIC~\cite{cui2018learningevaluateimagecaptioning}&Image Caption&No&Yes&Single\\
         RUSE~\cite{shimanaka-etal-2018-ruse}&Machine Translation&No&Yes&Single\\
         SentBLEU~\cite{zhang-van-genabith-2020-translation}&Machine Translation&No&Yes&Single\\
         ADEM~\cite{lowe-etal-2017-towards}&Dialogue Generation&Yes&Yes&Single\\
         COMET~\cite{rei-etal-2020-comet}&Machine Translation&Yes&No&Single\\
         DialogRPT~\cite{gao-etal-2020-dialogue}&Dialogue Generation&No&Yes&Single\\
         AlignScore~\cite{zha2023alignscoreevaluatingfactualconsistency}&General Text Generation&No&Yes&Single\\
         QAFactEval~\cite{fabbri2022qafacteval}&Text Summarization&No&Yes&Single\\
         SCConv~\cite{laban-etal-2022-summac}&Text Summarization&No&Yes&Single\\
        \bottomrule
        \end{tabular}
    }
    \caption{\label{tab:supervised_learning_based}The representative learning-based (supervised learning) evaluation methods.}
\end{table*}


\subsubsection{Supervised}
Supervised learning-based automatic evaluation methods train models to mimic human assessment using high-quality human-annotated datasets~\cite{mathur-etal-2019-putting,Zhang2021,weston-etal-2022-generative,liu-etal-2023-towards-interpretable,pu-etal-2021-learning,rubino-etal-2021-error,bulian-etal-2022-tomayto,wang-etal-2022-paratag}.
Table~\ref{tab:supervised_learning_based} provides a complete list of these methods, with representative approaches described below.

\paragraph{(1) Reference-based and Context-free Methods:} Early reference-based and context-free supervised methods typically fine-tuned hyper-parameters or trained pairwise classifiers or regression models based on RNN/LSTM architectures \cite{guzman-etal-2015-pairwise,guzman-etal-2014-learning,gupta-etal-2015-reval}. For instance, Q-BLEU4 \cite{nema-khapra-2018-towards} tunes two hyper-parameters on human evaluation scores, combining answerability scores with BLEU-4 scores for automatic question generation (AQG) systems. RUSE \cite{shimanaka-etal-2018-ruse} implements a regression model trained directly on human-rated translation quality scores.

\paragraph{(2) Reference-based and Context-based Methods:}
These methods evaluate generated text by considering both references and contextual information. A notable example is COMET~\cite{rei-etal-2020-comet}, which offers two variants—COMET-MQM and COMET-DARR—trained on human-evaluated machine translation corpora that incorporate source sentences and reference translations. Similarly, \citet{takahashi-etal-2020-automatic} developed machine translation evaluation methods based on the XLM model \cite{lample2019crosslingual}. This approach extends to other natural language generation tasks as well. For text simplification, LENS \cite{maddela-etal-2023-lens} outperforms strong baselines like BERTScore by training on human-evaluated simplified texts. RADE \cite{shi-etal-2023-rade} assesses dialog generation quality using human responses as references. Additionally, LERC \cite{chen-etal-2020-mocha} trains on human judgments from its MOCHA dataset for generative reading comprehension tasks, while ADEM \cite{lowe-etal-2017-towards} optimizes RNN models to evaluate dialog utterance quality based on conversation history and reference responses.

\paragraph{(3) Reference-free and Context-based Methods:}
These methods are widely used for open-domain generation tasks, where outputs that differ from limited reference examples may still be of high quality. In such cases, only contextual information is useful for evaluating generation quality.
For example, DialogRPT \cite{gao-etal-2020-dialogue} fine-tunes the DialoGPT model \cite{zhang2020dialogptlargescalegenerativepretraining} on human-rated datasets to create a ranking-based evaluation metric. DEB \cite{sai-etal-2020-improving} builds a high-quality human-annotated dataset specifically for training dialogue response evaluation models, incorporating multiple references and adversarial responses.
Additionally, several approaches focus on evaluating the faithfulness of generated content relative to its context~\cite{fabbri2022qafacteval,laban-etal-2022-summac,zha-etal-2023-alignscore}. QAFactEval~\cite{fabbri2022qafacteval} and SCConv~\cite{laban-etal-2022-summac} train models to predict faithfulness scores using sentence-level natural language inference (NLI) scores as feature vectors. AlignScore~\cite{zha-etal-2023-alignscore} employs a more comprehensive approach by pre-training its evaluation model on seven established tasks: NLI, question answering, paraphrasing, fact verification, information retrieval, semantic similarity, and summarization.

\paragraph{(4) Reference-free and Context-free Supervised Methods:}
This category represents a specialized type of supervised learning-based evaluation that primarily assesses the inherent quality of generated text without relying on references or context. Examples include StoryER \cite{chen-etal-2022-storyer} and ASE-Eval \cite{taghipour-ng-2016-neural}, which are designed for text simplification, story generation, and Automated Essay Scoring (ASE), respectively. These applications typically operate with minimal contextual constraints, focusing instead on intrinsic quality measures of the generated content.

\subsubsection{Self-supervised Learning Approaches}
Due to the substantial resources required for creating human-annotated datasets, researchers have developed various self-supervised strategies to construct evaluation datasets automatically. These self-supervised approaches can be classified into four main categories: (1) text perturbation techniques, which introduce controlled modifications to existing texts; (2) negative sampling methods, which sample negative samples from existing text corpus; (3) generative augmentation approaches, which utilizes language models to generate samples; and (4) positive sample augmentation strategies, which generate positive samples to balance the training dataset distribution.
A comprehensive overview of these self-supervised learning-based evaluation methods is presented in Table~\ref{tab:self_supervised_learning_based}.

\begin{table*}[htbp]
    \centering
    \resizebox{\textwidth}{!}{
    \begin{tabular}{ccccccc}
        \toprule
         \textbf{Metric} &\textbf{NLG Task}&\textbf{\begin{tabular}[c]{@{}c@{}}Category\end{tabular}}&\textbf{\begin{tabular}[c]{@{}c@{}}Self-Supervised\\Strategy\end{tabular}} & \textbf{\begin{tabular}[c]{@{}c@{}}Need\\Reference\end{tabular}} & \textbf{\begin{tabular}[c]{@{}c@{}}Need\\Context\end{tabular}} & \textbf{\begin{tabular}[c]{@{}c@{}}Evaluation\\Protocols\end{tabular}}\\
         \midrule
         DENSITY~\cite{park-etal-2023-density}&Dialogue Generation&Text Perturbation&\begin{tabular}[c]{@{}c@{}}Repetition\\Contradiction\\Sensitive Concatenation\end{tabular}&No&Yes&Single\\\hline
         ReFreeEval~\cite{wu-etal-2023-holistic}&Machine Translation&Text Perturbation&Reorder&No&Yes&Single\\\hline
         UNION~\cite{guan-huang-2020-union}&Story Generation&Text Perturbation&\begin{tabular}[c]{@{}c@{}}Repetition\\Substitution\\Reorder\\Negation Alteration\end{tabular}&No&Yes&Single\\\hline
         LS Score~\cite{wu-etal-2020-unsupervised}&Text Summarization&Text Perturbation&\begin{tabular}[c]{@{}c@{}}Delete,Add\\Reorder\end{tabular}&No&Yes&Single\\\hline
         SEScore~\cite{xu-etal-2022-errors}&General Text Generation&Text Perturbation&\begin{tabular}[c]{@{}c@{}}Add,Delete,Reorder\\Mask-and-Fill\\Substitution\end{tabular}&Yes&No&Single\\\hline
         PoE~\cite{10056996}&Dialogue Generation&\begin{tabular}[c]{@{}c@{}}Text Perturbation\\Negative Sampling\\Generation\end{tabular}&\begin{tabular}[c]{@{}c@{}}Back-Translation\\Delete,Reorder\\Repetition\\Mask-and-Fill\end{tabular}&No&Yes&Single\\\hline
         MDD-Eval~\cite{zhang2022mddevalselftrainingaugmenteddata}&Dialogue Generation&\begin{tabular}[c]{@{}c@{}}Text Perturbation\\Negative Sampling\\Generation\end{tabular}&\begin{tabular}[c]{@{}c@{}}Delete,Reorder\\Repetition\\Back-Translation\\Mask-and-Fill\end{tabular}&No&Yes&Single\\\hline
         RoMe~\cite{rony-etal-2022-rome}&General Text Generation&Text Perturbation&Text Attack~\cite{morris-etal-2020-textattack}&Yes&No&Single\\\hline
         BARTScore++~\cite{lu2022humanlikeevaluationnaturallanguage}&General Text Generation&Text Perturbation&Mask-and-Fill&Yes&No&Single\\\hline
         USL-H~\cite{phy-etal-2020-deconstruct}&Dialogue Generation&\begin{tabular}[c]{@{}c@{}}Text Perturbation\\Negative Sampling\end{tabular}&\begin{tabular}[c]{@{}c@{}}Delete,Reorder\\Substitution\\Repetition\\Mask-and-Fill\end{tabular}&No&Yes&Single\\\hline
         \textsc{UniEval}~\cite{zhong-etal-2022-towards}&General Text Generation&Text Perturbation&\begin{tabular}[c]{@{}c@{}}Repetition\\Delete,Reorder\\Substitution\end{tabular}&Yes&Yes&Single\\\hline
         FineD-Eval~\cite{zhang-etal-2022-fined}&Dialogue Generation&Text Perturbation&Utterance Shuffle&No&Yes&Single\\\hline
         DynaEval~\cite{zhang-etal-2021-dynaeval}&Dialogue Generation&\begin{tabular}[c]{@{}c@{}}Text Perturbation\\Negative Sampling\end{tabular}&\begin{tabular}[c]{@{}c@{}}Utterance Replacement\\Utterance Shuffle\end{tabular}&No&Yes&Single\\\hline
         SEScore2~\cite{xu-etal-2023-sescore2}&General Text Generation&Text Perturbation&\begin{tabular}[c]{@{}c@{}}Delete,Add,Substitution\end{tabular}&Yes&No&Single\\\hline
         BCR~\cite{NEURIPS2023_a8b14855}&Dialogue Generation&Text Perturbation&Utterance Shuffle&No&Yes&Single\\
         RUBER~\cite{tao2017ruberunsupervisedmethodautomatic}&Dialogue Generation&Negative Sampling&Random Sampling&Yes&Yes&Single\\\hline
         PoNe~\cite{lan2020ponenovelautomaticevaluation}&Dialogue Generation&\begin{tabular}[c]{@{}c@{}}Negative Sampling\\Generation,PA\end{tabular}&\begin{tabular}[c]{@{}c@{}}EDA~\cite{wei-zou-2019-eda}\\Generation\end{tabular}&Yes&No&Single\\\hline
         USR~\cite{mehri-eskenazi-2020-usr}&Dialogue Generation&Negative Sampling&Random Sampling&No&Yes&Single\\\hline
         BERT-RUBER~\cite{ghazarian-etal-2019-better}&Dialogue Generation&Negative Sampling&Random Sampling&Yes&Yes&Single\\\hline
         EREDAT~\cite{le-scao-gardent-2023-joint}&Data-to-Text&Negative Sampling&\begin{tabular}[c]{@{}c@{}}In-batch\\Negative Sampling\end{tabular}&No&Yes&Single\\\hline
         GRADE~\cite{huang-etal-2020-grade}&Dialogue Generation&Negative Sampling&\begin{tabular}[c]{@{}c@{}}Lexical Sampling\\Embedding Sampling\end{tabular}&No&Yes&Single\\\hline
         D-Score~\cite{10.1109/TASLP.2021.3074012}&Dialogue Generation&\begin{tabular}[c]{@{}c@{}}Text Perturbation\\Negative Sampling\end{tabular}&\begin{tabular}[c]{@{}c@{}}Utterance Swap\\Reorder\\Random Sampling\end{tabular}&No&Yes&Single\\\hline
         LearnCompare~\cite{zhou2020learningcomparebettertraining}&Dialogue Generation&Generation&\begin{tabular}[c]{@{}c@{}}Generation of \\Past Checkpoint\end{tabular}&No&Yes&Pair\\
         \bottomrule
    \end{tabular}

    }
    \caption{\label{tab:self_supervised_learning_based}The complete list of the self-supervised learning-based evaluation methods.}
\end{table*}

\paragraph{(1) Text Perturbation:} This widely-adopted strategy generates negative samples through various text modification operations, including repetition~\cite{park-etal-2023-density}, insertion~\cite{wu-etal-2023-holistic}, deletion~\cite{zhong-etal-2022-towards,guan-huang-2020-union}, substitution~\cite{guan-huang-2020-union}, reordering~\cite{wu-etal-2020-unsupervised}, back-translation~\cite{zhang2022mddevalselftrainingaugmenteddata}, and mask-and-fill~\cite{zhang2022mddevalselftrainingaugmenteddata}.
For example, DENSITY~\cite{park-etal-2023-density}, ReFreeEval~\cite{wu-etal-2023-holistic}, LS Score~\cite{wu-etal-2020-unsupervised}, and \textsc{UniEval}~\cite{zhong-etal-2022-towards} apply random-span repetition, insertion, deletion, substitution, and reordering operations to generate low-quality text with intentional flaws. UNION~\cite{guan-huang-2020-union} introduces negation alteration specifically for story generation tasks. Additionally, MDD-Eval~\cite{zhang2022mddevalselftrainingaugmenteddata} and PoE~\cite{10056996} use back-translation and mask-and-fill operations to generate text that appears fluent but contains subtle flaws.
Recently, BCR~\cite{NEURIPS2023_a8b14855} constructed medium-coherence negative responses by shuffling utterances from one speaker in dialogues.

\paragraph{(2) Negative Sampling:} This strategy samples negative examples from large-scale corpora~\cite{mehri-eskenazi-2020-unsupervised,ghazarian-etal-2019-better}. For instance, RUBER~\cite{tao2017ruberunsupervisedmethodautomatic} randomly samples dialogue utterances from a corpus to create negative examples for each user query. Since random negative samples are usually easily distinguishable, many researchers have developed hard negative sampling strategies~\cite{xu-etal-2023-sescore2,zhang-etal-2022-fined,zhang-etal-2021-dynaeval,zhou2020learningcomparebettertraining,10.1109/TASLP.2021.3074012}. For example, PoNe~\cite{lan2020ponenovelautomaticevaluation} and GRADE~\cite{huang-etal-2020-grade} select context-similar and reference-similar generations as challenging negative samples.

\paragraph{(3) Generative Augmentation:} This approach uses pre-trained language models (PLMs) to generate text that is linguistically fluent but contextually inappropriate. MDD-Eval demonstrates this by employing DialoGPT~\cite{zhang2019dialogpt} to create such samples.

\paragraph{(4) Positive Sample Augmentation:} Unlike the previous methods that focus primarily on collecting negative samples, this approach aims to balance datasets by adding positive examples~\cite{zhang2022mddevalselftrainingaugmenteddata}. Both PoNe~\cite{lan2020ponenovelautomaticevaluation} and PoE~\cite{10056996} implement this strategy.

Despite the efficiency of self-supervised methods in generating training data, they often struggle with noise in the form of false negatives. For instance, generative augmentation and hard negative sampling methods sometimes inadvertently include appropriate generations in the negative sample set. To address this challenge, several works~\cite{lan2020ponenovelautomaticevaluation,zhang2022mddevalselftrainingaugmenteddata,10056996} have developed pseudo-labeling techniques to identify and remove these false negatives. PoNe~\cite{lan2020ponenovelautomaticevaluation}, for example, implements an iterative optimization algorithm specifically designed to detect such problematic samples.

\subsubsection{Mixture Approaches}
Supervised learning approaches typically outperform self-supervised methods due to high-quality human annotations. However, self-supervised techniques offer scalable dataset construction without expensive manual labeling. Mixture approaches combine the strengths of both paradigms, providing a practical compromise.
Representative works in this category include BLEURT~\cite{sellam-etal-2020-bleurt}, FrugalScore~\cite{kamal-eddine-etal-2022-frugalscore}, and MisMATCH~\cite{murugesan-etal-2023-mismatch}. These methods follow a two-stage training process: first pre-training BERT models on large-scale synthetic reference-candidate pairs generated through self-supervised techniques, then fine-tuning on smaller sets of human-annotated scores. This approach leverages the scale of self-supervised data while benefiting from the quality of human annotations, resulting in evaluation metrics that are both robust and cost-effective.

\subsubsection{Benchmark-based Evaluation}

In addition to automatic evaluation methods, benchmark-based evaluation represents a distinct assessment approach. This method evaluates NLG models by measuring their consistency with human annotations on question-answer pairs, thus assessing the models' general capabilities.
Since human annotation is completed as a one-time effort, subsequent model evaluations only require verifying the match between model-generated answers and human annotations. This verification can be performed through simple answer checking, exact matching, or various automatic evaluation methods.

Table~\ref{tab:benchmark_based_evaluation} provides a complete list of widely used benchmarks for evaluating model capabilities. It can be found that, existing benchmarks can be categorized into six main types \cite{2023opencompass}:
(1) \textbf{Knowledge}: Evaluates general and domain-specific knowledge (e.g., C-Eval \cite{huang2023ceval} for STEM and humanities, MedBench for medical domains)
(2) \textbf{Reasoning}: Tests logical capabilities in mathematics \cite{cobbe2021gsm8k} and coding tasks \cite{chen-etal-2021-training}
(3) \textbf{Understanding}: Assesses context and query comprehension through benchmarks like MT-Bench101 \cite{bai2024mt} and XSum \cite{narayan2018dont}
(4) \textbf{Long Text}: Measures abilities to handle extended contexts via LongBench \cite{bai2023longbench} and Ada-LEval \cite{wang2024adaleval}
(5) \textbf{Agent}: Tests automated task solving and function calling capabilities \cite{chen2023t,ma2024agentboard}
(6) \textbf{Instruction-Following}: Evaluates adherence to instructions and human alignment using metrics like IFEval \cite{zhou2023instruction}

\begin{table}[htbp]
    \centering
    \resizebox{\textwidth}{!}{
    \begin{tabular}{ccc|ccc}
        \toprule
         \textbf{Benchmarks} & \textbf{Category} & \textbf{Metrics}&\textbf{Benchmarks} & \textbf{Category} & \textbf{Metrics}\\\midrule
         CMMLU~\cite{li2023cmmlu}&Knowledge&Acc.&HumanEval-X~\cite{zheng2023codegeex}&Reasoning&Pass@K\\
         ARC~\cite{clark2018think}&Knowledge& Acc.&MBPP~\cite{austin2021program}&Reasoning&Pass@K\\
         AGIEval~\cite{zhong2023agieval}&Knowledge&Acc.&TACO~\cite{li2023taco}&Reasoning&Pass@K\\
         C-Eval~\cite{huang2023ceval}&Knowledge&Acc.&GSM8K~\cite{cobbe2021gsm8k}&Reasoning&Acc.\\
         MMLU~\cite{hendryckstest2021}&Knowledge&Acc.&MATH~\cite{hendrycksmath2021}&Reasoning&Acc.\\
         GAOKAO-Bench~\cite{Zhang2023EvaluatingTP}&Knowledge&Acc.&RACE~\cite{lai2017race}&Understanding&Acc.\\
         CFBenchmark~\cite{lei2024cfbenchmarkchinesefinancialassistant}&Knowledge&\begin{tabular}[c]{@{}c@{}}F1,\\Embedding-based\end{tabular}&LAMBDA~\cite{paperno2016lambada}&Understanding&Acc.\\
         LawBench~\cite{fei2023lawbenchbenchmarkinglegalknowledge}&Knowledge&Acc,F1,ROUGE&MT-Bench-101~\cite{bai2024mt}&Understanding&LLM-based\\
         MedBench~\cite{liu2024medbenchcomprehensivestandardizedreliable}&Knowledge&\begin{tabular}[c]{@{}c@{}}BLEU,ROUGE,\\Acc,F1\end{tabular}&XSum~\cite{narayan2018dont}&Understanding&ROUGE\\
         TriviaQA~\cite{JoshiTriviaQA2017}&Knowledge&EM,F1&HaluEval~\cite{li2023halueval}&Safety&Acc.\\
         CommonsenseQA~\cite{talmor2018commonsenseqa}&Knowledge&Acc.&SaftyBench~\cite{zhang2023safetybench}&Safety&Acc.\\
         NQ~\cite{kwiatkowski2019natural}&Knowledge&F1&SALAD-Bench~\cite{li-etal-2024-salad}&Safety&\begin{tabular}[c]{@{}c@{}}F1,Acc,\\LLM-based\end{tabular}\\
         GPQA~\cite{rein2024gpqa}&Knowledge&Acc.&Flames~\cite{huang2023flames}&Safety&Acc.\\
         TruthfulQA~\cite{lin2022truthfulqameasuringmodelsmimic}&Knowledge&LLM-based&LongBench~\cite{bai2023longbench}&LongText&ROUGE-L,F1\\
         BBH~\cite{suzgun2022challenging}&Reasoning&EM&BABILong~\cite{kuratov2024babilong}&LongText&Acc.\\
         HellaSwag~\cite{zellers2019hellaswag}&Reasoning&Acc.&Ada-LEval~\cite{wang2024adaleval}&LongText&Acc.\\
         TheoremQA~\cite{chen2023theoremqa}&Reasoning&Acc.&T-Eval~\cite{chen2023t}&Agent&\begin{tabular}[c]{@{}c@{}}F1,Acc,EM,\\Embedding-based\end{tabular}\\
         WinoGrande~\cite{sakaguchi2019winogrande}&Reasoning&Acc.&AgentBoard~\cite{ma2024agentboard}&Agent&PassRate\\
         BigCodeBench~\cite{zhuo2024bigcodebench}&Reasoning&Pass@K&IFEval~\cite{zhou2023instruction}&Instruction&Acc.\\
         HumanEval~\cite{chen2021codex}&Reasoning&Pass@K&InfoBench~\cite{qin2024infobench}&Instruction&\begin{tabular}[c]{@{}c@{}}Decomposed Requirements\\Following Ratio\end{tabular} \\
         AlignBench~\cite{liu2024alignbenchbenchmarkingchinesealignment}&Instruction&LLM-based&-&-&-\\
         \bottomrule
    \end{tabular}
    }
    \caption{The list of benchmarks for evaluating the generation models.}
    \label{tab:benchmark_based_evaluation}
\end{table}
\subsection{LLM-based Evaluation}
\label{sec:llm_based_evaluation}

Learning-based evaluation methods heavily rely on high-quality and diverse training samples, which limits their effectiveness in out-of-domain tasks and their ability to generalize across evaluation criteria. 
With the development of large language models (LLMs) that demonstrate strong instruction following, context understanding, and zero-shot generalization capabilities, researchers increasingly implement automatic evaluation systems based on these models. This approach is known as \textbf{\textit{LLM-based Evaluation}} or LLM-as-a-judge~\cite{li2024generationjudgmentopportunitieschallenges}.

Compared to previous heuristic, embedding-based, and learning-based automatic evaluation methods, LLM-based evaluation provides not only assessment scores but also textual rationales that analyze flaws in generated content and offer valuable revision suggestions. This results in a more fine-grained and interpretable evaluation paradigm.
Current LLM-based evaluation methods fall into two main categories~\cite{li-etal-2024-leveraging-large}: 
(1) Prompt-based methods, which incorporate evaluation guidelines in prompts to direct LLMs to function as annotators;
(2) Fine-tuning-based methods, which enhance the evaluation capabilities of smaller, more efficient LLMs to address the high computational costs associated with advanced models like GPT-4.

\subsubsection{Prompt-based Methods}

Through well-designed prompts, LLMs can effectively evaluate various natural language generation (NLG) tasks across different criteria~\cite{fu2023gptscoreevaluatedesire,lin-chen-2023-llm,es-etal-2024-ragas,tu2024spagent}. For example, \citet{wang2023chatgptgoodnlgevaluator} and \citet{liu2023gevalnlgevaluationusing} demonstrate that GPT-3.5 and GPT-4 perform well as zero-shot evaluators for diverse NLG tasks.
Building on these findings, researchers have proposed several techniques to enhance the robustness of LLM evaluation capabilities:

\paragraph{(1) Reference:} \citet{zheng2023judgingllmasajudgemtbenchchatbot} and \citet{badshah2024referenceguidedverdictllmsasjudgesautomatic} evaluate response quality by comparing it with reference responses.
Recently, BatchEval~\cite{yuan-etal-2024-batcheval} improves LLM-based evaluation by using in-batch examples as references.

\paragraph{(2) Criteria:} \citet{qian-etal-2024-large,lu2024exactmatchsemanticallyreassessing} provide detailed evaluation criteria and rubrics to guide LLM assessment processes.

\paragraph{(3) Demonstration:} GPTScore~\cite{fu2023gptscoreevaluatedesire}, ICE~\cite{jain-etal-2023-multi}, MSoR~\cite{song2025manyshotincontextlearninghelp}, Little Giants~\cite{kotonya-etal-2023-little}, and ALLURE~\cite{hasanbeig2023allureauditingimprovingllmbased} enhance LLM-based evaluation by incorporating few-shot demonstrations.

\paragraph{(4) Swapping:} Swapping operations are widely used to reduce positional bias in pair-wise evaluations~\cite{lan2024criticevalevaluatinglargelanguage,lee2024rlaifvsrlhfscaling,zheng2023judgingllmasajudgemtbenchchatbot}.

\paragraph{(5) Self-consistency:} \citet{zhang2024selfcontrastbetterreflectioninconsistent,zheng2023judgingllmasajudgemtbenchchatbot,manakul2023selfcheckgpt,cohen2023lm} show that LLM-as-a-judge performance improves with self-consistency prompting strategies, which sample multiple evaluation rationales and use majority voting for the final result.

\paragraph{(6) Multi-Agent:} Addressing single-model bias concerns, researchers have developed multi-agent debate frameworks for evaluation tasks, including ChatEval~\cite{chan2023chatevalbetterllmbasedevaluators}, PoLL~\cite{verga2024replacingjudgesjuriesevaluating}, and PRD~\cite{li2024prdpeerrankdiscussion}.

\paragraph{(7) Claim Decomposition:} Since evaluated responses may contain multiple independent claims, researchers propose decomposing responses into atomic information units~\cite{min2023factscore,sun2024critiquecritique} and verifying the quality of each unit individually before aggregating results to determine overall response quality.

In summary, most prompt-based methods described above propose prompting strategies to address the inconsistency problem in LLM outputs~\cite{liang2024internalconsistencyselffeedbacklarge,zhang2024selfcontrastbetterreflectioninconsistent,wang2023selfconsistency}.
LLMs achieve more accurate judgments when evaluating response quality by following guidelines and clues provided in demonstrations, detailed criteria, and reference responses. Additionally, self-consistency techniques and multi-agent frameworks are widely employed to reduce inconsistencies and biases inherent in single-model evaluations.

Although LLMs can achieve performance comparable to human judgments, their substantial inference costs limit applications in large-scale evaluation scenarios.
Recent research has focused on reducing these costs. 
For example, UniCBE~\cite{yuan2025unicbe} introduces a unified uniformity-driven CBE framework that optimizes tuple sampling and preference aggregation strategies in pairwise evaluation protocols.
Similarly, TailoredBench~\cite{yuan2025onesizefitsalltailoredbenchmarksefficient} proposes a customized evaluation approach tailored to each target model, achieving an average reduction of 31.4\% in Mean Absolute Error (MAE) of accuracy estimates under the same inference budget constraints.


\subsubsection{Fine-tuning-based Methods}
The most direct approach to reduce the cost of LLM-based evaluation is improving the evaluation capabilities of smaller, more efficient models.
Many researchers have developed fine-tuning methods to enhance the evaluation performance of small-scale LLMs.
These efficient evaluation models address the computational demands of large-scale evaluations, and are widely-used in the recent RLHF training procedure~\cite{mahan2024generativerewardmodels,lee2024rlaifvsrlhfscaling,ouyang2022traininglanguagemodelsfollow,deepseekai2024deepseekv3technicalreport}.
So far, current fine-tuning-based LLM evaluation methods can be categorized across four dimensions~\cite{zeng2024scalingsearchlearningroadmap,li-etal-2024-leveraging-large}: (1) Evaluation Interpretability; (2) Evaluation Granularity; (3) Optimization Methods; and (4) Data Sources.
Table~\ref{tab:fine_tuning_llm_evaluation} provides a comprehensive list of these methods, and we describe representative approaches from each dimension below\footnote{Table~\ref{tab:fine_tuning_llm_evaluation} includes selected representative reward models.}.

\paragraph{(1) Evaluation Interpretability:} 
Based on interpretability, current fine-tuned LLM-based evaluation models fall into two categories: reward models and critique models. Reward models~\cite{lambert2024rewardbenchevaluatingrewardmodels,ouyang2022traininglanguagemodelsfollow} are trained to mimic human preferences and serve as key components in Reinforcement Learning with Human Feedback (RLHF)~\cite{ouyang2022traininglanguagemodelsfollow,10.5555/3495724.3495977}. While these models help align LLMs with human preferences, they only provide numerical scores without explanatory feedback, which limits their reliability and interpretability.
In contrast, critique models deliver textual analysis of generated content, offering more detailed and interpretable feedback~\cite{tian-etal-2024-sheng,liu2024alignbenchbenchmarkingchinesealignment}. Recent advances have introduced generative reward models~\cite{ye2024improvingrewardmodelssynthetic,mahan2024generativerewardmodels}, such as Critic-RM~\cite{yu2024selfgeneratedcritiquesboostreward}, which produce chain-of-thought analysis before generating scores. This approach enhances both interpretability and training data efficiency.

\paragraph{(2) Evaluation Granularity:} 
Early evaluation approaches focused on whole-response assessment, known as outcome-based reward models (ORMs)~\cite{lightman2023letsverifystepstep}. Recent research~\cite{zeng2024scalingsearchlearningroadmap} has shown that fine-grained, process reward models (PRMs) offer better transparency and more effective feedback, though they require significant human labor for annotation~\cite{lightman2023letsverifystepstep}.
Most fine-tuning-based evaluation models remain outcome-based, assessing the correctness of entire responses~\cite{cui2023ultrafeedback,kim2024prometheus2opensource}. To improve transparency and explanations, some approaches identify specific flaws in responses and provide detailed analysis, including InstructScore~\cite{xu-etal-2023-instructscore}, TIGERScore~\cite{jiang2024tigerscorebuildingexplainablemetric}, and MultiCritique~\cite{lan2024traininglanguagemodelscritique}. However, these evaluation signals remain sparse since responses typically contain few flaws.

A growing trend involves providing denser evaluation signals for all intermediate steps in responses through process-based reward models (PRMs). For example, \citet{lightman2023letsverifystepstep} annotate the correctness of each intermediate reasoning step in mathematical solutions. While PRMs offer detailed feedback, annotating intermediate steps is significantly more resource-intensive than evaluating complete responses.
To address this challenge, several researchers employ Monte Carlo Tree Search (MCTS) to automatically generate process signals for reasoning tasks with known ground-truth answers, as seen in Math-Shepherd~\cite{wang-etal-2024-math} and PSRLM~\cite{zhang2025processbasedselfrewardinglanguagemodels}. Nevertheless, most existing PRMs are limited to reasoning tasks with definitive answers, and developing process-based evaluation models for broader domains remains a huge challenging~\cite{zeng2024scalingsearchlearningroadmap}.

\paragraph{(3) Optimization Methods:} 
Supervised fine-tuning (SFT) represents the most prevalent approach for enhancing LLMs' evaluation capabilities~\cite{li2023generativejudgeevaluatingalignment,cui2023ultrafeedback}. Notable examples include Auto-J~\cite{li2023generativejudgeevaluatingalignment}, UltraCM~\cite{cui2023ultrafeedback}, and Prometheus~\cite{kim-etal-2024-prometheus}, which optimize Llama models using synthetic datasets generated by GPT-4.
More recently, researchers have begun exploring preference learning techniques such as Direct Preference Optimization (DPO)~\cite{rafailov2024directpreferenceoptimizationlanguage} and Reinforcement Learning (RL)~\cite{ouyang2022traininglanguagemodelsfollow} to overcome the limitations of behavior cloning approaches. Behavior cloning often struggles with distributional shift and fails to capture the nuanced reasoning behind human preferences. Models like Critic-RM~\cite{wang2024selftaughtevaluators} and SFR-Judge~\cite{wang2024directjudgementpreferenceoptimization} demonstrate this shift by optimizing Llama models on preference evaluation datasets, resulting in more robust evaluation capabilities that better align with human judgment patterns.

\paragraph{(4) Data Sources:} 
Similar to the Learning-based evaluation methods described in Section~\ref{sec:learning_based_methods}, dataset is the core of fine-tuned-based LLM-based evaluation methods.
So far, the datasets for optimizing fine-tuned-based evaluation methods originate from three primary sources:
(1) \textbf{Human annotation:} This represents the most direct approach to constructing evaluation datasets. For instance, Shepherd~\cite{wang2023shepherd} employs domain experts to annotate textual evaluation samples for training Llama-7B, while CriticGPT~\cite{mcaleese2024llmcriticshelpcatch} develops a human-annotated preference critique dataset for training the GPT-4 model. Although human-annotated datasets provide reliable quality, their substantial cost significantly limits scalability, particularly when constructing detailed textual critique datasets.
(2) \textbf{Advanced Teacher Models:} Given the considerable resources required for human annotation of textual evaluations, numerous researchers utilize advanced teacher models such as GPT-4 to construct the synthetic dataset for training. Examples include Auto-J~\cite{li2024generationjudgmentopportunitieschallenges} and UltraCM~\cite{cui2023ultrafeedback}. However, these model-generated datasets often contain significant noise that may compromise the robustness of the resulting evaluation models. Recently, MultiCritique~\cite{lan2024traininglanguagemodelscritique} proposes to address the noise of one single model by aggregating the diverse critique opinions from multi-agent.
(3) \textbf{Human-machine collaborative annotation:} This hybrid approach combines the advantages of both previously mentioned methods. Specifically, LLMs generate draft evaluation samples, which are subsequently validated using human-annotated scores. For example, Themis~\cite{hu2024themisreferencefreenlgevaluation} and SFR~\cite{wang2024directjudgementpreferenceoptimization} filter high-quality GPT-4 generated samples by measuring the consistency between GPT-4's judgments and human assessments, resulting in more reliable training data while maintaining cost efficiency. This approach could leverage simple human-annotated evaluation scores or preferences as ground-truth to collect reliable detailed textual critiques, thereby avoiding the substantial expense of annotating detailed critiques from scratch.

\begin{table*}[htbp]
    \centering
    \resizebox{\textwidth}{!}{
    \begin{tabular}{cccccccc}
        \toprule
         \textbf{Metric} & \textbf{\begin{tabular}[c]{@{}c@{}}Interpre\\tability\end{tabular}}&\textbf{Granularity} & \textbf{Optimization} & \textbf{Data Source} & \textbf{Reference} & \textbf{Protocols} & \textbf{Base Model}\\
         \midrule
         CriticGPT~\cite{mcaleese2024llmcriticshelpcatch}&Yes&Outcome&RL&Human&No&Single&GPT-4\\
         PRM\cite{lightman2023letsverifystepstep}&No&Process&SFT&Human&No&Single&GPT-4\\
         Math-Shepherd~\cite{wang-etal-2024-math}&No&Process&SFT&Human-Model&No&Single&Llama2-70B\\
         Qwen2.5-Math-PRM-7B&No&Process&SFT&Human-Model&No&Single&Qwen2.5-7B\\
         InternLM2-20B-Reward~\cite{cai2024internlm2technicalreport}& No & Outcome &SFT&-&No&Single&InternLM2-20B\\
         Skyword-Reward-8B~\cite{liu2024skywork}& No & Outcome & SFT & - & No & Single & Llama-3.1-8B \\
         Skyword-Critic-8B~\cite{liu2024skywork}& No & Outcome & SFT & - & No & Pair & Llama-3.1-8B \\
         Auto-J~\cite{li2023generativejudgeevaluatingalignment}&Yes&Outcome&SFT&GPT-4&No&Single/Pair&Llama-2-13B\\
         UltraCM~\cite{cui2023ultrafeedback}&Yes&Outcome&SFT&GPT-4&No&Single&Llama-2-13B\\
         Shepherd~\cite{wang2023shepherd}&Yes&Outcome&SFT&Human&No&Single&Llama-7B\\
         Prometheus~\cite{kim2024prometheusinducingfinegrainedevaluation}&Yes&Outcome&SFT&GPT-4&Yes&Single&Llama2-13B\\
         Prometheus2~\cite{kim2024prometheus2opensource}&Yes&Outcome&SFT&GPT-4&Yes&Single/Pair&Llama2,Mistral\\
         Self-Taught~\cite{wang2024selftaughtevaluators}&Yes&Outcome&SFT&LLM&No&Pair&Llama3-70B\\
         Meta-Rewarding~\cite{wu2024metarewardinglanguagemodelsselfimproving}&Yes&Outcome&PL&LLM&No&Single&Llama3-8B\\
         InstructScore~\cite{xu-etal-2023-instructscore}&Yes&Process&SFT&GPT-4&No&Single&Llama-7B\\
         TIGERScore~\cite{jiang2024tigerscorebuildingexplainablemetric}&Yes&Process&SFT&GPT-4&Yes&Single&Llama2-7B/13B\\
         Themis~\cite{hu2024themisreferencefreenlgevaluation}&Yes&Outcome&SFT/PL&Human-Model&No&Single&Llama3-8B\\
         SFR~\cite{wang2024directjudgementpreferenceoptimization}&Yes&Outcome&SFT/PL&Llama3-70B&No&Pair&Llama3.1-8B/70B\\
         Critic-RM~\cite{yu2024selfgeneratedcritiquesboostreward}&Yes&Outcome&SFT/PL&Human-Model&No&Single&Llama3.1-70B\\
         MultiCritique~\cite{lan2024traininglanguagemodelscritique}&Yes&Process&SFT/PL&Multi-Agent&Yes&Single&InternLM2-7B\\
         PandaLM~\cite{wang2024pandalmautomaticevaluationbenchmark}&Yes&Outcome&SFT&GPT-3.5&Yes&Pair&Llama-7B\\
         JudgeLM~\cite{zhu2023judgelm}&Yes&Outcome&SFT&GPT-4&Yes&Single/Pair&Llama2\\
         CritiqueLLM~\cite{ke2024critiquellminformativecritiquegeneration}&Yes&Outcome&SFT&GPT-4&Yes&Single/Pair&ChatGLM3-6B\\
         CompassJudger~\cite{cao2024compassjudger1allinonejudgemodel}&Yes&Outcome&SFT&Human,GPT-4&No&Single/Pair&InternLM2.5-7B\\
         X-Eval~\cite{liu2024xevalgeneralizablemultiaspecttext}&Yes&Outcome&SFT&Human&No&Single/Pair&Llama-7B\\
         FLAMe~\cite{vu2024foundationalautoraterstaminglarge}&Yes&Outcome&SFT&Human&No&Single/Pair&PaLM-2-24B\\
         AttrScore~\cite{yue-etal-2023-automatic}&Yes&Outcome&SFT&Human&No&Single&Llama\\
         Self-Judge~\cite{lee2024aligninglargelanguagemodels}&Yes&Outcome&SFT&Human&No&Pair&Llama2-7B\\
         Self-Rationalize~\cite{trivedi2024selfrationalizationimprovesllmfinegrained}&Yes&Outcome&PL&LLM&No&Single&Llama3.1-8B\\
         SCRIT~\cite{tang2025enablingscalableoversightselfevolving}&Yes&Process&SFT&Human-Model&Yes&Single&Qwen2.5-72B\\
         ANAH~\cite{ji-etal-2024-anah}&Yes&Outcome&SFT&Human&No&Single&InternLM2-20B\\
         ANAH-v2~\cite{gu2024anah}&Yes&Outcome&SFT&LLM&No&Single&InternLM2-20B\\
         RAGTruth~\cite{niu-etal-2024-ragtruth}&No&Outcome&SFT&Human&No&Single&Llama2-13B\\
         TrueTeacher~\cite{gekhman2023trueteacher}&Yes&Outcome&SFT&PaLM 540B&No&Single&T5-11B\\
         PerSE~\cite{wang-etal-2024-learning-personalized}&Yes&Outcome&SFT&Human&No&Single/Pair&Llama-7B\\
         SorryBench~\cite{xie2024sorry}&No&Outcome&SFT&Human&No&Single&Llama3-8B\\
         MATHMinos~\cite{gao2024llm}&Yes&Process&SFT&Human-Model&No&Single&MetaMATH\\
         Halu-J~\cite{wang2024halujcritiquebasedhallucinationjudge}&Yes&Outcome&SFT/PL&Human-Model&No&Single&Mistral-7B\\
         Offsetbias~\cite{park-etal-2024-offsetbias}&No&Outcome&SFT&GPT-4&No&Pair&Llama3-8B\\
         DeepSeek-GRM~\cite{liu2025inferencetimescalinggeneralistreward}&Yes&Outcome&PL&Human&No&Single&DeepSeek-V3\\
         \bottomrule
    \end{tabular}
    }
    \caption{The list of the fine-tuning-based LLM-based evaluation models. As for the Data Source, Human-Model indicates that the training data are generated by using LLMs and verified by human annotations. The human annotations could be reference answers in reasoning tasks~\cite{tang2025enablingscalableoversightselfevolving} or human-annotated preference labels~\cite{wang2024directjudgementpreferenceoptimization,hu2024themisreferencefreenlgevaluation,yu2024selfgeneratedcritiquesboostreward}.}
    \label{tab:fine_tuning_llm_evaluation}
\end{table*}

\subsection{Comparing Automatic Evaluations}
\label{sec:comparing_automatic_evaluation}

In this section, we present both qualitative and quantitative analyses of existing automatic evaluation paradigms for text generation.

\subsubsection{Qualitative Analysis}

Table~\ref{tab:comparing_ae} summarizes our comparative analysis of existing automatic evaluation methods across four key dimensions: (1) Evaluation Flexibility and Generalization, which examines the adaptability of methods across different tasks and evaluation criteria; (2) Training Data Source, which identifies the origin and nature of data used to develop these methods; (3) Evaluation Granularity, which considers the level of detail in the assessment process; and (4) Evaluation Cost, which accounts for computational resources and time requirements. These dimensions provide a comprehensive framework for understanding the strengths and limitations of current automatic evaluation approaches.

\begin{table}[ht]
    \centering
    \resizebox{\textwidth}{!}{
        \begin{tabular}{cccccccc}
        \toprule
         \multicolumn{2}{c}{\textbf{Evaluation Methods}}&\textbf{\begin{tabular}[c]{@{}c@{}}Flexibility and\\Generalization\end{tabular}}&\textbf{\begin{tabular}[c]{@{}c@{}}Data\\Resource\end{tabular}}&\textbf{\begin{tabular}[c]{@{}c@{}}Evaluation\\Granularity\end{tabular}}&\textbf{\begin{tabular}[c]{@{}c@{}}Output\\Format\end{tabular}}&\textbf{\begin{tabular}[c]{@{}c@{}}Evaluation \\Cost\end{tabular}}&\textbf{\begin{tabular}[c]{@{}c@{}}Correlation\\with Human\end{tabular}}\\\midrule
         \multicolumn{2}{c}{\textbf{Benchmark-based Evaluation}}&\begin{tabular}[c]{@{}c@{}}Limited Task\\and Criteria\end{tabular}&Human Annotation&Instance-level&-&Small&Strong\\\midrule
         \multirow{3}{*}{\textbf{\begin{tabular}[c]{@{}c@{}}Heuristic\\Evaluation\end{tabular}}}
         &\textbf{Word-overlap}&\multirow{3}{*}{\begin{tabular}[c]{@{}c@{}}Limited Task\\and Criteria\end{tabular}}&-&\multirow{3}{*}{\begin{tabular}[c]{@{}c@{}}Instance-level\\Corpus-level\end{tabular}}&Scalar&Small&Weak\\
         &\textbf{Edit-Distance}&&-&&Scalar&Small&Weak\\
         &\textbf{Probability}&&-&&Scalar&Moderate&Weak\\\midrule
         \multirow{3}{*}{\textbf{\begin{tabular}[c]{@{}c@{}}Embedding-based\\Evaluation\end{tabular}}}
         &\textbf{Cosine Similarity}&\multirow{3}{*}{\begin{tabular}[c]{@{}c@{}}Limited Task\\and Criteria\end{tabular}}&-&\multirow{3}{*}{\begin{tabular}[c]{@{}c@{}}Instance-level\\Corpus-level\end{tabular}}&Scalar&Moderate&Weak\\
         &\textbf{WMD}&&-&&Scalar&Moderate&Weak\\
         &\textbf{Divergence}&&-&&Scalar&Moderate&Weak\\\midrule
         \multirow{4}{*}{\textbf{\begin{tabular}[c]{@{}c@{}}Learning-based\\Evaluation\end{tabular}}}
         &\textbf{Supervised}&\multirow{4}{*}{\begin{tabular}[c]{@{}c@{}}Limited Task\\and Criteria\end{tabular}}&\begin{tabular}[c]{@{}c@{}}Human-annotated Score\end{tabular}&\multirow{4}{*}{\begin{tabular}[c]{@{}c@{}}Instance-level\end{tabular}}&Scalar&Moderate&Moderate\\
         &\textbf{Self-Supervised}&&Synthetic Data&&Scalar&Moderate&Moderate\\
         &\textbf{Mixture}&&\begin{tabular}[c]{@{}c@{}}Human-annotated Score\\and Synthetic Data\end{tabular}&&Scalar&Moderate&Moderate\\\midrule
         \multirow{3}{*}{\textbf{\begin{tabular}[c]{@{}c@{}}LLM-based\\Evaluation\end{tabular}}}
         &\textbf{Prompt-based}&\multirow{3}{*}{\begin{tabular}[c]{@{}c@{}}General Task\\and Customized\\ Criteria\end{tabular}}&-&\multirow{3}{*}{\begin{tabular}[c]{@{}c@{}}Instance-level\\Process-level\end{tabular}}&\begin{tabular}[c]{@{}c@{}}Scalar and\\Rationale\end{tabular}&Huge&Strong\\
         &\textbf{Fine-tuning-based}&&\begin{tabular}[c]{@{}c@{}}Human Annotation\\or Synthetic Data\end{tabular}&&\begin{tabular}[c]{@{}c@{}}Scalar and\\Rationale\end{tabular}&Huge&Strong$^*$\\
         \bottomrule
        \end{tabular}
    }
    \caption{The comparison results of five evaluation paradigm on six dimensions. Scalar denotes the scalar-based evaluation outcome, such as scores and preference labels. Process-level indicates recent process-based reward models~\cite{zeng2024scalingsearchlearningroadmap} that evaluates the quality of each process in the generation.}
    \label{tab:comparing_ae}
\end{table}

\paragraph{Evaluation Flexibility and Generalization}
The evolution of NLG evaluation techniques shows a clear shift from task-specific methods toward more generalizable approaches. Early techniques (heuristic, embedding-based, and learning-based) were limited to specific tasks and criteria~\cite{fu2023gptscoreevaluatedesire}. In contrast, modern LLM-based methods achieve remarkable flexibility through simple prompt engineering, enabling general evaluation across diverse tasks and customized criteria~\cite{liu2024hdevalaligninglargelanguage}.

Recently, fine-tuned LLM-based automatic evaluation methods have emerged as alternatives to resource-intensive prompt-based approaches. However, these trained evaluation methods often sacrifice generalization capability, particularly for tasks and evaluation criteria not covered in their training data~\cite{lan2024traininglanguagemodelscritique}. 
The explicit learning of evaluation criteria represents a crucial direction for future development. Recent work has begun exploring this area. For example, HD-Eval~\cite{liu2024hdevalaligninglargelanguage} learns to construct a hierarchical criteria tree using human-annotation labels and demonstrates that this criteria structure improves LLM-based evaluation methods like G-Eval~\cite{liu2023gevalnlgevaluationusing}. Similarly, MultiCritique~\cite{lan2024traininglanguagemodelscritique} jointly learns customized criteria and evaluation generation within a multi-task learning framework, significantly enhancing evaluation generalization.

\paragraph{Training Data Source}

The reliability of learning-based or fine-tuning-based (LLM-based) automatic evaluation methods heavily depends on training data quality. 
As shown in Table~\ref{tab:comparing_ae}, training data sources come from either human annotation or synthetic data generation~\cite{wang2023shepherd,ye2024improvingrewardmodelssynthetic,hu2024themisreferencefreenlgevaluation}. 
Human annotation provides precise labels but incurs substantial costs that prohibit scaling. This limitation is especially significant for recent interpretable fine-tuning-based automatic evaluation methods. Synthetic data offers a cost-efficient alternative but suffers from quality limitations inherent to the generative models' capabilities.
To address this challenge, recent approaches have developed human-machine collaborative annotation method. Methods like SCRIT~\cite{tang2025enablingscalableoversightselfevolving} and Critic-RM~\cite{yu2024selfgeneratedcritiquesboostreward} validate synthetic data with human-annotated labels, creating more efficient and scalable solutions to improve evaluation capability.
As task complexity increases, a significant challenge emerges: neither teacher models nor human experts alone can provide sufficiently reliable supervision. Future research should focus on developing methods to accurately supervise generations for challenging tasks, such as complex reasoning problems or open-ended questions.

\paragraph{Evaluation Granularity}
The evolution of evaluation techniques has led to increasingly fine-grained assessment capabilities. Traditional methods (heuristic, embedding-based, and learning-based) employ outcome-based evaluation~\cite{zeng2024scalingsearchlearningroadmap}, providing holistic quality scores or binary preference labels without detailed justification.
In contrast, LLM-based evaluation methods leverage their advanced understanding and generation capabilities to deliver multi-dimensional assessments. These methods not only produce quantitative metrics such as numerical scores or preference rankings, but also provide fine-grained, explainable evaluation rationales~\cite{liu2023gevalnlgevaluationusing}. This explanatory capability operates at both process and step levels—examining the reasoning path and individual decision points rather than just the final output.

\paragraph{Evaluation Cost}
While evaluation techniques have become more advanced and robust, their computational cost has also increased dramatically, especially for recent LLM-based evaluation methods. Large foundation models like GPT-4 require significant computational resources, making large-scale evaluations prohibitively expensive and time-consuming for many research teams and organizations.

To address this efficiency challenge, fine-tuning-based evaluation methods have emerged as a promising solution within the LLM-based evaluation paradigm. These approaches aim to distill the sophisticated evaluation capabilities of large-scale LLMs like GPT-4 into smaller, more efficient models. The resulting compact evaluators significantly reduce inference time and computational requirements while maintaining comparable evaluation quality.
These smaller evaluation models offer two key advantages: First, they enable efficient large-scale evaluations that would be cost-prohibitive with larger models. Second, they can serve as more robust and reliable reward models in Reinforcement Learning from Human Feedback (RLHF) training pipelines~\cite{deepseekai2024deepseekv3technicalreport,ye2024improvingrewardmodelssynthetic}, where repeated evaluations are required during the optimization process. This balance between evaluation quality and computational efficiency represents an important direction for making advanced evaluation techniques more accessible and practical.

\subsubsection{Quantitative Analysis}

This section systematically examines performance differences among representative works across various automatic evaluation methods through comparative experiments, extending beyond qualitative analysis.
Before presenting our quantitative results, we introduce the concept of meta-evaluation—the process of evaluating automatic evaluation methods.

\paragraph{Meta-Evaluation}\label{sec:pre_meta_eval}

Meta-evaluation aims to evaluate the reliability of the automatic evaluation.
So far, the meta-evaluation could be classified into two categories: objective evaluation and subjective evaluation.

Meta-evaluation assesses the reliability of automatic evaluation methods, \textit{i.e.,} whether the automatic evaluation correlates with human judgments. Current meta-evaluation approaches fall into two main categories: objective and subjective evaluation~\cite{lan2024criticevalevaluatinglargelanguage}.

\noindent\textbf{(1) Objective Evaluation:} 
Objective meta-evaluation methods fall into the following categories:
\begin{itemize}
    \item \textbf{Single-wise evaluation protocol} measures correlation between automatic evaluation methods and human judgments. Spearman, Pearson, and Kendall correlation coefficients are widely used in established meta-evaluation benchmarks such as SummEval~\cite{fabbri2021summeval}, FED~\cite{mehri-eskenazi-2020-unsupervised}, and OpenMEVA~\cite{guan-etal-2021-openmeva}.
    \item \textbf{Pair-wise evaluation protocol} measures agreement (accuracy) between model-generated preferences $p$ and human-annotated preferences, like RewardBench~\cite{lambert2024rewardbenchevaluatingrewardmodels} and Auto-J~\cite{li2023generativejudgeevaluatingalignment}.
    \item \textbf{Corpus-wise evaluation protocol} calculates correlation between quality judgments from automatic evaluation methods and human annotators, where a quality judgment means choosing a particular (generative model and decoder) setting~\cite{NEURIPS2021_260c2432}.
\end{itemize}

\noindent\textbf{(2) Subjective Evaluation:} 
Subjective meta-evaluation methods primarily assess the quality of evaluation rationales. 
Recent studies employ GPT-4 as a judge to score rationale quality~\cite{ligenerative,cui2023ultrafeedback,wang2023shepherd,kim2023prometheus}.
However, due to the complexity of automatic evaluation tasks, even GPT-4 cannot consistently provide reliable subjective meta-evaluation~\cite{wang2023shepherd,mcaleese2024llmcriticshelpcatch}.
Recent work by CriticEval~\cite{lan2024criticevalevaluatinglargelanguage} and MetaCritique~\cite{sun2024critiquecritique} shows that GPT-4 can achieve reliable meta-evaluation when human-annotated evaluation rationales are provided as reference evaluation rationale (critiques).

Numerous meta-evaluation benchmarks have been proposed to measure the reliability of automatic evaluations. 
As described in Table~\ref{tab:meta_evaluation_benchmarks}, early meta-evaluation benchmarks primarily focused on specific NLG tasks, including machine translation~\cite{freitag-etal-2022-results,fu2023gptscoreevaluatedesire}, text summarization~\cite{fabbri2021summeval,Bhandari-2020-reevaluating,grusky-etal-2018-newsroom}, data-to-text generation~\cite{mairesse-etal-2010-phrase,wen-etal-2015-semantically,zhou-lampouras-2020-webnlg,lin-etal-2020-commongen}, dialog generation~\cite{mehri-eskenazi-2020-unsupervised,mehri-eskenazi-2020-usr}, and story generation~\cite{guan-etal-2021-openmeva,wang-etal-2022-paratag}.
With the advanced generation and generalization capabilities of LLMs, researchers have developed systematic general-domain meta-evaluation benchmarks to assess their evaluation capabilities, such as MT-Bench, Chat-Arena~\cite{zheng2023judging}, RewardBench~\cite{lambert2024rewardbenchevaluatingrewardmodels}, and RM-Bench~\cite{liu2024rmbenchbenchmarkingrewardmodels}.
Additionally, specialized meta-evaluation benchmarks measure LLMs' evaluation performance on challenging tasks, including reasoning~\cite{luo2023critiqueabilitylargelanguage,lin-etal-2024-criticbench,tan2024judgebenchbenchmarkevaluatingllmbased}, safety alignment~\cite{yuan2024rjudgebenchmarkingsafetyrisk}, and information retrieval~\cite{lu2024exactmatchsemanticallyreassessing}.

\begin{table}[htbp]
    \centering
    \resizebox{\textwidth}{!}{
    \begin{tabular}{cccc|cccc}
        \toprule
         \textbf{\begin{tabular}[c]{@{}c@{}}Meta-Evaluation\\Benchmark\end{tabular}} &\textbf{Task}&\textbf{Type} & \textbf{Protocols}&\textbf{\begin{tabular}[c]{@{}c@{}}Meta-Evaluation\\Benchmark\end{tabular}} &\textbf{Task}&\textbf{Type} & \textbf{Protocols}\\\midrule
         WMT-22~\cite{freitag-etal-2022-results} & MT & Objective&Single&RAEE~\cite{lu2024exactmatchsemanticallyreassessing} & EE & Objective&Single \\
         MQM-2020~\cite{fu2023gptscoreevaluatedesire} & MT  &Objective& Single&R-Judge~\cite{yuan2024rjudgebenchmarkingsafetyrisk}&Safety&\begin{tabular}[c]{@{}c@{}}Subjective\\\&Objective\end{tabular}&Pair\\
         SummEval~\cite{fabbri2021summeval}& SM &Objective&Single&FLASK~\cite{ye2024flaskfinegrainedlanguagemodel} & General& Objective&Single\\
         REALSumm~\cite{Bhandari-2020-reevaluating}& SM  &Objective&Single&Auto-J~\cite{li2023generativejudgeevaluatingalignment} & General& Objective&Pair\\
         NEWSROOM~\cite{grusky-etal-2018-newsroom}& SM &Objective&Single&MetaCritique~\cite{sun2024critiquecritique} & General &Subjective& Single\\
         QAGS\_XSUM~\cite{wang-etal-2020-asking}& SM &Objective&Single&CriticEval~\cite{lan2024criticevalevaluatinglargelanguage} & General&\begin{tabular}[c]{@{}c@{}}Subjective\\\&Objective\end{tabular} & \begin{tabular}[c]{@{}c@{}}Single\\\&Pair\end{tabular}\\
         BAGEL~\cite{mairesse-etal-2010-phrase} & D2T &Objective& Single&MT-Bench~\cite{zheng2023judging}&General&Objective&Pair\\
         SFRES~\cite{wen-etal-2015-semantically} & D2T &Objective& Single&Chatbot Arena~\cite{zheng2023judging}&General&Objective&Pair\\
         WebNLG~\cite{zhou-lampouras-2020-webnlg} & D2T &Objective& Single&MM-Eval~\cite{son2024mmevalmultilingualmetaevaluationbenchmark}&General&Objective&Single\\
         Commongen~\cite{lin-etal-2020-commongen} & D2T &Objective& Single&LLM-Judge-Eval~\cite{wei2024systematicevaluationllmasajudgellm}&General&Objective&Single\\
         FED~\cite{mehri-eskenazi-2020-unsupervised} & Dialog &Objective& Single&DHP~\cite{wang2024dhpbenchmarkllmsgood}&General&Objective&Pair\\
         TopicalChat~\cite{mehri-eskenazi-2020-usr} & Dialog&Objective& Single&EvalBiasBench~\cite{park2024offsetbiasleveragingdebiaseddata}&General&Objective&Pair\\
         PersonaChat~\cite{mehri-eskenazi-2020-usr} & Dialog&Objective& Single&KUDGE~\cite{son2024llmasajudgerewardmodel}&General&Objective&Single\\
         CBench~\cite{luo2023critiqueabilitylargelanguage} & Reason&Objective& Single&~\citet{murugadoss2024evaluatingevaluatormeasuringllms}&General&Objective&Single \\
         CriticBench~\cite{lin-etal-2024-criticbench} &Reason &Objective& Single&~\citet{thakur2024judgingjudgesevaluatingalignment}&General&Objective&Pair \\
         JudgeBench~\cite{tan2024judgebenchbenchmarkevaluatingllmbased}&Reason&Objective&Pair&CALM~\cite{ye2024justiceprejudicequantifyingbiases}&General&Objective&Single\\
         OpenMEVA~\cite{guan-etal-2021-openmeva} & SG &Objective& Single&RewardBench~\cite{lambert2024rewardbenchevaluatingrewardmodels}& General &Objective& Pair\\
         Per-MPST/DOC~\cite{wang2024learning} & SG & Objective&Single&RM-Bench~\cite{liu2024rmbenchbenchmarkingrewardmodels} & General &Objective& Pair\\
         LLMBar~\cite{zeng2024llmbar}& General & Objective & Pair&ProcessBench~\cite{zheng2024processbench}&Reason&Objective&Single \\
         RealCritique~\cite{tang2025realcriticeffectivenessdrivenevaluationlanguage}&Reason&Objective&Single&Feedback-Collection~\cite{kim2023prometheus}&General&Objective&Single\\
         Preference-Collection~\cite{kim-etal-2024-prometheus}&General&Objective&Pair&SummaC~\cite{laban-etal-2022-summac}&SM&Objective&Single\\
         \bottomrule
    \end{tabular}
    }
    \caption{The widely used meta-evaluation benchmarks for NLG tasks. MT, SM, D2T, Dialog, Reason, SG, EE and Safety represent the machine translation, text summarization, data-to-text, dialogue generation, logic reasoning, story generation, event extraction and safety NLG tasks. General task type indicate that the meta-evaluation benchmarks cover multiple diverse domains. 
    }
    \label{tab:meta_evaluation_benchmarks}
\end{table}

\paragraph{Systematic Meta-evaluation Results}

\begin{table*}[t]
    \centering
    \resizebox{\textwidth}{!}{
        \begin{tabular}{c|c|c|ccccccc}
        \toprule
         \multicolumn{3}{c|}{\textbf{Evaluation Methods}}&\begin{tabular}[c]{@{}c@{}}\textbf{Summ}\\\textbf{-Eval}\end{tabular}&\begin{tabular}[c]{@{}c@{}}\textbf{Topical}\\\textbf{-Chat}\end{tabular}&\textbf{FED}&\begin{tabular}[c]{@{}c@{}}\textbf{WMT}\\\textbf{-22}\end{tabular}&\begin{tabular}[c]{@{}c@{}}\textbf{Open}\\\textbf{MEVA}\end{tabular}&\textbf{BAGEL}&\begin{tabular}[c]{@{}c@{}}\textbf{Web}\\\textbf{-NLG}\end{tabular}\\\midrule
         \multirow{4}{*}{\textbf{\begin{tabular}[c]{@{}c@{}}Heuristic\\Evaluation\end{tabular}}}
         &\multirow{2}{*}{\textbf{Word-overlap}}&BLEU~\cite{papineni-etal-2002-bleu}&12.0&21.6&-&19.7&-1.17&16.8&20.7\\
         &&ROUGE-L~\cite{lin-2004-rouge}&14.5&23.7&24.4&17.8&2.34&14.2&35.5\\ \cline{2-10}\rule{0pt}{10pt}
         &\textbf{Edit-Distance}&TER~\cite{snover-etal-2006-study}&-12.0&1.07&-&21.95&6.20&-0.09&-0.08\\ 
         \cline{2-10}
         \rule{0pt}{10pt}
         &\textbf{Probability}&BARTScore~\cite{yuan2021bartscoreevaluatinggeneratedtext}&17.2&39.0&12.8&33.7&17.4&20.7&56.8\\
         \midrule
         \multirow{2}{*}{\textbf{\begin{tabular}[c]{@{}c@{}}Embedding-based\\Evaluation\end{tabular}}}
         &\textbf{Cosine Similarity}&BERTScore~\cite{Zhang2020BERTScore:}&23.7&32.3&27.3&42.4&2.9&28.2&50.4\\\cline{2-10}\rule{0pt}{10pt}
         &\textbf{WMD}&MoverScore~\cite{zhao-etal-2019-moverscore}&19.1&31.0&-&27.1&8.53&20.8&36.5\\\midrule
         \multirow{5}{*}{\textbf{\begin{tabular}[c]{@{}c@{}}Learning-based\\Evaluation\end{tabular}}}
         &\textbf{Supervised}&COMET22~\cite{rei-etal-2020-comet}&33.8&11.6&-&\textbf{56.4}&39.2&13.8&40.9\\\cline{2-10}\rule{0pt}{10pt}
         &\multirow{2}{*}{\textbf{Self-Supervised}}&UniEval~\cite{zhong-etal-2022-towards}&47.5&53.5&21.5&21.9&44.5&30.3&38.4\\
         &&SEScore2~\cite{xu-etal-2023-sescore2}&39.9&-37.9&-&44.9&30.6&32.5&48.4\\\cline{2-10}\rule{0pt}{10pt}
         &\multirow{2}{*}{\textbf{Mixture}}&BLEURT~\cite{sellam-etal-2020-bleurt}&17.3&38.8
         &-&48.4&27.5&22.9&16.8\\
         &&MisMatch~\cite{murugesan-etal-2023-mismatch}&41.0&-&-&-&-&-&49.0\\\midrule
         \multirow{9}{*}{\textbf{\begin{tabular}[c]{@{}c@{}}LLM-based\\Evaluation\end{tabular}}}
         &\multirow{3}{*}{\textbf{\begin{tabular}[c]{@{}c@{}}Fine-tuning-based\\Critique Models\end{tabular}}}&TIGERScore~\cite{jiang2024tigerscorebuildingexplainablemetric}&36.8&34.6&-&45.0&46.4&-&42.4\\
         &&InstructScore~\cite{xu-etal-2023-instructscore}&26.3&24.1&-&51.9&16.1&34.2&\textbf{59.0}\\
         &&Auto-J~\cite{li2023generativejudgeevaluatingalignment}&4.8&42.8&37.6&0.4&30.1&20.4&22.2\\\cline{2-10}\rule{0pt}{10pt}
         &\multirow{2}{*}{\textbf{\begin{tabular}[c]{@{}c@{}}Fine-tuning-based\\Reward Models\end{tabular}}}&InternLM2-20B-Reward&48.5&65.0&43.9&45.4&43.7&27.5&20.1\\
         &&Skywork-Reward-8B&44.3&43.3&42.3&30.1&39.1&25.9&25.5\\\cline{2-10}
         &\multirow{4}{*}{\textbf{Prompt-based}}&GPTScore~\cite{fu2023gptscoreevaluatedesire}&41.7&53.5&39.2&28.8&23.9&\textbf{41.3}&28.8\\
         &&G-Eval (GPT-4)~\cite{liu2023gevalnlgevaluationusing}&51.4&\textbf{73.2}&45.5&-&\textbf{47.5}&27.8&43.1\\
         &&DeepSeek-V3~\cite{deepseekai2024deepseekv3technicalreport}&\textbf{57.6}&66.4&53.6&-&44.9&39.4&42.2\\
         &&DeepSeek-R1~\cite{deepseekai2025deepseekr1incentivizingreasoningcapability}&52.1&64.9&\textbf{54.4}&-&42.9&38.0&41.5\\
         \bottomrule
        \end{tabular}
    }
    \caption{\label{tab:evaluation_results}Performance (Spearman correlation scores) of representative automatic evaluation methods on 7 NLG meta-evaluation benchmarks. Partial results remain empty due to the incomplete evaluations in previous studies. Due to the limited work on \textbf{Embedding-based - Divergence} methods, results are not feasible. }
\end{table*}

\begin{table*}[t]
    \centering
    \resizebox{\textwidth}{!}{
        \begin{tabular}{c|c|c|cccccc}
        \toprule
         \multicolumn{3}{c|}{\textbf{Evaluation Methods}}&\begin{tabular}[c]{@{}c@{}}\textbf{Critic}\\\textbf{-Bench}\end{tabular}&\begin{tabular}[c]{@{}c@{}}\textbf{CriticEval}\\\textbf{-Single}\end{tabular}&\begin{tabular}[c]{@{}c@{}}\textbf{CriticEval}\\\textbf{-Pair}\end{tabular}&\textbf{Auto-J}&\begin{tabular}[c]{@{}c@{}}\textbf{Preference}\\\textbf{-Bench}\end{tabular}&\begin{tabular}[c]{@{}c@{}}\textbf{Reward}\\\textbf{Bench}\end{tabular}\\\midrule
         \multirow{7}{*}{\textbf{\begin{tabular}[c]{@{}c@{}}LLM-based\\Evaluation\end{tabular}}}
         &\multirow{2}{*}{\textbf{\begin{tabular}[c]{@{}c@{}}Fine-tuning-based\\Critique Models\end{tabular}}}&Auto-J~\cite{li2023generativejudgeevaluatingalignment}&67.4&36.1&49.3&75.6&74.0&42.8\\
         &&UltraCM~\cite{cui2023ultrafeedback}&59.4&21.5&38.0&-&78.2&-\\\cline{2-9}\rule{0pt}{10pt}
         &\multirow{2}{*}{\textbf{\begin{tabular}[c]{@{}c@{}}Fine-tuning-based\\Reward Models\end{tabular}}}&InternLM2-20B-Reward&-&58.3&61.6&\textbf{85.6}&51.0&90.8\\
         &&Skywork-Reward-8B&-&52.7&51.2&76.8&51.0&\textbf{92.5}\\\cline{2-9}
         &\multirow{3}{*}{\textbf{Prompt-based}}&G-Eval (GPT-4o)~\cite{liu2023gevalnlgevaluationusing}&78.8&\textbf{68.2}&59.1&80.8&\textbf{89.0}&86.7\\
         &&DeepSeek-V3~\cite{deepseekai2024deepseekv3technicalreport}&73.6&57.6&71.2&74.9&87.8&91.9\\
         &&DeepSeek-R1~\cite{deepseekai2025deepseekr1incentivizingreasoningcapability}&\textbf{85.9}&65.0&\textbf{71.5}&77.1&88.8&91.1\\
         \bottomrule
        \end{tabular}
    }
    \caption{\label{tab:evaluation_results_2}Performance of LLM-based evaluation methods on 5 meta-evaluation benchmarks on general domains. Metric for CriticEval-single is the Spearman correlation, and metrics for other five meta-benchmarks are preference accuracy.
    }
\end{table*}

We performed systematic comparisons and evaluations of representative works from four automatic evaluation approaches (heuristic, embedding-based, learning-based, and LLM-based methods\footnote{Benchmark-based evaluation methods do not require meta-evaluation.}) across 12 key meta-evaluation benchmarks:
(1) SummEval~\cite{fabbri2021summeval} for text summarization; (2) Topical-Chat~\cite{mehri-eskenazi-2020-usr} for personalized dialogue generation; (3) FED~\cite{mehri-eskenazi-2020-unsupervised} for general open-domain dialogue generation; (4) WMT22~\cite{freitag-etal-2022-results} for machine translation; (5) OpenMEVA~\cite{guan-etal-2021-openmeva} for story generation; (6) BAGEL~\cite{mairesse-etal-2010-phrase} for data-to-text generation; (7) WebNLG~\cite{zhou-lampouras-2020-webnlg} for data-to-text generation; (8) CriticBench~\cite{lin-etal-2024-criticbench} for reasoning tasks; (9) CriticEval~\cite{lan2024criticevalevaluatinglargelanguage} for 9 diverse NLG tasks on single-wise (CriticEval-single) and pair-wise (CriticEval-pair) evaluation protocols; (10) Auto-J~\cite{li2024generationjudgmentopportunitieschallenges} for 58 diverse NLG tasks; (11) PreferenceBench~\cite{kim-etal-2024-prometheus} for diverse NLG tasks; and (12) RewardBench~\cite{lambert2024rewardbenchevaluatingrewardmodels} for evaluating reward models across chat, safty and reasoning tasks. Please refer to more details about these meta-evaluation benchmarks in Table~\ref{tab:meta_evaluation_benchmarks}.

Since the last five meta-evaluation benchmarks span diverse domains, we only tested representative LLM-based evaluation methods, as heuristic, embedding-based, and learning-based evaluation approaches cannot be effectively evaluated across these domains.
Based on the results presented in Table~\ref{tab:evaluation_results} and Table~\ref{tab:evaluation_results_2}, we can draw several important conclusions:

\begin{itemize}
    \item LLM-based automated evaluation methods currently outperform other approaches significantly. Across the seven benchmarks, LLM-based methods are only slightly less effective than COMET-22~\cite{rei-etal-2020-comet} on the WMT-22 benchmark for machine translation tasks.
    
    \item Fine-tuned LLM-based evaluation methods also substantially outperform heuristic, embedding-based, and learning-based methods. This demonstrates that distilling evaluation capabilities from strong LLMs can produce efficient, high-quality automated evaluation models.
    
    \item Table~\ref{tab:evaluation_results_2} shows that reward models fine-tuned on human-annotated preference datasets outperform interpretable critique models and even surpass prompt-based methods on several benchmarks (e.g., Auto-J and RewardBench). This highlights the effectiveness of human-annotated training data for evaluation tasks. However, Table~\ref{tab:evaluation_results} reveals that reward models significantly underperform compared to prompt-based methods on seven meta-evaluation benchmarks for specific NLG tasks. This finding indicates that generalization capability remains a major limitation of reward models. Future work should prioritize enhancing the generalization ability of these models.
    
    \item Recent reasoning models optimized for mathematical and coding questions, such as OpenAI o1 and DeepSeek-R1, demonstrate strong critique abilities when solving complex problems. This raises an important question: \textbf{Are reasoning models more suitable for evaluation than LLMs?} As shown in Tables~\ref{tab:evaluation_results} and~\ref{tab:evaluation_results_2}, the state-of-the-art reasoning model DeepSeek-R1 outperforms its base model DeepSeek-V3 and GPT-4 only on specific benchmarks like CriticBench and CriticEval-Pair. This indicates that reasoning models are not universally superior to traditional LLM evaluation methods. DeepSeek-R1 performs exceptionally well on CriticBench, which includes diverse meta-evaluation tasks focused on complex reasoning, suggesting it may be particularly suitable for evaluating generation quality in complex reasoning tasks.
\end{itemize}
\section{Automatic Evaluation for Vision Generation}
\tikzstyle{my-box}=[
    rectangle,
    draw=hidden-draw,
    rounded corners,
    text opacity=1,
    minimum height=1.5em,
    minimum width=5em,
    inner sep=2pt,
    align=center,
    fill opacity=.5,
]
\tikzstyle{leaf}=[my-box, minimum height=1.5em,
    fill=hidden-orange!60, text=black, align=left,font=\scriptsize,
    inner xsep=2pt,
    inner ysep=4pt,
]
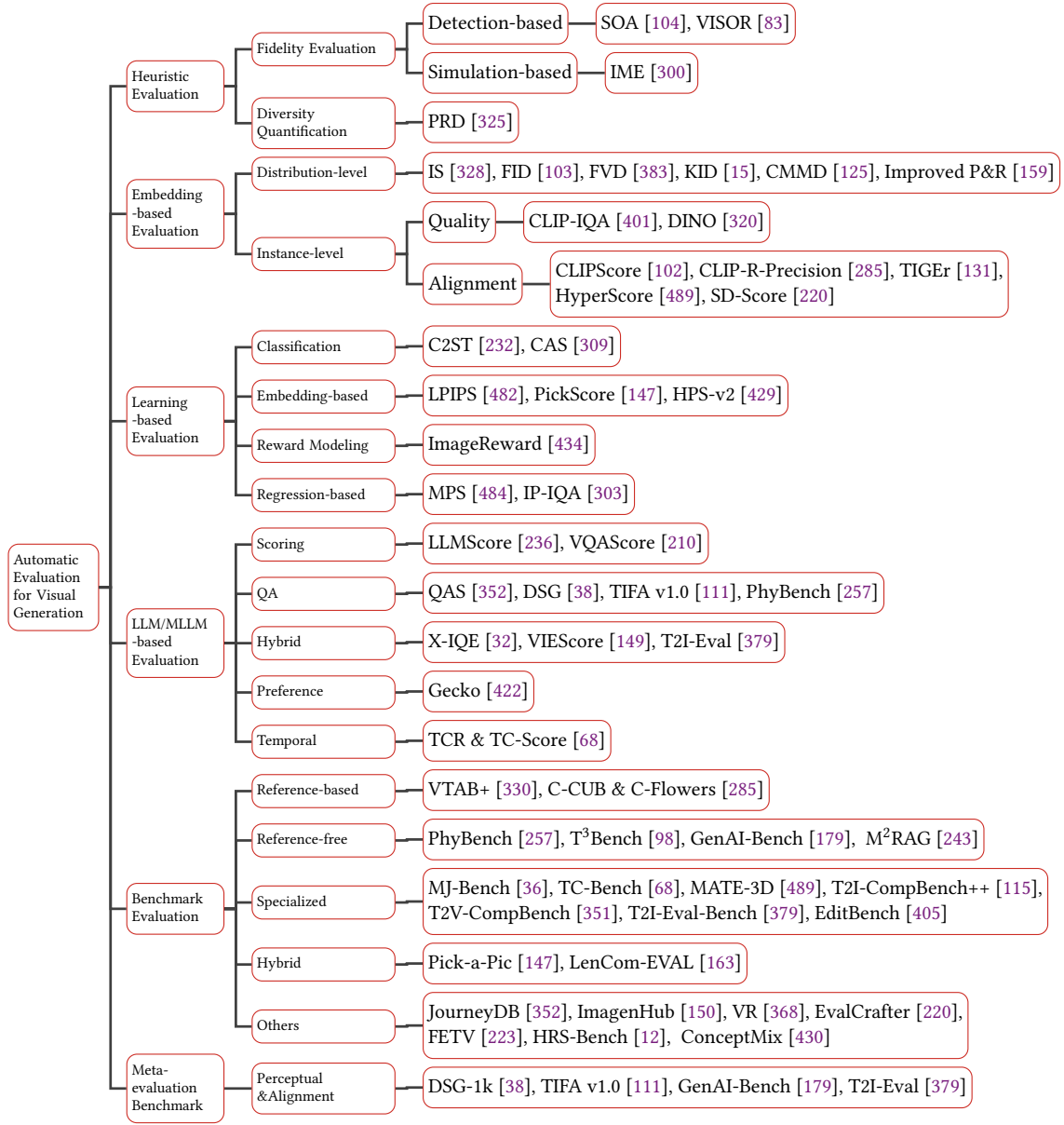
\begin{figure*}[t]
    \centering
    \resizebox{\textwidth}{!}{
        \begin{forest}
            forked edges,
            for tree={
                grow=east,
                reversed=true,
                anchor=base west,
                parent anchor=east,
                child anchor=west,
                base=left,
                font=\small,
                rectangle,
                draw=hidden-draw,
                rounded corners,
                align=left,
                minimum width=3em,
                edge+={darkgray, line width=1pt},
                s sep=3pt,
                inner xsep=2pt,
                inner ysep=3pt,
                ver/.style={rotate=90, child anchor=north, parent anchor=south, anchor=center},
            },
            where level=0{text width=3.5em,font=\scriptsize,}{},
            where level=1{text width=3.8em,font=\scriptsize,}{},
            where level=2{text width=5.8em,font=\scriptsize,}{},
            [ 
                Automatic \\Evaluation\\ for Visual \\ Generation
                [
                    Heuristic\\ Evaluation\\
                    [
                        Fidelity Evaluation\\
                        [
                            Detection-based\\
                            [
                                SOA~\cite{hinz2020semantic}{,}
                                VISOR~\cite{gokhale2022benchmarking}
                            ]
                        ]
                        [
                            Simulation-based\\
                            [
                                IME~\cite{qin2024worldsimbenchvideogenerationmodels}
                            ]
                        ]
                    ]
                    [
                        Diversity\\Quantification\\
                        [
                            PRD~\cite{sajjadi2018assessing}
                        ]
                    ]
                ]
                [
                    Embedding\\-based\\ Evaluation \\
                    [
                        Distribution-level
                        [
                            IS~\cite{salimans2016improved}{,}
                            FID~\cite{heusel2017gans}{,}
                            FVD~\cite{unterthiner2019fvd}{,}
                            KID~\cite{binkowski2018demystifying}{,}
                            CMMD~\cite{jayasumana2024rethinking}{,}
                            Improved P\&R~\cite{kynkaanniemi2019improved}
                        ]
                    ]
                    [
                        Instance-level
                        [
                            Quality
                            [
                                CLIP-IQA~\cite{wang2023exploring}{,}
                                DINO~\cite{ruiz2023dreambooth}
                            ]
                        ]
                        [
                            Alignment
                            [
                                CLIPScore~\cite{hessel2021clipscore}{,}
                                CLIP-R-Precision~\cite{park2021benchmark}{,}
                                TIGEr~\cite{jiang2019tiger}{,}\\
                                HyperScore~\cite{zhang2024benchmarking}{,}
                                SD-Score~\cite{liu2024evalcrafter}
                            ]
                        ]
                    ]
                ]
                [
                    Learning\\-based\\ Evaluation\\
                    [
                        Classification
                        [
                            C2ST~\cite{lopez2016revisiting}{,}
                            CAS~\cite{ravuri2019classification}
                        ]
                    ]
                    [
                        Embedding-based
                        [
                            LPIPS~\cite{zhang2018unreasonable}{,}
                            PickScore~\cite{kirstain2023pick}{,}
                            HPS-v2~\cite{wu2023human}
                        ]
                    ]
                    [
                        Reward Modeling
                        [
                            ImageReward~\cite{xu2024imagereward}
                        ]
                    ]
                    [
                        Regression-based
                        [
                            MPS~\cite{zhang2024learning}{,}
                            IP-IQA~\cite{qu2024bringing}
                        ]
                    ]
                ]
                [
                    LLM/MLLM\\-based\\ Evaluation\\
                    [
                        Scoring
                        [
                            LLMScore~\cite{lu2024llmscore}{,}
                            VQAScore~\cite{lin2025evaluating}
                        ]
                    ]
                    [
                        QA
                        [
                            QAS~\cite{sun2024journeydb}{,}
                            DSG~\cite{cho2023davidsonian}{,}
                            TIFA v1.0~\cite{hu2023tifa}{,}
                            PhyBench~\cite{meng2024phybench}
                        ]
                    ]
                    [
                        Hybrid
                        [
                            X-IQE~\cite{chen2023x}{,}
                            VIEScore~\cite{ku2023viescore}{,}
                            T2I-Eval~\cite{tu2024automatic}
                        ]
                    ]
                    [
                        Preference
                        [
                            Gecko~\cite{wiles2024revisiting}
                        ]
                    ]
                    [
                        Temporal
                        [
                            TCR \& TC-Score~\cite{feng2024tc}
                        ]
                    ]
                ]
                [
                    Benchmark\\Evaluation\\
                    [
                        Reference-based
                        [
                            VTAB+~\cite{schuhmann2022laion}{,}
                            C-CUB \& C-Flowers~\cite{park2021benchmark}
                        ]
                    ]
                    [
                        Reference-free
                        [
                            PhyBench~\cite{meng2024phybench}{,}
                            T$^3$Bench~\cite{he2023t}{,}
                            GenAI-Bench~\cite{li2024genai}{, }
                            M$^2$RAG~\cite{ma2025multimodalretrievalaugmentedmultimodal}
                        ]
                    ]
                    [
                        Specialized
                        [
                            MJ-Bench~\cite{chen2024mj}{,}
                            TC-Bench~\cite{feng2024tc}{,}
                            MATE-3D~\cite{zhang2024benchmarking}{,}
                            T2I-CompBench++~\cite{huang2023t2i}{,}\\
                            T2V-CompBench~\cite{sun2024t2v}{,}
                            T2I-Eval-Bench~\cite{tu2024automatic}{,}
                            EditBench~\cite{wang2023imagen}
                        ]
                    ]
                    [
                        Hybrid
                        [
                            Pick-a-Pic~\cite{kirstain2023pick}{,}
                            LenCom-EVAL~\cite{lakhanpal2024refining}
                        ]
                    ]
                    [
                        Others
                        [
                            JourneyDB~\cite{sun2024journeydb}{,}
                            ImagenHub~\cite{ku2023imagenhub}{,}
                            VR~\cite{tewel2022zerocap}{,}
                            EvalCrafter~\cite{liu2024evalcrafter}{,}\\
                            FETV~\cite{liu2024fetv}{,}
                            HRS-Bench~\cite{bakr2023hrs}{, }
                            ConceptMix~\cite{wu2024conceptmix}
                        ]
                    ]
                ]
                [
                    Meta-\\evaluation\\Benchmark\\
                    [
                        Perceptual\\\&Alignment\\
                        [
                            DSG-1k~\cite{cho2023davidsonian}{,}
                            TIFA v1.0~\cite{hu2023tifa}{,}
                            GenAI-Bench~\cite{li2024genai}{,}
                            T2I-Eval~\cite{tu2024automatic}
                        ]
                    ]
                ]
            ]
        \end{forest}}
    \caption{Taxonomy of representative automatic evaluation methods in Visual Generation.}
    \label{fig:taxonomy_visual}
\end{figure*}

Before this section, we have systematically survey automatic evaluation methods for text generation. Building on this foundation, we now extend the taxonomy to automatic evaluation techniques for vision generation tasks, like text-to-image and text-to-video generation tasks.
As illustrated in Figure~\ref{fig:taxonomy_visual}, existing automatic evaluation for vision generation have progressed through five paradigms:
\textbf{(1) Heuristic Evaluation}: Relies on handcrafted rules or features (e.g., pixel‐wise differences) to quantify simple aspects of visual content quality;
\textbf{(2) Embedding‐based Evaluation}: Leverages learned visual feature embeddings from deep neural networks to assess perceptual or semantic similarity between generated and reference content;
\textbf{(3) Learning‐based Evaluation}: Trains models to predict human‐annotated quality scores, aligning metric outputs more closely with human judgment;
\textbf{(4) LLM/MLLM‐based Evaluation}: Utilizes LLMs and MLLMs with tailored prompts to perform nuanced assessment across diverse visual evaluation criteria;
\textbf{(5) Benchmark Evaluation}: Employs curated datasets and gold‐standard references to directly compare system outputs against established performance baselines.

After describing these works, we then introduce the most widely used benchmark suites for evaluating visual content (Section~\ref{subsec:vision_heuristic}–\ref{subsec:vision_meta_eval}). Finally, we discuss current challenges and outline promising directions for future research (Section~\ref{subsec:challenges_future_visual}).

\subsection{Heuristic Evaluation\label{subsec:vision_heuristic}}

Heuristic evaluation of vision generation relies on predefined metrics and standardized procedures—either computed algorithmically or via structured human inspection—to assess model outputs. In our taxonomy, each method is characterized along three axes: (1) \emph{reference dependency}, (2) \emph{evaluation protocol}, and (3) \emph{focus} (fidelity vs.\ diversity). Table~\ref{tab:heuristic_based_evaluation_vision} summarizes representative methods according to these criteria and their underlying metric types.

\subsubsection{Fidelity Evaluation}
Fidelity evaluation measures how well generated content semantically aligns with given conditions (e.g., text prompts). Most approaches leverage pretrained detectors or simulative models to verify object presence, attributes, and cross‑modal consistency:
\begin{itemize}
  \item \textbf{Detection‑based metrics.} Semantic Object Accuracy (SOA) \cite{hinz2020semantic} uses object detectors to confirm target entities, while VISOR \cite{gokhale2022benchmarking} extends this by validating spatial relationships among detected objects.
  \item \textbf{Simulation‑based metrics.} Implicit Manipulative Evaluation (IME) \cite{qin2024worldsimbenchvideogenerationmodels} assesses video generation quality by simulating agent actions with video‑to‑action models, probing the model’s world‑simulation capabilities.
\end{itemize}

\subsubsection{Diversity Quantification}
Diversity metrics evaluate the variety and coverage of generated outputs, often by comparing distributions or estimating entropy across samples:
\begin{itemize}
  \item \textbf{Distributional comparisons.} Precision‑Recall for Distributions (PRD) \cite{sajjadi2018assessing} analyzes precision‑recall curves between real and generated data distributions, offering a more nuanced view than single‑value scores.
  \item \textbf{Entropy‑based measures.} Entropy-based heuristics quantify an image’s information content to proxy perceptual quality without reference. For example, ENIQA~\cite{chen2019no} computes entropy in the spatial and frequency domain to predict a continuous quality score. Despite its simplicity, ENIQA achieves strong agreement with human judgments, demonstrating that entropy alone can effectively capture both loss of detail and distortion diversity.
\end{itemize}

\begin{table}[htbp]
    \centering
    \resizebox{\textwidth}{!}{
    \begin{tabular}{cccccc}
        \toprule
        \textbf{Method} & \textbf{Task} & \textbf{Objective} & \textbf{Reference} & \textbf{Protocols} & \textbf{Metric Type} \\
        \midrule
        SOA~\cite{hinz2020semantic} & Text-to-Image Generation & Fidelity & No & Single & Detection \\
        TIAM~\cite{grimal2024tiam} & Text-to-Image Generation & Fidelity & No & Corpus & Hybrid \\
        VISOR~\cite{gokhale2022benchmarking} & Text-to-Image Generation & Fidelity & No & Corpus & Detection \\
        PRD~\cite{sajjadi2018assessing} & Text-to-Image Generation & Diversity & Yes & Single & Distribution \\
        IME~\cite{qin2024worldsimbenchvideogenerationmodels} & Text-to-Video Generation & Fidelity & No & Single & Simulation \\
        \bottomrule
    \end{tabular}
    }
    \caption{Heuristic evaluation methods for visual generative tasks, categorized by focus (fidelity vs.\ diversity), reference dependency, evaluation protocol, and metric type.}
    \label{tab:heuristic_based_evaluation_vision}
\end{table}

\subsubsection{Limitations and Future Directions}
While straightforward and computationally efficient, heuristic metrics have notable drawbacks:
\begin{itemize}
  \item \textbf{Bias and coverage.} Detection‑based approaches inherit biases from pretrained models and may overlook novel or out‑of‑vocabulary concepts.
  \item \textbf{Human correlation.} Many metrics correlate weakly with human judgments of aesthetics and realism.
  \item \textbf{Single‑axis focus.} Existing methods often emphasize either fidelity or diversity, but not both simultaneously.
\end{itemize}
Recent efforts such as TIAM’s multi‑stage verification and VISOR’s relational checks point toward richer, multi‑facet evaluation frameworks. Future work should aim to integrate reference‑free perceptual metrics, joint fidelity–diversity measures, and learned evaluators that better mirror human preferences.

\subsection{Embedding-based Evaluation}

Embedding-based evaluation methods leverage semantic representations learned by deep neural networks to assess generated visual content with greater flexibility and semantic depth. These methods have evolved along two complementary research tracks:
\begin{itemize}
  \item \textbf{Distribution‐level metrics}, which compare aggregate feature distributions between real and generated outputs, progressing from early CNN‐based divergences to sophisticated, unbiased, and spatiotemporal measures.
  \item \textbf{Instance‐level metrics}, which evaluate each sample individually, split into \textbf{Quality metrics}—assessing visual fidelity—and \textbf{Alignment metrics}—measuring consistency with textual or temporal references—by harnessing advances in vision–language and self‐supervised embeddings.
\end{itemize}

\subsubsection{Distribution‐level Methods}

These metrics offer a global assessment by comparing statistical properties of real versus generated feature distributions. Over time, research has introduced more robust divergences, reduced bias, and extended coverage to new domains such as video.
The Inception Score (IS)~\cite{salimans2016improved} is the first to introduce embedding models in image evaluation task. It computes the KL divergence between the conditional label distribution and its marginal over generated samples:
\begin{equation}
    \mathrm{IS}(G)=\exp\bigl[\mathbb{E}_{x\sim p_g}\,D_{KL}(p(y\mid x)\;\|\;p(y))\bigr].
\end{equation}
This pioneering metric captures both sample fidelity (low‐entropy $p(y\mid x)$) and diversity (high‐entropy $p(y)$), though it can be insensitive to mode collapse when different modes share labels \cite{salimans2016improved}.
Building upon the dual focus of IS, Fréchet Inception Distance (FID)~\cite{heusel2017gans} models real and generated Inception embeddings as Gaussians $\mathcal{N}(\mu,\Sigma)$ and $\mathcal{N}(\mu',\Sigma')$, then computes the Fréchet (2‐Wasserstein) distance between them:
\begin{equation}
    \mathrm{FID}=\|\mu-\mu'\|^2 + \mathrm{Tr}(\Sigma+\Sigma' - 2(\Sigma\Sigma')^{1/2}).
\end{equation}
This approach directly compares feature distributions and demonstrates stronger correlation with human judgment by mitigating mode collapse \cite{heusel2017gans}.
To relax the Gaussian assumption in FID, Kernel Inception Distance (KID)~\cite{binkowski2018demystifying} employs an unbiased maximum‐mean‐discrepancy (MMD) estimator with a polynomial kernel applied to Inception features. This substitution eliminates finite‐sample bias while maintaining reliable distribution similarity estimation \cite{binkowski2018demystifying}.
Aiming to disentangle fidelity from diversity, the Improved Precision \& Recall~\cite{kynkaanniemi2019improved} estimates precision (sample quality) and recall (coverage) manifolds nonparametrically. By evaluating these metrics separately, it provides clearer insight into generative performance \cite{kynkaanniemi2019improved}.
Extending the FID framework to video, Fréchet Video Distance (FVD) calculates the Fréchet distance over 3D convolutional features across frames. This extension captures both spatial and temporal aspects, aligning closely with human evaluations of video quality \cite{unterthiner2019fvd}.
CMMD~\cite{jayasumana2024rethinking} moves beyond simple Gaussian assumptions by pairing rich CLIP embeddings with a kernel-based two-sample test. Instead of modeling only means and covariances, it computes the Maximum Mean Discrepancy (MMD) using a Gaussian RBF kernel over CLIP features, allowing it to detect complex distributional differences between real and generated images without any normality assumptions—resulting in a more robust, unbiased, and sample-efficient quality metric.

\subsubsection{Instance‐level Methods}

While distribution-level metrics capture global trends across many samples, instance-level methods focus on evaluating each generated image (or video) individually by leveraging rich embedding representations. These approaches break down into two complementary categories: \textbf{(1) Quality Metrics}, which assess the perceptual and structural fidelity of each sample, and \textbf{(2) Alignment Metrics}, which measure how well an image (or frame) matches its textual or temporal reference.

\paragraph{Quality Metrics.}
Quality metrics aim to predict human perceptual judgments on a per-image basis. CLIP-IQA~\cite{wang2023exploring} operates in CLIP’s shared image–text embedding space, using carefully crafted text prompts (e.g., antonym pairs) to produce zero-shot scores for both aesthetic and technical quality; these scores have been shown to closely mirror human evaluations of image fidelity . In scenarios where the generation is driven by a specific subject (e.g., ``insert this person into a new scene''), the DINO score~\cite{ruiz2023dreambooth} employs self-supervised DINO features to quantify how faithfully the model preserves that subject’s appearance, tying structural consistency in the embedding space to perceived visual fidelity .

\paragraph{Alignment Metrics.}
Alignment metrics evaluate semantic consistency between the generated visual content and its prompt (or, in video, its temporal context). CLIPScore~\cite{hessel2021clipscore} computes the cosine similarity between an image’s CLIP embedding and the embedding of its prompt text, enabling robust, reference-free assessment of text-to-image alignment without any human-provided ground truth . Extending this retrieval perspective, CLIP-R-Precision~\cite{park2021benchmark} ranks a set of candidate captions by their similarity scores to a given image, offering a retrieval-based measure of alignment in text-to-image tasks . TIGEr~\cite{jiang2019tiger} adapts cross-modal retrieval techniques to measure semantic alignment more broadly, comparing images and descriptions across diverse scenarios for prompt-to-image consistency . For 3D asset generation, HyperScore~\cite{zhang2024benchmarking} projects embeddings into a hyperbolic space—capturing hierarchical and geometric relationships—to assess spatial coherence in generated geometry . Finally, for dynamic content in text-to-video tasks, SD-Score~\cite{liu2024evalcrafter} combines frame-level image–text alignment with a temporal coherence penalty, thus reflecting both visual quality and motion fidelity over time .

\subsubsection{Summary and Future Directions}

Embedding-based vision metrics split into \textit{distribution-level} and \textit{instance-level} approaches. Distribution-level methods (e.g., FID, KID, FVD) compare summary statistics of deep features from real and generated data, providing a quick global quality check but sometimes overlooking fine details .

Instance-level methods use per-sample embeddings to assess either \textit{visual fidelity} (e.g., CLIP-IQA, DINO score) or \textit{prompt alignment} (e.g., CLIPScore, CLIP-R-Precision). These zero-shot metrics correlate well with human judgments on individual images and videos without requiring ground-truth references .

Together, these techniques offer a flexible, reference-free toolkit: distribution tests flag large-scale mode collapse or drift, while instance-level scores diagnose per-sample quality and relevance. Future work will likely unify both views into single, efficient metrics that capture overall fidelity, semantic alignment, and perceptual subtlety.

\begin{table}[htbp]
  \centering
  \resizebox{0.9\textwidth}{!}{
  \begin{tabular}{lllll}
    \toprule
    \textbf{Method} & \textbf{Task} & \textbf{Category} & \textbf{Reference} & \textbf{Protocol} \\
    \midrule
    \multicolumn{5}{l}{\emph{Distribution‑level}} \\
    \midrule
    IS~\cite{salimans2016improved} & Text-to-Image Generation & Distribution & Yes  & Corpus \\
    FID~\cite{heusel2017gans}   & Text-to-Image Generation & Distribution & Yes & Corpus \\
    KID~\cite{binkowski2018demystifying} & Text-to-Image Generation & Distribution & Yes & Corpus \\
    Improved P\&R~\cite{kynkaanniemi2019improved} & Text-to-Image Generation & Distribution & Yes & Corpus \\
    CMMD~\cite{jayasumana2024rethinking} & Text-to-Image Generation & Distribution & Yes & Corpus \\
    FVD~\cite{unterthiner2019fvd} & Video Generation & Distribution & Yes & Corpus \\
    \midrule
    \multicolumn{5}{l}{\emph{Instance‑level: Quality Metrics}} \\
    \midrule
    CLIP‑IQA~\cite{wang2023exploring} & Text-to-Image Generation & Quality‑specific & No & Single \\
    DINO~\cite{ruiz2023dreambooth}    & Subject‑Driven Generation & Quality‑specific & Yes & Single \\
    AC-T2I~\cite{bakr2023hrs} & Text-to-Image Generation & Fidelity & Yes & Single \\
    \midrule
    \multicolumn{5}{l}{\emph{Instance‑level: Alignment Metrics}} \\
    \midrule
    CLIPScore~\cite{hessel2021clipscore} & Text-to-Image Generation & Similarity & No & Single \\
    CLIP‑R‑Precision~\cite{park2021benchmark} & Text-to-Image Generation & Retrieval & No & Single \\
    TIGEr~\cite{jiang2019tiger}        & Image Captioning & Semantic & Yes & Single \\
    HyperScore~\cite{zhang2024benchmarking} & Text-to-3D Generation & Geometric & No & Single \\
    SD‑Score~\cite{liu2024evalcrafter} & Text-to-Video Generation & Temporal & Yes & Single \\
    \bottomrule
  \end{tabular}
  }
  \caption{Embedding‑based evaluation metrics for visual generative tasks, grouped by distribution‑ vs. instance‑level approaches, with details on reference requirements and evaluation protocol.}
  \label{tab:embedding_based_evaluation}
\end{table}

\subsection{Learning-based Evaluation}

Learning‐based evaluation trains neural predictors on human‐annotated data to approximate perceptual judgments. Below we organize four major paradigms—classification, embedding, reward‐modeling, and regression—each introduced with a brief overview, followed by concise descriptions of representative methods.

\paragraph{Classification‐based Methods.}
These methods pose quality assessment as a discrimination task: a classifier is trained to distinguish real from generated images or to recognize high- versus low-quality outputs, and its accuracy serves as the evaluation score. C2ST~\cite{lopez2016revisiting} trains a binary classifier on mixed real and synthetic samples; under the null that the two distributions match, test‐set accuracy should be around 50 \%, so any deviation quantifies perceptual divergence.
CAS~\cite{ravuri2019classification} builds a classifier using only generated image–label pairs and measures its Top-1/Top-5 accuracy on real data; the drop in accuracy reveals semantic or class‐conditional fidelity gaps in conditional generators.

\paragraph{Embedding‐based Methods.}
Embedding‐based approaches reuse pre‐trained vision–language encoders to map images (and often their prompts) into a shared feature space, where similarity or distance estimates quality. PickScore~\cite{kirstain2023pick} fine-tunes a CLIP backbone on 500 K user preference judgments, achieving superhuman correlation with human rankings and supporting both absolute scores and pairwise comparisons. HPS-v2~\cite{wu2023human} adapts CLIP on the HPD v2 corpus of 798 K human preference pairs, predicting which image in a pair users prefer across diverse models and data; it generalizes robustly and is entirely reference-free.

\paragraph{Reward-modeling Methods.}
Reward models learn to predict human preferences directly from pairwise comparisons, yielding a scalar reward that reflects nuanced quality distinctions.
ImageReward~\cite{xu2024imagereward} trained on 137 K expert‐curated comparisons, it outperforms CLIP and aesthetic predictors by over 30\% in matching human choices, and can also serve as a reinforcement‐learning signal to fine-tune diffusion models for better alignment.

\paragraph{Regression-based Methods.}
Regression-based evaluators map images (and their prompts) to continuous quality scores by learning on mean opinion ratings, offering interpretable, absolute assessments. IP-IQA~\cite{qu2024bringing} enhances CLIP with an Image2Prompt pretraining task and cross‐attention fusion, injecting both image and prompt into a special token; it achieves state-of-the-art absolute scores on AGIQA‐1K/3K benchmarks. MPS~\cite{zhang2024learning} decomposes human preference into four axes—\textit{aesthetics}, \textit{semantic alignment}, \textit{detail quality}, and \textit{overall}—and trains specialized regressors on 918 K comparisons, enabling multi-dimensional absolute scoring that closely matches user judgments.

\begin{table}[htbp]
    \centering
    \resizebox{0.8\textwidth}{!}{
    \begin{tabular}{ccccc}
        \toprule
        \textbf{Method} & \textbf{Task} & \textbf{Category} & \textbf{Reference} & \textbf{Protocol} \\
        \midrule
        LPIPS~\cite{zhang2018unreasonable} & Image Generation & Embedding & Yes & Single \\
        C2ST~\cite{lopez2016revisiting} & Text-to-Image & Classification & No & Pair \\
        CAS~\cite{ravuri2019classification} & Text-to-Image & Classification & No & Single \\
        DreamSim~\cite{fu2023dreamsim} & Text-to-Image & Embedding & Yes & Single \\
        ImageReward~\cite{xu2024imagereward} & Text-to-Image & Reward & No & Single \& Pair \\
        PickScore~\cite{kirstain2023pick} & Text-to-Image & Embedding & No & Single \& Pair \\
        HPS-v2~\cite{wu2023human} & Text-to-Image & Embedding & No & Pair \\
        MPS~\cite{zhang2024learning} & Text-to-Image & Regression & No & Single \\
        IP-IQA~\cite{qu2024bringing} & Text-to-Image & Regression & No & Single \\
        \bottomrule
    \end{tabular}
    }
    \caption{Categorization of learning-based evaluation methods for visual generation tasks. Methods are classified by technical approach (Category), need for reference inputs (Reference), and evaluation protocol (Protocol). Hybrid protocols support both absolute scoring and pairwise comparisons.}
    \label{tab:learning_based_evaluation_vision}
\end{table}

\subsection{LLM/MLLM-based Evaluation}
Leveraging the reasoning and multimodal capabilities of recent large language models (LLMs) and multimodal LLMs (MLLMs), we propose a unified taxonomy that captures five core evaluation paradigms for visual generation tasks. As summarized in Table~\ref{tab:llm_based_evaluation_vision}, our taxonomy distinguishes methods by their mechanism, interpretability, and protocol, encompassing Direct Scoring, QA Frameworks, Explainable/Hybrid Metrics, Preference Modeling, and Temporal Consistency assessments.

\paragraph{Direct Scoring Methods} These methods harness instruction-following LLMs to compute alignment scores in a single pass. Early work in this space framed alignment as a semantic similarity task, with LLMScore leveraging textual descriptions to estimate how well a generated image matches its prompt~\cite{lu2024llmscore}, and VQAScore interpreting probabilistic QA outputs for a compositional score~\cite{lin2025evaluating}. While both methods are reference-free and lightweight, VQAScore’s black-box probability aggregation sacrifices interpretability in favor of efficiency, whereas LLMScore’s structured prompting yields more transparent reasoning.

\paragraph{QA Frameworks} QA-based methods decompose evaluation into discrete question–answer interactions that probe different facets of image fidelity. Multimodal QA methods such as QAS apply open-ended visual question answering to verify both alignment and perceptual details~\cite{sun2024journeydb}, and Diagnostic QA techniques like DSG generate structured scene-graph questions to ensure exhaustive semantic coverage~\cite{cho2023davidsonian}. Hierarchical QA, exemplified by TIFA v1.0, further refines this approach by organizing questions into levels (object, counting, relational) to pinpoint specific failure modes~\cite{hu2023tifa}. Rubric-based QA, embodied by PhyBench, introduces detailed criteria for physical commonsense reasoning—addressing a gap in earlier QA methods by explicitly assessing physics-based plausibility~\cite{meng2024phybench}. Transitioning from generic QA to these specialized rubrics resolves the coarse-grained limitations of earlier frameworks and better quantifies reasoning depth.

\paragraph{Explainable and Hybrid Methods} These methods leverage MLLMs to generate natural-language explanations and combine reference-free and reference-based cues. X-IQE produces chain-of-thought rationales alongside scores for alignment, aesthetics, and authenticity~\cite{chen2023x}, while VIEScore uses visual instruction tuning to fuse alignment and perceptual judgments within a unified score~\cite{ku2023viescore}. The latest T2I-Eval toolkit enhances this fusion by dynamically selecting between reference-based comparison and prompt-based QA, improving robustness across diverse generation styles~\cite{tu2024automaticevaluationtexttoimagegeneration}.

\paragraph{Preference Modeling} Preference-based methods reframes quality assessment as pairwise ranking. Gecko employs LLM-driven preference judgments to order candidate outputs by semantic fidelity, mitigating absolute scoring biases inherent to single-instance methods and enabling more nuanced comparisons between high-quality images~\cite{wiles2024revisiting}.

\paragraph{Temporal} Temporal methods extend alignment evaluation to video generation by verifying assertions over time. The TCR metric checks transition completion across frames to ensure narrative coherence, and TC-Score aggregates frame-level checks into a corpus-level measure, providing scalable assessment for dynamic content~\cite{feng2024tc}.

\begin{table}[htbp]
    \centering
    \resizebox{\textwidth}{!}{
    \begin{tabular}{ccccccc}
        \toprule
        \textbf{Method} & \textbf{Task} & \textbf{Interpretability} & \textbf{Category} & \textbf{Dimension} & \textbf{Reference} & \textbf{Protocol} \\
        \midrule
        LLMScore~\cite{lu2024llmscore} & Text-to-Image & High & Direct Scoring & Alignment & No & Single \\
        VQAScore~\cite{lin2025evaluating} & Text-to-Image & Low & Direct Scoring & Alignment & No & Single \\
        QAS~\cite{sun2024journeydb} & Text-to-Image & Medium & Multimodal QA & Alignment \& Perceptual & No & Single \\
        X-IQE~\cite{chen2023x} & Text-to-Image & High & Hybrid Interpretable Metric & Alignment \& Perceptual & No & Single \\
        DSG~\cite{cho2023davidsonian} & Text-to-Image & High & Diagnostic QA & Alignment & No & Single \\
        TIFA v1.0~\cite{hu2023tifa} & Text-to-Image & High & Hierarchical QA & Alignment & No & Single \\
        PhyBench~\cite{meng2024phybench} & Text-to-Image & High & Rubric-based QA & Alignment \& Commonsense & No & Single \\
        VIEScore~\cite{ku2023viescore} & Text-to-Image & Medium & Hybrid Interpretable Metric & Alignment \& Perceptual & No & Single \\
        Gecko~\cite{wiles2024revisiting} & Text-to-Image & Medium & Preference Modeling & Alignment & No & Pair \\
        T2I-Eval~\cite{tu2024automaticevaluationtexttoimagegeneration} & Text-to-Image & High & Hybrid QA & Alignment \& Perceptual & Both & Single \\
        TCR~\cite{feng2024tc} & Video Gen & High & Temporal Verification & Alignment & No & Corpus \\
        TC-Score~\cite{feng2024tc} & Video Gen & High & Frame-level Checking & Alignment & No & Single \\
        \bottomrule
    \end{tabular}
    }
    \caption{Taxonomy of LLM/MLLM-based evaluation methods categorized by assessment protocol (single-instance, pairwise, corpus-level) and primary evaluation dimensions. The reference requirement indicates whether ground truth comparisons are needed.}
    \label{tab:llm_based_evaluation_vision}
\end{table}

Notable methodological innovations include T2I-Eval's~\cite{tu2024automaticevaluationtexttoimagegeneration} hybrid approach combining reference-based and reference-free assessment through dynamic prompt selection, and TCR's~\cite{feng2024tc} novel application of temporal assertion verification for video generation. However, current limitations persist in evaluating open-domain creative generation and quantifying subtle perceptual qualities, suggesting future directions for MLLM-based evaluation research.

\subsection{Benchmark-based Evaluation}
Benchmark-based evaluation methodologies employ systematically constructed datasets with human annotations or synthetic criteria to establish standardized assessment frameworks. These benchmarks enable reproducible and quantifiable comparisons across generative models through three principal dimensions: (1) \textit{Reference requirement} (reference-based vs. reference-free), (2) \textit{Evaluation paradigm} (automatic metrics vs. human judgments), and (3) \textit{Evaluation criteria} (task-specific capabilities vs. cross-modal alignment). As shown in Table~\ref{tab:benchmark_based_evaluation_vision}, contemporary benchmarks exhibit three evolutionary trends: increasing specialization in task-specific evaluation (e.g., temporal consistency in video generation), integration of multimodal large language models (MLLMs) for semantic alignment assessment, and hybrid approaches combining automatic metrics with human preference modeling.

\textbf{Reference-based methods} require ground-truth samples for comparative evaluation. For instance, \textit{VTAB+}~\cite{schuhmann2022laion} establishes multi-task benchmarks across 35 vision tasks using accuracy metrics, while \textit{C-CUB \& C-Flowers}~\cite{park2021benchmark} focus on compositional image generation through fine-grained attribute matching. These approaches excel in controlled comparisons but face scalability challenges in open-ended generation tasks.

\textbf{Reference-free methods} leverage model-driven assessment without requiring target outputs. The \textit{PhyBench}~\cite{meng2024phybench} benchmark evaluates physical commonsense in generated images through LLM-based reasoning chains, whereas \textit{T$^3$Bench}~\cite{he2023t} combines neural scorers with LLM evaluators for 3D asset alignment. Recent advances like \textit{M$^2$RAG}~\cite{ma2025multimodalretrievalaugmentedmultimodal} demonstrate the potential of retrieval-augmented evaluation for multimodal coherence assessment.

\textbf{Specialized assessment taxonomies} emerge across modalities:
\begin{itemize}
    \item \textit{Image generation}: Hierarchical evaluation in \textit{MJ-Bench}~\cite{chen2024mj} spans safety checks (NSFW detection), semantic alignment (CLIPScore), and perceptual quality (FID)
    \item \textit{Video generation}: \textit{TC-Bench}~\cite{feng2024tc} introduces temporal compositionality metrics using LLM-based trajectory analysis
    \item \textit{3D generation}: \textit{MATE-3D}~\cite{zhang2024benchmarking} combines geometric consistency metrics with commonsense reasoning evaluations
\end{itemize}

\textbf{Emerging hybrid approaches} blend multiple evaluation paradigms. \textit{Pick-a-Pic}~\cite{kirstain2023pick} combines human preference modeling with pairwise comparison metrics, while \textit{LenCom-EVAL}~\cite{lakhanpal2024refining} integrates OCR verification with neural semantic scores for complex text rendering. The field is gradually shifting from single-metric evaluation towards multidimensional assessment frameworks that address both low-level perceptual quality and high-level semantic fidelity.

\begin{table}[htbp]
    \centering
    \resizebox{\textwidth}{!}{
    \begin{tabular}{cccc}
        \toprule
        \textbf{Benchmark} & \textbf{Task} & \textbf{Evaluation Criteria} & \textbf{Metrics} \\
        \midrule
        Pick-a-Pic~\cite{kirstain2023pick} & Text-to-Image & Human preference modeling & Pairwise comparison \\
        JourneyDB~\cite{sun2024journeydb} & Text-to-Image & Prompt-image comprehension & Heuristic scoring \\
        TC-Bench~\cite{feng2024tc} & Text/Image-to-Video & Temporal consistency & LLM-based trajectory analysis \\
        PhyBench~\cite{meng2024phybench} & Text-to-Image & Physical commonsense & LLM reasoning chains \\
        VTAB+~\cite{schuhmann2022laion} & Multi-task & Cross-task generalization & Accuracy \\
        T2I-CompBench++~\cite{huang2023t2i} & Text-to-Image & Compositional alignment & Attribute matching \\
        T2V-CompBench~\cite{sun2024t2v} & Text-to-Video & Cross-frame coherence & MLLM/Detection/Tracking \\
        MJ-Bench~\cite{chen2024mj} & Text-to-Image & Safety \& Quality & NSFW detection, CLIPScore \\
        C-CUB/Flowers~\cite{park2021benchmark} & Text-to-Image & Fine-grained alignment & CLIPScore, Human eval \\
        ImagenHub~\cite{ku2023imagenhub} & Image Edit & Semantic preservation & Human evaluation \\
        VR~\cite{tewel2022zerocap} & Image-to-Text & Visual relation capture & Heuristic scoring \\
        EditBench~\cite{wang2023imagen} & Image Inpainting & Context consistency & Human eval \\
        LenCom-EVAL~\cite{lakhanpal2024refining} & Text-to-Image & Complex text rendering & CLIPScore, OCR, NLD \\
        MATE-3D~\cite{zhang2024benchmarking} & Text-to-3D & Geometric consistency & Point cloud analysis \\
        EvalCrafter~\cite{liu2024evalcrafter} & Text-to-Video & Multi-aspect quality & Dover, IS, CLIPScore \\
        FETV~\cite{liu2024fetv} & Text-to-Video & Temporal alignment & BLIPScore, CLIPScore \\
        HRS-Bench~\cite{bakr2023hrs} & Text-to-Image & Human resemblance & Face detection metrics \\
        GenAI-Bench~\cite{li2024genai} & Multi-modal & Cross-modal alignment & VQAScore \\
        T$^3$Bench~\cite{he2023t} & Text-to-3D & Semantic fidelity & LLM-based scoring \\
        T2I-Eval-Bench~\cite{tu2024automaticevaluationtexttoimagegeneration} & Text-to-Image & Perceptual \& Alignment & MLLM-based \\
        ConceptMix~\cite{wu2024conceptmix} & Text-to-Image & Concept integration & LLM-based QA \\
        M$^2$RAG~\cite{ma2025multimodalretrievalaugmentedmultimodal} & Multi-modal & Contextual coherence & Retrieval accuracy \\
        \bottomrule
    \end{tabular}
    }
    \caption{Taxonomy of benchmark-based evaluation methods for visual generation tasks, categorized by modality, evaluation focus, and metric paradigm. The table highlights three key dimensions: reference requirement (ground-truth dependency), evaluation paradigm (automatic vs. human), and assessment focus (modality-specific capabilities vs. cross-modal alignment).}
    \label{tab:benchmark_based_evaluation_vision}
\end{table}

\subsection{Meta-Evaluation Benchmarks for Vision Generation\label{subsec:vision_meta_eval}}

\begin{table}[htbp]
    \centering
    \resizebox{\textwidth}{!}{
    \begin{tabular}{ccccc}
        \toprule
            \textbf{Benchmark} & \textbf{Task} & \textbf{Type} & \textbf{Focus} & \textbf{Protocol} \\
        \midrule
            TIFA v1.0~\cite{hu2023tifa} & Text-to-Image & Objective & Alignment & Single \\
            DSG-1k~\cite{cho2023davidsonian} & Text-to-Image & Objective & Alignment & Single \& Corpus \\
            GenAI-Bench~\cite{li2024genai} & Text-to-Image/Video & Objective & Overall Quality & Pair \\
            T2I-Eval~\cite{tu2024automaticevaluationtexttoimagegeneration} & Text-to-Image & Objective \& Subjective & Perceptual \& Alignment & Single \\
        \bottomrule
    \end{tabular}
    }
    \caption{Taxonomy of meta-evaluation benchmarks for visual generation tasks, categorized by modality, evaluation focus, and metric paradigm. The table highlights three key dimensions: reference requirement (ground-truth dependency), evaluation paradigm (automatic vs. human), and assessment focus (modality-specific capabilities vs. cross-modal alignment).}
    \label{tab:meta_eval_vision}
\end{table}

As shown in Table~\ref{tab:meta_eval_vision}, there are currently only a handful of publicly available benchmarks designed to assess how well automatic metrics correlate with human judgments in text‑to‑vision generation tasks. These benchmarks differ along three principal dimensions:
(1) \emph{Task Modality:} All four focus primarily on text‑to‑image synthesis, but GenAI‑Bench further extends to video generation evaluations.
(2) \emph{Evaluation Type:} While TIFA v1.0 and DSG‑1k employ purely objective, reference‑based measures, T2I‑Eval incorporates both objective and human (subjective) ratings to capture perceptual quality.
(3) \emph{Protocol:} TIFA v1.0 and T2I‑Eval follow a single‑instance protocol (each candidate is judged independently), whereas DSG‑1k and GenAI‑Bench additionally support corpus‑level aggregation, enabling evaluation of metric consistency across a batch of examples.

Due to the early stage of research on meta‑evaluation for visual generative models, the existing benchmarks provide only a limited view of current metric reliability.  For instance, TIFA v1.0 was built on a relatively small set of high‑quality image captions, emphasizing calibration of CLIP‑based scores; in contrast, DSG‑1k offers a larger, more diverse dataset but remains purely automatic.  GenAI‑Bench’s inclusion of video data marks an important step toward multimodal assessment, yet it still relies on standard objective protocols. Finally, T2I‑Eval’s hybrid design demonstrates the value of supplementing reference‑based scores with human perceptual judgments, but its scope is restricted to single images.

Based on the observations of the current research progress, we can conclude some key insights and future directions: 
\begin{itemize}
    \item \textbf{Incorporate richer subjective evaluations.} As T2I‑Eval shows, objective scores alone can miss nuances of visual coherence and creativity; expanding large‑scale human annotations—ideally via scalable, low‑cost methods such as crowdsourced pairwise comparisons—would strengthen benchmark validity. 
    \item \textbf{Broaden protocol diversity.} Corpus‑level protocols capture inter‑image consistency and robustness to dataset biases; future benchmarks should systematically compare single‑ versus corpus‑level correlations across a wider variety of content genres (e.g., cartoons, medical imagery).  
    \item \textbf{Extend modality coverage.} With video, 3D, and interactive graphics becoming mainstream, meta‑evaluation frameworks must evolve to handle temporal dynamics and spatial complexity, perhaps by integrating spatiotemporal alignment scores and human motion judgments.  
    \item \textbf{Foster open, modular benchmark suites.}  A community‑driven benchmark that unifies objective libraries (e.g., CLIP, ViT‑based scores) with plug‑and‑play human‑in‑the‑loop modules would accelerate progress by allowing researchers to evaluate new metrics under consistent, extensible protocols.
\end{itemize}

By addressing these gaps—richer human signals, diverse protocols, expanded modalities, and modular design—future meta‑evaluation benchmarks can more comprehensively characterize the true strengths and weaknesses of text‑to‑vision metrics, ultimately guiding the development of more reliable and human‑aligned evaluation methods.

\subsection{Comparison of Current Methods and Promising Future Directions\label{subsec:challenges_future_visual}}

\begin{table*}[htbp]
    \small
    \centering
    \caption{\label{tab:vision_meta_eval} Comparison of evaluation methods on text-to-image generative task.}
    \vspace{-0.1cm}
    \resizebox{0.75\linewidth}{!}{
        \begin{tabular*}{0.71\linewidth}{cl|cc|cc}
        \toprule
             \multirow{2}{*}[-2.0pt]{\textbf{Category}} & \multirow{2}{*}[-2.0pt]{\textbf{Method}} & \multicolumn{2}{c|}{\textbf{T2I-Eval}} & \multicolumn{2}{c}{\textbf{Tifa v1.0}} \\
        \cmidrule{3-6}
            & & $\rho$ & $\tau$ & $\rho$ & $\tau$ \\
        \midrule
            \multirow{3}{*}[-1.5pt]{\textbf{Embedding-based}} & FID~\cite{heusel2017gans} & -0.1231 & -0.0862 & - & - \\
            & CLIPScore~\cite{hessel2021clipscore} & 0.1505 & 0.1016 & 0.3382 & 0.2456 \\
            & BLIPv2Score~\cite{li2023blip2bootstrappinglanguageimagepretraining} & 0.2152 & 0.1423 & 0.4049 & 0.2944 \\
        \midrule
            \multirow{4}{*}[-1.5pt]{\textbf{Learning-based}} & LPIPS~\cite{zhang2018unreasonable} & -0.1244 & -0.0856 & - & - \\
            & DreamSim~\cite{fu2023dreamsim} & -0.1382 & -0.0968 & - & - \\
            & PickScore~\cite{kirstain2023pick} & 0.3944 & 0.2803 & 0.4279 & 0.3137 \\
            & ImageReward~\cite{xu2024imagereward} & 0.4046 & 0.2839 & 0.6211 & 0.4659 \\
        \midrule
            \multirow{7}{*}[-1.5pt]{\textbf{\begin{tabular}[c]{@{}c@{}}LLM-based \&\\MLLM-based\end{tabular}}} & LLMScore$_\textbf{GPT-4}$~\cite{lu2024llmscore} & 0.3096 & 0.2228 & 0.4969 & 0.3753 \\
            & TIFA$_\textbf{mPLUG}$~\cite{hu2023tifa} & 0.3252 & 0.2455 & 0.5922 & 0.4717 \\
            & DSG$_\textbf{Dependent}$~\cite{cho2023davidsonian} & 0.4582 & 0.3512 & 0.6046 & 0.4893 \\
            & DSG$_\textbf{Independent}$ & 0.4704 & 0.3655 & 0.6108 & 0.4954 \\
            & VQAScore$_\textbf{CLIP-FlanT5}$~\cite{lin2025evaluating} & 0.5116 & 0.3712 & 0.6950 & 0.5321 \\
            & VIEScore$_\textbf{GPT-4o}$~\cite{ku2023viescore} & 0.5545 & 0.4170 & 0.5388 & 0.4065 \\
            & T2I-Eval$_\textbf{MiniCPM-V-2.6}$~\cite{tu2024automatic} & 0.5802 & \textbf{0.4409} & 0.6061 & 0.4692 \\
            & T2I-Eval-R1~\cite{ma2025t2ievalr1reinforcementlearningdrivenreasoning} & \textbf{0.5874} & 0.4380 & \textbf{0.7043} & \textbf{0.5510} \\
        \bottomrule
        \end{tabular*} 
    }
    \vspace{-0.3cm}
\end{table*}

A quantitative comparison on two human-annotated text-to-image benchmarks (T2I-Eval and TIFA v1.0) reveals three clear trends (Table ~\ref{tab:vision_meta_eval}):

\textbf{Embedding-based methods} steadily improve as backbone models advance—from Inception-v3 (FID) to CLIP (CLIPScore) and then to BLIP-v2 (BLIPv2Score)—but their correlations with human judgments remain modest (e.g., BLIPv2Score achieves $\rho$=0.4049 on T2I-Eval and $\rho$=0.2944 on TIFA v1.0) .

\textbf{Learning-based metrics} that train directly on human preferences, such as PickScore and ImageReward, offer stronger alignment (ImageReward achieves $\rho$=0.6211 on T2I-Eval, $\rho$=0.4659 on TIFA v1.0). These models capture nuanced perceptual and semantic judgments beyond static embeddings.

\textbf{LLM/MLLM-based evaluators} deliver the highest correlations. For instance, T2I-Eval with MiniCPM-V-2.6 and VQAScore with CLIP-FlanT5 outperform all other paradigms. Their ability to decompose visual content into rich, instructive reasoning appears critical for human-like assessment.

Despite these advances, vision evaluation remains nascent. Current benchmarks are limited in scale, modality coverage, and evaluation protocols. To drive progress, we identify four promising directions:

\begin{itemize}
    \item \textbf{Expand and diversify meta-evaluation datasets:}  Beyond static image pairs, future benchmarks should include video, 3D assets, and interactive graphics, annotated for temporal consistency, geometry, and user experience. Larger, crowd-sourced collections will improve statistical reliability and cover rare or creative failure modes.
    \item \textbf{Develop multidimensional, hybrid metrics:}  Single-score outputs obscure trade-offs between fidelity, diversity, and aesthetics. Combining distributional measures (e.g., PRD), embedding alignment, and LLM-derived critiques in a modular framework would allow researchers to tailor evaluation to specific application needs.
    \item \textbf{Integrate human–model collaboration:}  Leveraging LLMs to propose candidate critiques which are then refined or validated by human annotators can yield high-quality labels at lower cost. This pipeline could bootstrap learning-based evaluators and continuously update benchmarks as models evolve.
    \item \textbf{Standardize extensible, open evaluation platforms:}  A community-driven library of plug-and-play modules—for reference-based metrics, reference-free scorers, LLM/MLLM evaluators, and human-in-the-loop interfaces—would enable apples-to-apples comparisons and rapid testing of new methods under consistent protocols.
\end{itemize}

By addressing these gaps—richer annotations, hybrid evaluation strategies, human–AI collaboration, and open tooling—we can move toward more robust, generalizable, and human-aligned evaluation of vision generation.

\section{Automatic Evaluation for Audio and Speech-related Generation} 
\label{sec:speech_evaluation}

\tikzstyle{my-box}=[
    rectangle,
    draw=hidden-draw,
    rounded corners,
    text opacity=1,
    minimum height=1.5em,
    minimum width=5em,
    inner sep=2pt,
    align=center,
    fill opacity=.5,
]
\tikzstyle{leaf}=[my-box, minimum height=1.5em,
    fill=hidden-orange!60, text=black, align=left,font=\scriptsize,
    inner xsep=2pt,
    inner ysep=4pt,
]
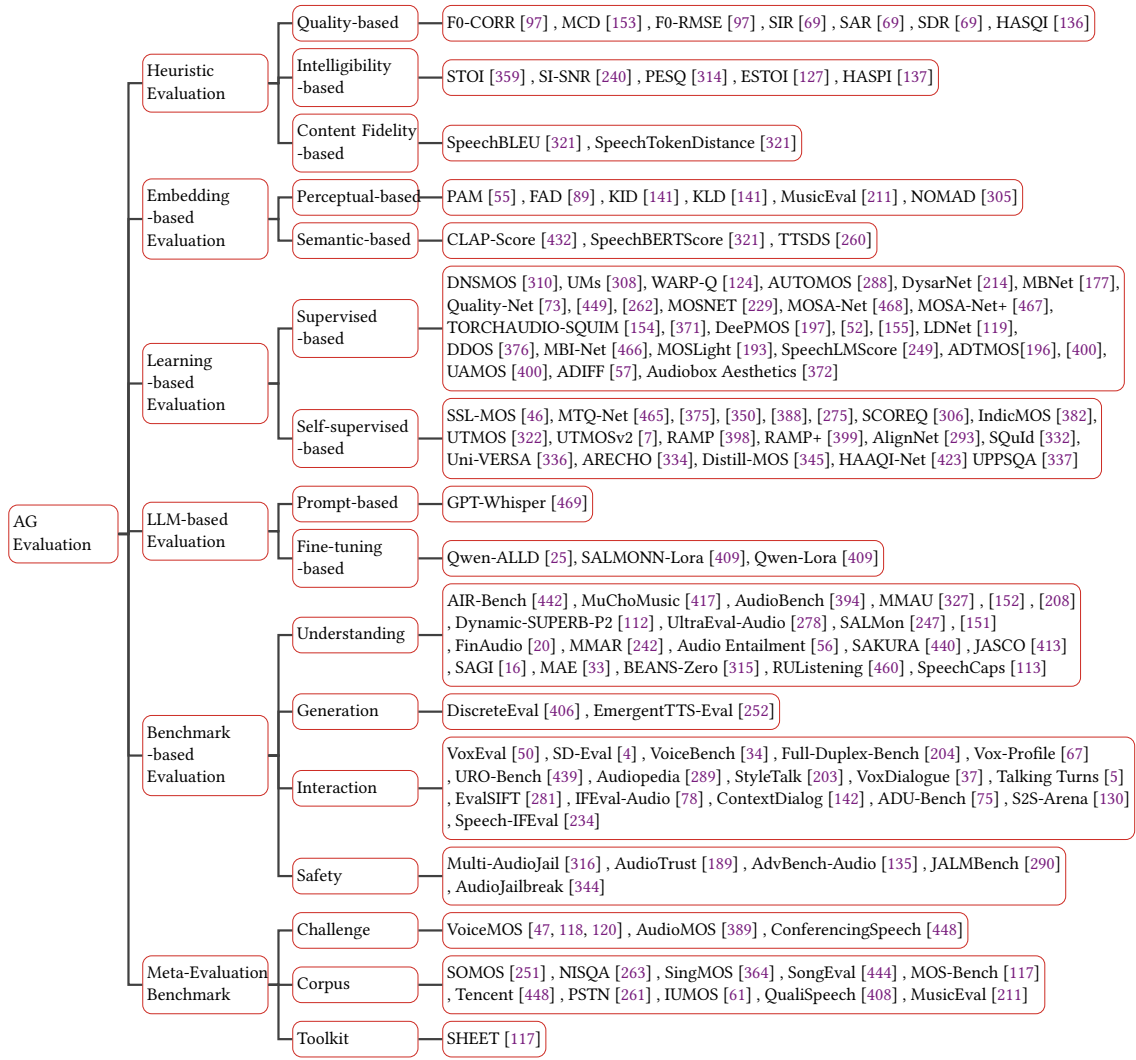
\begin{figure*}[t]
    \centering
    \resizebox{\textwidth}{!}{
        \begin{forest}
            forked edges,
            for tree={
                grow=east,
                reversed=true,
                anchor=base west,
                parent anchor=east,
                child anchor=west,
                base=left,
                font=\small,
                rectangle,
                draw=hidden-draw,
                rounded corners,
                align=left,
                minimum width=3em,
                edge+={darkgray, line width=1pt},
                s sep=3pt,
                inner xsep=2pt,
                inner ysep=3pt,
                ver/.style={rotate=90, child anchor=north, parent anchor=south, anchor=center},
            },
            where level=0{text width=5em,font=\small,}{},
            where level=1{text width=5.8em,font=\small,}{},
            where level=2{text width=5.8em,font=\small,}{},
            [AG\\ Evaluation
                [
                    Heuristic\\ Evaluation
                    [Quality-based\\
                        [
                            F0-CORR \cite{hayashi2021espnet2}
                            {,} MCD \cite{kubichek1993mel}
                            {,} F0-RMSE \cite{hayashi2021espnet2}
                            {,} SIR \cite{fevotte2005bss_eval}
                            {,} SAR \cite{fevotte2005bss_eval}
                            {,} SDR \cite{fevotte2005bss_eval}
                            {,} HASQI \cite{kates2010hearing} 
                        ]
                    ]
                    [Intelligibility\\-based
                        [
                            STOI \cite{taal2011algorithm}
                            {,} SI-SNR \cite{luo2018tasnet}
                            {,} PESQ \cite{rix2001perceptual}
                            {,} ESTOI \cite{jensen2016algorithm}
                            {,} HASPI \cite{kates2014hearing}
                        ]
                    ]
                    [Content Fidelity\\-based 
                        [
                            SpeechBLEU \cite{saeki2024speechbertscore}
                            {,} SpeechTokenDistance \cite{saeki2024speechbertscore}
                        ]
                    ]
                ]
                [Embedding\\-based\\ Evaluation
                    [Perceptual-based\\
                        [
                            PAM \cite{deshmukh2024pam}
                            {,} FAD \cite{gui2024adapting}
                            {,} KID \cite{kilgour2018fr}
                            {,} KLD \cite{kilgour2018fr}
                            {,} MusicEval \cite{liu2025musiceval}
                            {,} NOMAD \cite{ragano2024nomad}
                        ]
                    ]
                    [Semantic-based\\
                        [
                            CLAP-Score \cite{xiao2024reference}
                            {,} SpeechBERTScore \cite{saeki2024speechbertscore}
                            {,} TTSDS \cite{minixhofer2024ttsds}
                        ]
                    ]
                ]
                [Learning\\-based\\ Evaluation
                    [Supervised\\-based
                        [
                            DNSMOS \cite{reddy2021dnsmos}{,}
                            UMs \cite{ravuri2024uncertainty}{,}
                            WARP-Q \cite{jassim2021warp}{,}
                            AUTOMOS \cite{patton2016automos}{,}
                            DysarNet \cite{liu2024non}{,}
                            MBNet \cite{leng2021mbnet}{,} \\
                            Quality-Net \cite{fu2018quality}{,}
                            \cite{yoshimura2016hierarchical}{,}
                            \cite{mittag2021deep}{,}
                            MOSNET \cite{lo2019mosnet}{,}
                            MOSA-Net \cite{MOSA-Net}{,}
                            MOSA-Net+ \cite{MOSA-Net+}{,} \\
                            TORCHAUDIO-SQUIM \cite{kumar2023torchaudio}{,}
                            \cite{tian2022multi}{,}
                            DeePMOS \cite{liang2023deepmos}{,}
                            \cite{cumlin2025impairments}{,}
                            \cite{kunevsova2023ensemble}{,}
                            LDNet \cite{huang2022ldnet}{,} \\
                            DDOS \cite{tseng2022ddos}{,}
                            MBI-Net \cite{zezario2022mbi}{,}
                            MOSLight \cite{li2023moslight}{,}
                            SpeechLMScore \cite{maiti2023speechlmscore}{,}
                            ADTMOS\cite{liang2025adtmos}{,}
                            \cite{wang2024uncertainty}{,} \\
                            UAMOS \cite{wang2024uncertainty}{,}
                            ADIFF \cite{deshmukh2025adiff}{,}
                            Audiobox Aesthetics \cite{tjandra2025meta}
                        ]
                    ]
                    [
                        Self-supervised\\-based
                        [
                            SSL-MOS \cite{ssl-mos}{,}
                            MTQ-Net \cite{mtqnet}{,}
                            \cite{ssr-mos}{,}
                            \cite{sultana2025pre}{,}
                            \cite{vioni2023investigating}{,}
                            \cite{nguyen2024exploring}{,}
                            SCOREQ \cite{ragano2024scoreq}{,}
                            IndicMOS \cite{udupa2024indicmos}{,} \\
                            UTMOS \cite{saeki2022utmos}{,}
                            UTMOSv2 \cite{baba2024t05}{,}
                            RAMP \cite{wang2023ramp}{,}
                            RAMP+ \cite{wang2025ramp+}{,}
                            AlignNet \cite{pieper2024alignnet}{,}
                            SQuId \cite{sellam2023squid}{,} \\
                            Uni-VERSA \cite{shi2025uni}{,}
                            ARECHO \cite{shi2025arecho}{,}
                            Distill-MOS \cite{stahl2025distillation}{,}
                            HAAQI-Net \cite{wisnu2025haaqi}
                            UPPSQA \cite{UPPSQA}
                        ]
                    ]
                ]
                [
                    LLM-based\\ Evaluation
                    [
                        Prompt-based
                        [
                            GPT-Whisper \cite{GPT-Whisper}
                        ]
                    ]
                    [
                        Fine-tuning\\-based
                        [
                            Qwen-ALLD \cite{alld}{,}
                            SALMONN-Lora \cite{lora}{,}
                            Qwen-Lora \cite{lora}
                        ]
                    ]
                ]
                [
                    Benchmark\\-based\\ Evaluation \\
                    [
                        Understanding\\
                        [
                            AIR-Bench \cite{yang2024airbench}
                            {,} MuChoMusic \cite{weck2024muchomusic}
                            {,} AudioBench \cite{wang2024audiobench}
                            {,} MMAU~\cite{sakshi2024mmau} 
                            {,} \cite{kuan2025can}
                            {,} \cite{gpt4o} \\
                            {,} Dynamic-SUPERB-P2 \cite{huang2024dynamic} 
                            {,} UltraEval-Audio \cite{ultraeval-audio2025}
                            {,} SALMon \cite{maimon2025salmon}
                            {,} \cite{kuan2024understanding} \\
                            {,} FinAudio \cite{cao2025finaudio} 
                            {,} MMAR \cite{ma2025mmar}
                            {,} Audio Entailment \cite{deshmukh2025audio}
                            {,} SAKURA \cite{yang2025sakura}
                            {,} JASCO \cite{wang2025they} \\
                            {,} SAGI \cite{bu2024roadmap} 
                            {,} MAE \cite{chen2024beyond} 
                            {,} BEANS-Zero \cite{robinson2024naturelm}
                            {,} RUListening \cite{zang2025you}
                            {,} SpeechCaps \cite{huang2025speechcaps} 
                        ]
                    ]
                    [
                        Generation\\
                        [
                            DiscreteEval \cite{wang2024evaluating}
                            {,} EmergentTTS-Eval \cite{manku2025emergenttts}
                        ]
                    ]
                    [
                        Interaction\\
                        [
                            VoxEval \cite{cui2025voxeval}
                            {,} SD-Eval \cite{ao2024sd-eval}
                            {,} VoiceBench \cite{chen2024voicebench}
                            {,} Full-Duplex-Bench \cite{Full-Duplex-Bench}
                            {,} Vox-Profile \cite{feng2025vox} \\
                            {,} URO-Bench \cite{yan2025uro}
                            {,} Audiopedia \cite{penamakuri2025audiopedia}
                            {,} StyleTalk \cite{lin2024advancing} 
                            {,} VoxDialogue \cite{cheng2025voxdialogue} 
                            {,} Talking Turns \cite{arora2025talking} \\
                            {,} EvalSIFT \cite{pandey2025sift}
                            {,} IFEval-Audio \cite{gao2025ifeval}
                            {,} ContextDialog \cite{kim2025does}
                            {,} ADU-Bench \cite{gao2024benchmarking}
                            {,} S2S-Arena \cite{jiang2025s2s} \\
                            {,} Speech-IFEval \cite{Speech-ifeval}
                        ]
                    ]
                [
                    Safety\\
                    [
                        Multi-AudioJail
                        \cite{roh2025multilingual}
                        {,} AudioTrust \cite{li2025audiotrust}
                        {,} AdvBench-Audio \cite{kang2024advwave}
                        {,} JALMBench \cite{peng2025jalmbench} \\
                        {,} AudioJailbreak \cite{song2025audio}
                    ]
                ]
            ]
                [
                    Meta-Evaluation\\Benchmark \\
                    [
                        Challenge \\
                        [
                            VoiceMOS \cite{huang2022voicemos,cooper2023voicemos,huang2024voicemos}
                            {,} AudioMOS \cite{audiomos2025}
                            {,} ConferencingSpeech \cite{yi2022conferencingspeech}
                        ]
                    ]
                    [
                        Corpus\\
                        [
                            SOMOS \cite{maniati2022somos}
                            {,} NISQA \cite{mittag2021nisqa}
                            {,} SingMOS \cite{tang2024singmos}
                            {,} SongEval \cite{yao2025songeval}
                            {,} MOS-Bench \cite{huang2024mosbench} \\
                            {,} Tencent \cite{yi2022conferencingspeech}
                            {,} PSTN \cite{mittag2020dnn}
                            {,} IUMOS \cite{dong2020pyramid}
                            {,} QualiSpeech \cite{wang2025qualispeech}
                            {,} MusicEval \cite{liu2025musiceval} 
                        ]
                    ]
                    [
                        Toolkit\\
                        [
                            SHEET \cite{huang2024mosbench}
                        ] 
                    ]
                ]
        ]
        \end{forest}
    }
    \caption{Taxonomy of research of automatic evaluation metrics in Audio and Speech Generation (AG). The detailed classifications about automatic evaluation methods could be found in Section~\ref{sec:speech_evaluation}.}
    \label{img:speech_taxonomy}
\end{figure*}

In this section, we review the development of automatic evaluation methodologies for audio and speech generation tasks, including text-to-speech (TTS), voice conversion (VC), text-to-audio generation (TTA), and generative audio-language models (ALMs). As the capabilities of generation systems in these areas have grown, evaluation techniques have evolved accordingly. The progression of these methodologies can be categorized into five distinct paradigms, as illustrated in Figure~\ref{img:speech_taxonomy}: \textbf{(1) Heuristic Evaluation.} Early approaches are grounded in simple rule-based metrics, such as signal-to-noise ratio, log-spectral distance, and prosody statistics, which are used to quantify distortion or fidelity. While computationally lightweight, these methods are often poorly aligned with perceptual human judgment, particularly in complex prosodic or speaker-sensitive settings. \textbf{(2) Embedding-based Evaluation. }With the rise of self-supervised learning, embedding-based methods leverage learned audio representations from models such as Wav2vec\cite{schneider2019wav2vec,baevski2020wav2vec}, HuBERT\cite{hsu2021hubert}, and VGGish \cite{hershey2017cnn} to measure acoustic similarity, speaker identity, and temporal dynamics. These embeddings capture fine-grained characteristics of audio signals and enable more perceptually relevant comparisons than handcrafted metrics.\textbf{ (3) Learning-based Evaluation.} This category of methods relies on supervised or self-supervised models trained on audio modality data such as speech, general audio, or music, along with either human-annotated or automatically generated quality scores. From the perspective of speech generation, these models are designed to approximate general-purpose and expert-level subjective evaluation metrics, including the Mean Opinion Score (MOS). By learning to predict these scores directly, the models aim to estimate key perceptual attributes such as naturalness and intelligibility. Through learning from annotated data that reflects human preferences, these models typically show improved alignment with subjective judgments and demonstrate robust generalization across tasks. In the broader context of audio generation, this paradigm often employs neural networks trained on datasets that contain audio samples paired with quality scores. These annotations may be obtained through manual labeling or enriched using large language models. The objective is to enable automatic evaluation along multiple perceptual dimensions such as difference, aesthetic quality, and content consistency. For example, methods like ADIFF\cite{deshmukh2025adiff} incorporate label enhancements generated by large language models to increase the diversity and richness of training data. Although these techniques utilize large models to refine the annotation process, the underlying evaluation approach remains fundamentally rooted in supervised learning. Therefore, they are categorized under the learning-based evaluation paradigm. \textbf{(4) LLM/ALM-based Evaluation.} Recent advances in large language models and audio-language models have enabled more holistic evaluation mechanisms for audio and speech generation. These models not only produce scalar predictions comparable to Mean Opinion Score (MOS) ratings, but also provide explanatory feedback and descriptive assessments based on both acoustic characteristics and semantic content. Audio-language models trained on large-scale audio-text corpora exhibit strong generalization capabilities, and their performance can be further enhanced through fine-tuning on speech evaluation datasets. Their ability to align multimodal representations across speech and text allows for more comprehensive and interpretable assessments. As a result, this paradigm represents a shift from traditional learning-based evaluation toward a new generation of LLM- and ALM-driven evaluation methods. In this survey, we systematically integrate recent work in this area and provide a comparative analysis between learning-based and LLM/ALM-based approaches in Section~\ref{sec:comparision_mos_eval}, focusing on both evaluation performance and generalization capabilities. \textbf{ (5) Benchmark-based Evaluation.} To support robust and standardized system-level comparisons, recent years have seen the emergence of several benchmark-based evaluation suites for ALMs and MLLMs. These benchmarks aim to systematically assess model capabilities across multiple dimensions, including understanding, interaction, reasoning, and generation, in modalities such as speech, audio, and music. Compared with well-established benchmarks in text to image or text to video generation, evaluation protocols for text to speech, audio, or music generation remain relatively underexplored. Although high-quality datasets such as LibriSpeech \cite{panayotov2015librispeech} and LibriTTS \cite{zen2019libritts} are widely used for training speech generation models, and efforts like TTS-Arena \cite{tts-arena} provide rigorous leaderboard-based evaluation through human-machine interactive scoring, and Seed-TTS \cite{anastassiou2024seed} offers standardized and reliable test sets, these efforts do not fall within the scope of what we define as automated, general-purpose, and multi-task extensible benchmarks for evaluating the capabilities of ALMs and MLLMs. Accordingly, we argue that developing a unified, scalable evaluation framework for speech and audio generation at the ALM level represents an important and underexplored research direction. In this context, we define benchmark-based evaluation as an automated assessment paradigm in which a suite of standardized tasks is used to evaluate ALMs' multimodal capabilities in speech and audio-related scenarios, including understanding, interaction, reasoning, and generation. It is worth noting that most recent benchmark-based evaluation suites have increasingly adopted speech or audio generated by speech and audio generation models as inputs to ALMs or MLLMs, in order to assess their capabilities in understanding, interaction, reasoning, and related tasks. While these benchmarks were not originally designed for generation evaluation, they still offer meaningful signals for assessing generation quality. In particular, evaluating aspects such as semantic consistency, factual alignment, and audio-text coherence in understanding tasks provides critical diagnostic insights for model refinement. These emerging benchmarks thus form the core foundation of what we consider benchmark-based evaluation for ALMs.

Following this conceptual overview, we introduce representative benchmark suites used for meta-evaluating speech and audio generation models (Section~\ref{sec:speech_meta_evaluation_benchmarks}). We conclude with a discussion of persistent challenges and potential directions for advancing automatic evaluation in the era of ALMs and MLLMs. (Section~\ref{sec:speech_conclusion_challenge_future}).

\subsection{Heuristic Evaluation}

Heuristic evaluation refers to the use of predefined, often rule-based or statistical metrics to assess synthesized speech. These methods are lightweight, interpretable, and widely used for evaluating core attributes of audio and speech signals. Based on their assessment objectives, We categorize these methods into three major types, as illustrated in Fig.\ref{img:speech_taxonomy} , and summarize them in Table \ref{tab:heuristic_based_evaluation_speech} :
(1) \textit{quality-based evaluation}, which measures the acoustic fidelity and signal-level similarity of generated speech;
(2) \textit{intelligibility-based evaluation}, which assesses how clearly speech can be perceived and understood under various conditions; and
(3) \textit{content fidelity-based evaluation}, which focuses on the symbolic or semantic accuracy of the generated content, typically via transcription-based comparisons.

\subsubsection*{Quality-based evaluation}
Quality-based evaluation assesses the fidelity of synthesized speech against natural recordings in terms of acoustic characteristics. A range of metrics has been developed to capture aspects such as spectral similarity, pitch accuracy, and signal distortion. Among spectral metrics, Mel Cepstral Distortion (MCD)\cite{kubichek1993mel} is widely used to quantify frame-level spectral differences between generated and reference utterances, enabling fine-grained analysis of short-term frequency alignment. To evaluate prosodic accuracy, F0-CORR and F0-RMSE\cite{hayashi2021espnet2} measure the alignment and deviation of pitch contours, reflecting naturalness in intonation and rhythm. Beyond frame-level features, signal-level metrics provide holistic assessments. Specifically, Signal-to-Distortion Ratio (SDR), Signal-to-Interference Ratio (SIR), and Signal-to-Artifacts Ratio (SAR)\cite{fevotte2005bss_eval} are standard in enhancement and separation tasks, capturing overall distortion, source interference, and artifact presence, respectively. Perceptually motivated metrics have also been introduced, particularly in hearing-aid contexts. The Hearing-Aid Speech Quality Index (HASQI)\cite{kates2010hearing}, for instance, integrates an auditory model to estimate perceived quality and is effective under hearing impairment scenarios.

\subsubsection*{Intelligibility-based evaluation}
Intelligibility based evaluation measures how clearly synthesized speech can be perceived and understood, particularly in noisy or degraded environments. These metrics are essential for evaluating speech usability in real-world conditions. The Short Time Objective Intelligibility (STOI)\cite{taal2011algorithm} estimates intelligibility by correlating the temporal envelopes of clean and degraded speech, showing strong robustness in stationary noise. Its extension, the Extended STOI (ESTOI)\cite{jensen2016algorithm}, accounts for interband dependencies, improving sensitivity in modulated and non-stationary noise. Beyond envelope correlation, perceptual models offer more direct estimates of intelligibility. The Perceptual Evaluation of Speech Quality (PESQ)\cite{rix2001perceptual} simulates auditory perception using models of masking and loudness to assess degradation from noise and distortion. The Scale Invariant Signal to Noise Ratio (SI SNR)\cite{luo2018tasnet} measures intelligibility by comparing target and estimated signals with scale normalization, making it particularly effective in speech separation and enhancement. The Hearing Aid Speech Perception Index (HASPI)~\cite{kates2014hearing} models auditory periphery responses to predict intelligibility across a variety of degradation conditions. Together, these methods provide complementary insights at different abstraction levels, from acoustic detail to perceptual relevance, and are widely used in evaluating speech restoration systems.

\subsubsection*{Content fidelity-based evaluation}
Content fidelity based evaluation measures the symbolic and semantic consistency between synthesized speech and its reference, typically by comparing their transcribed or tokenized forms. This is crucial for text-conditioned generation tasks, where preserving linguistic content is paramount. SpeechBLEU~\cite{saeki2024speechbertscore} extends the traditional BLEU metric to the speech domain by computing n-gram overlaps on discrete speech tokens from self-supervised models, effectively capturing lexical similarity while remaining invariant to acoustic details. In parallel, SpeechTokenDistance~\cite{saeki2024speechbertscore} evaluates sequence-level consistency using edit distance metrics such as Levenshtein or Jaro Winkler, offering a finer measure of token alignment and order. These methods are particularly useful for assessing semantic accuracy in applications like text to speech synthesis and audio captioning. Overall, content fidelity metrics offer a symbolic perspective that complements acoustic and intelligibility based evaluations, and are essential for validating the semantic integrity of text conditioned speech generation systems.

\vspace{5pt}
Heuristic metrics remain widely adopted for evaluating synthesized speech due to their simplicity, efficiency, and interpretability. However, each category presents inherent limitations. Quality-based metrics emphasize acoustic similarity but often fail to capture perceptual naturalness and semantic coherence, resulting in limited alignment with human judgments in expressive generative tasks. Intelligibility-based metrics reliably assess speech clarity under controlled degradation but are sensitive to distortions and typically overlook content fidelity. In contrast, content fidelity-based metrics evaluate symbolic and semantic accuracy, often through transcription-level comparisons, yet they may be affected by acoustic–prosodic mismatches. Among them, edit-distance measures have demonstrated stronger correlations with human perception, particularly in capturing content consistency across speakers and variable-length inputs. Overall, these metrics offer complementary insights and remain essential for benchmarking language-conditioned speech generation, while informing the development of more perceptually and semantically aligned evaluation frameworks.

\begin{table}[htbp]
    \centering
    \resizebox{\textwidth}{!}{
    \begin{tabular}{ccccc}
        \toprule
        \textbf{Method} & \textbf{Task} & \textbf{Objective} & \textbf{Dimension} & \textbf{Function Type} \\
        \midrule
        MCD~\cite{kubichek1993mel} & Speech Generation & Signal Quality & Frame & Spectral Distance \\
        F0-RMSE~\cite{hayashi2021espnet2} & Signal Restoration & Signal Quality & Frame & Pitch Error \\
        SDR~\cite{fevotte2005bss_eval} & Signal Restoration & Signal Quality & Utterance & Signal-to-Noise Ratio \\
        SIR~\cite{fevotte2005bss_eval} & Signal Restoration & Signal Quality & Utterance & Interference Ratio \\
        SAR~\cite{fevotte2005bss_eval} & Signal Restoration & Signal Quality & Utterance & Artifact Ratio \\
        HASQI~\cite{kates2010hearing} & Speech Generation & Signal Quality & Utterance & Perceptual Quality Model \\
        HASPI~\cite{kates2014hearing} & Speech Generation & Intelligibility & Utterance & Perceptual Intelligibility Model \\
        STOI~\cite{taal2011algorithm} & Signal Restoration & Intelligibility & Utterance & Temporal Envelope Correlation \\
        ESTOI~\cite{jensen2016algorithm} & Signal Restoration & Intelligibility & Utterance & Inter-band Temporal Correlation \\
        PESQ~\cite{rix2001perceptual} & Signal Restoration & Intelligibility & Utterance & Perceptual Evaluation Model \\
        SI-SNR~\cite{luo2018tasnet} & Signal Restoration & Intelligibility & Utterance & Scale-Invariant SNR \\
        SpeechBLEU~\cite{saeki2024speechbertscore} & Speech Generation & Content Consistency & Utterance & n-gram Token Overlap \\
        SpeechTokenDistance~\cite{saeki2024speechbertscore} & Speech Generation & Content Consistency & Utterance & Token Sequence Edit Distance \\
        \bottomrule
    \end{tabular}
    }
    \caption{
    Taxonomy of heuristic evaluation methods for speech generation, organized by evaluation objective (signal quality, intelligibility, or content consistency), evaluation granularity (dimension), and function type. The dimension column indicates the temporal resolution at which the metric is applied. Frame-level metrics operate on short, fixed-length segments of the speech waveform, enabling fine-grained analysis of local acoustic properties such as pitch or spectral shape. In contrast, utterance-level metrics are computed over entire spoken segments and capture global characteristics like intelligibility, distortion, or semantic consistency. The function type column reflects the methodological basis of each metric, indicating how it operationalizes similarity, distortion, or semantic alignment in the evaluation process.
    }
    \label{tab:heuristic_based_evaluation_speech}
\end{table}

\subsection{Embedding-based Evaluation}

Embedding-based evaluation methods have emerged as a promising alternative to heuristic metrics by addressing their limitations through the use of high-level perceptual and semantic embeddings learned from deep neural networks. These methods facilitate more content-aware and perceptually aligned evaluation frameworks, particularly in scenarios where low-level acoustic metrics fail to capture quality aspects aligned with human perception. A summary of representative embedding-based evaluation methods, along with their evaluation objectives, reference requirements, and protocols, is provided in Table~\ref{tab:embedding_based_evaluation_speech}. Depending on their evaluation focus, these methods can be grouped into two major categories: (1) \textit{perceptual-based metrics}, which aim to approximate human auditory perception, and (2) \textit{semantic-based metrics}, which assess the alignment between audio and corresponding linguistic or contextual information.

\subsubsection*{Perceptual-based methods}

Perceptual-based metrics evaluate audio quality by modeling perceptual similarity within embedding spaces, rather than relying on direct comparisons of acoustic features. A representative example is PAM~\cite{deshmukh2024pam}, which computes pairwise distances between audio samples using prompting audio-language models in order to approximate human judgments. Distribution-based metrics such as Fréchet Audio Distance (FAD)\cite{kilgour2018fr}, Kernel Inception Distance (KID), and Kullback–Leibler Divergence (KLD) operate on the statistical distributions of embeddings extracted from real and generated audio signals. These approaches are commonly adopted in general-purpose audio generation tasks and are particularly suitable when reference signals are unavailable or when evaluating long-form, high-dimensional audio content. MusicEval-Score~\cite{liu2025musiceval} extends this distributional framework to the music domain by leveraging CLAP-based embeddings to capture corpus-level perceptual characteristics. In addition, NOMAD~\cite{ragano2024nomad} introduces an unsupervised approach that estimates perceptual audio quality by comparing degraded speech signals to unrelated clean references in the embedding space, thereby removing the requirement for ground-truth alignments or human annotations.

\subsubsection*{Semantic-based methods}

Semantic-based metrics focus on assessing whether the generated audio content maintains semantic alignment with the associated text or contextual information. CLAP-Score~\cite{xiao2024reference} applies cross-modal embeddings to evaluate semantic consistency between audio and its reference text under a single-reference setting. SpeechBERTScore~\cite{saeki2024speechbertscore} captures contextual alignment by comparing embeddings generated from pretrained speech encoders across paired utterances. More recently, TTSDS~\cite{minixhofer2024ttsds} introduces token-level distance metrics to quantify semantic fidelity in text-to-speech (TTS) systems. These methods are particularly effective in language-conditioned generation scenarios such as TTS and audio captioning, where preserving the semantic content of the input is essential. Furthermore, they are capable of detecting semantic errors including omissions, hallucinated content, and topic drift, which are often overlooked by conventional acoustic-based evaluation methods.

\vspace{5pt}
Embedding-based evaluation provides a unified and extensible framework that simultaneously captures perceptual similarity and semantic fidelity, effectively addressing key limitations of traditional heuristic metrics. Specifically, perceptual-based methods approximate human auditory perception using learned embeddings, enabling robust and reference-efficient assessment across a variety of acoustic conditions. Complementing this, semantic-based methods focus on preserving linguistic meaning and ensuring contextual alignment, which are essential for evaluating language-conditioned generation tasks such as TTS and TTA. Looking ahead, the advancement of large-scale multimodal embedding architectures, such as those exemplified by Gemini \cite{lee2025gemini} and related vision-language-speech models \cite{wu2024next,girdhar2023imagebind}, is expected to expand the capabilities of embedding-based evaluation. These developments may facilitate more comprehensive and generalizable evaluation strategies for complex generative systems involving cross-modal alignment and audio-grounded semantic understanding. As a result, embedding-based methods are anticipated to play a central role in the future of perceptually grounded, semantically coherent, and scalable evaluation frameworks.

\begin{table}[htbp]
\centering
\scalebox{0.9}{
\begin{tabular}{ccccc}
\toprule
\textbf{Method} & \textbf{Task} & \textbf{Category} & \textbf{Reference} & \textbf{Protocol} \\
\midrule
PAM~\cite{deshmukh2024pam} & Audio Generation & Perceptual-based & No & Pairwise \\
FAD~\cite{kilgour2018fr} & Audio Generation & Perceptual-based & Yes & Corpus \\
KID~\cite{kilgour2018fr} & Audio Generation & Perceptual-based & Yes & Corpus \\
KLD~\cite{kilgour2018fr} & Audio Generation & Perceptual-based & Yes & Corpus \\
MusicEval-Score~\cite{liu2025musiceval} & Music Generation & Perceptual-based & Yes & Corpus \\
NOMAD~\cite{ragano2024nomad} & Audio/Speech Enhancement & Perceptual-based & Yes & Pairwise \\
CLAP-Score~\cite{xiao2024reference} & Audio Generation & Semantic-based & Yes & Single \\
SpeechBERTScore~\cite{saeki2024speechbertscore} & Speech Generation & Semantic-based & Yes & Pairwise \\
TTSDS~\cite{minixhofer2024ttsds} & Speech Generation & Semantic-based & Yes & Corpus \\
\bottomrule
\end{tabular}
}
\caption{Taxonomy of embedding-based evaluation methods for speech, music, and audio generation tasks, categorized by evaluation objective (perceptual or semantic), reference requirement, and evaluation protocol.}
\label{tab:embedding_based_evaluation_speech}
\end{table}

\subsection{Learning-based Evaluation}

Learning-based evaluation seeks to predict human-perceived quality by training models on subjective quality annotations. This paradigm has become a cornerstone in evaluating speech and audio generation systems, particularly for estimating perceptual properties such as the Mean Opinion Score (MOS). Broadly, existing methods can be divided into two major categories which are shown in Table~\ref{tab:learning_based_evaluation_speech} : (1) \textit{supervised-based methods}, which train neural networks directly on human-labeled MOS datasets to simulate subjective evaluation behavior, and (2) \textit{self-supervised-based methods}, which leverage acoustic representations derived from large-scale unlabeled corpora to enhance prediction performance, especially in data-scarce scenarios.

\begin{table}[htbp]
\centering
\scalebox{0.9}{
\begin{tabular}{ccccc}
\toprule
\textbf{Method} & \textbf{Task} & \textbf{Category} & \textbf{Reference} & \textbf{Protocol} \\
\midrule
DNSMOS~\cite{reddy2021dnsmos} & Text-to-Speech & Supervised-based & No & Single \\
UMs~\cite{ravuri2024uncertainty} & Text-to-Speech & Supervised-based & No & Single \\
WARP-Q~\cite{jassim2021warp} & Text-to-Speech & Supervised-based & Yes & Single \\
AUTOMOS~\cite{patton2016automos} & Text-to-Speech & Supervised-based & No & Single \\
DysarNet~\cite{liu2024non} & Text-to-Speech & Supervised-based & No & Single \\
MBNet~\cite{leng2021mbnet} & Text-to-Speech & Supervised-based & No & Single \\
Quality-Net~\cite{fu2018quality} & Text-to-Speech & Supervised-based & No & Single \\
\cite{yoshimura2016hierarchical} & Text-to-Speech & Supervised-based & No & Single \\
\cite{mittag2021deep} & Text-to-Speech & Supervised-based & No & Single \\
MOSNET~\cite{lo2019mosnet} & Text-to-Speech & Supervised-based & No & Single \\
MOSA-Net~\cite{MOSA-Net} & Text-to-Speech & Supervised-based & No & Single \\
MOSA-Net+~\cite{MOSA-Net+} & Text-to-Speech & Supervised-based & No & Single \\
TORCHAUDIO-SQUIM~\cite{kumar2023torchaudio} & Text-to-Speech / Text-to-Audio & Supervised-based & No & Single \\
\cite{tian2022multi} & Text-to-Speech & Supervised-based & No & Single \\
DeePMOS~\cite{liang2023deepmos} & Text-to-Speech & Supervised-based & No & Single \\
\cite{cumlin2025impairments} & Text-to-Speech & Supervised-based & No & Single \\
\cite{kunevsova2023ensemble} & Text-to-Speech & Supervised-based & No & Single \\
LDNet~\cite{huang2022ldnet} & Text-to-Speech & Supervised-based & No & Single \\
DDOS~\cite{tseng2022ddos} & Text-to-Speech & Supervised-based & No & Single \\
MBI-Net~\cite{zezario2022mbi} & Text-to-Speech & Supervised-based & No & Single \\
MOSLight~\cite{li2023moslight} & Text-to-Speech & Supervised-based & No & Single \\
SpeechLMScore~\cite{maiti2023speechlmscore} & Text-to-Speech & Supervised-based & No & Single \\
ADT-MOS~\cite{liang2025adtmos} & Text-to-Speech & Supervised-based & No & Single \\
UAMOS~\cite{wang2024uncertainty} & Text-to-Speech & Supervised-based & No & Single \\
ADIFF~\cite{deshmukh2025adiff} & Text-to-Audio & Supervised-based & No & Pairwise \\
Audiobox Aesthetics~\cite{tjandra2025meta} & Text-to-Audio & Supervised-based & No & Single \\
\midrule
SSL-MOS~\cite{ssl-mos} & Text-to-Speech & SSL-based & No & Single \\
MTQ-Net~\cite{mtqnet} & Text-to-Speech & SSL-based & No & Single \\
\cite{ssr-mos} & Text-to-Speech & SSL-based & No & Single \\
\cite{sultana2025pre} & Text-to-Speech & SSL-based & No & Single \\
\cite{vioni2023investigating} & Text-to-Speech & SSL-based & No & Single \\
\cite{nguyen2024exploring} & Text-to-Speech & SSL-based & No & Single \\
SCOREQ~\cite{ragano2024scoreq} & Text-to-Speech & SSL-based & Optional & Single \\
IndicMOS~\cite{udupa2024indicmos} & Text-to-Speech & SSL-based & No & Single \\
UTMOS~\cite{saeki2022utmos} & Text-to-Speech & SSL-based & No & Single \\
UTMOSv2~\cite{baba2024t05} & Text-to-Speech & SSL-based & No & Single \\
RAMP~\cite{wang2023ramp} & Text-to-Speech & SSL-based & No & Single \\
RAMP+~\cite{wang2025ramp+} & Text-to-Speech & SSL-based & No & Single \\
AlignNet~\cite{pieper2024alignnet} & Text-to-Speech & SSL-based & No & Single \\
SQuId~\cite{sellam2023squid} & Text-to-Speech & SSL-based & No & Single \\
Uni-VERSA~\cite{shi2025uni} & Text-to-Speech & SSL-based & Yes & Single \\
ARECHO~\cite{shi2025arecho} & Text-to-Speech & SSL-based & Optional & Single \\
Distill-MOS~\cite{stahl2025distillation} & Text-to-Speech & SSL-based & No & Single \\
HAAQI-Net~\cite{wisnu2025haaqi} & Text-to-Audio & SSL-based & No & Single \\
UPPSQA~\cite{UPPSQA} & Text-to-Speech & SSL-based & No &  Pairwise \\
\bottomrule
\end{tabular}
}
\caption{Categorization of learning-based evaluation methods for speech and audio generation. Methods are categorized by generation task (Text-to-Speech or Text-to-Audio), learning strategy (Supervised or Self-Supervised), reference dependency (whether a reference speech or signal is required), and evaluation protocol (Single, Pairwise, or Corpus-level). This taxonomy reflects the diversity of current mainstream learning-based approaches.}
\label{tab:learning_based_evaluation_speech}
\end{table}

\subsubsection*{Supervised-based methods}
Supervised models are trained on annotated datasets to learn direct mappings from acoustic signals to perceived quality. Early works, such as DNSMOS~\cite{reddy2021dnsmos}, Quality-Net~\cite{fu2018quality}, and MBNet~\cite{leng2021mbnet}, adopt spectral or perceptual features as input (e.g., STFT or log-mel), and employ convolutional or recurrent networks to perform MOS regression. To enhance temporal modeling and capacity, more recent models integrate advanced architectural designs. For instance, MOSNET~\cite{lo2019mosnet} and MOSA-Net~\cite{MOSA-Net} leverage attention mechanisms to capture long-range dependencies, while MOSA-Net+~\cite{MOSA-Net+} deepens the network through residual connections. Lightweight models such as LDNet~\cite{huang2022ldnet}, MOSLight~\cite{li2023moslight}, and DeePMOS~\cite{liang2023deepmos} emphasize inference efficiency using compression-friendly components. Several supervised methods also address domain generalization and uncertainty. SpeechLMScore~\cite{maiti2023speechlmscore} uses speech-language pretraining for more transferable representations. UAMOS~\cite{wang2024uncertainty} and ADT-MOS~\cite{liang2025adtmos} incorporate uncertainty modeling and domain adaptation to enhance robustness in mismatched conditions. ADIFF~\cite{deshmukh2025adiff}, focusing on alignment-sensitive scenarios, introduces a cross-projection module and a multi-stage training strategy to capture fine-grained semantic and emotional degradations between audio recordings. Audiobox Aesthetics~\cite{tjandra2025meta} proposes a no-reference aesthetic assessment model that decomposes perceptual judgment into four human-interpretable dimensions, enabling nuanced prediction of audio pleasantness in generative settings. In contrast to fully blind approaches, methods like WARP-Q~\cite{jassim2021warp} retain reference-based estimation to ensure stronger perceptual anchoring. As a result, supervised models provide a wide range of trade-offs between interpretability, scalability, and perceptual fidelity, making them well-suited for MOS estimation under varied conditions.

\subsubsection*{Self-supervised-based methods}
To overcome the reliance on labeled data, self-supervised learning (SSL) methods utilize speech representations pre-trained on large unannotated corpora. These embeddings, derived from foundation models such as Wav2vec 2.0~\cite{baevski2020wav2vec}, HuBERT~\cite{hsu2021hubert}, or WavLM~\cite{chen2022wavlm}, provide rich contextual and phonetic cues for downstream quality prediction tasks. Foundational SSL-based models include SSL-MOS~\cite{ssl-mos} and MTQ-Net~\cite{mtqnet}, which combine frozen or fine-tuned speech encoders with simple regressors. More advanced systems such as RAMP~\cite{wang2023ramp} enhance decoder performance by retrieving relevant instances from a large corpus of annotated data, using a fusion network to dynamically adjust retrieval scope and confidence-aware weighting. RAMP+\cite{wang2025ramp+} further incorporates prior knowledge and adaptive retrieval mechanisms to improve domain robustness. UTMOSv2\cite{baba2024t05}, in contrast, fuses spectrogram-based and SSL-based features through separate predictors and a fine-tuning stage that improves calibration and generalization. SQuId~\cite{sellam2023squid} applies knowledge distillation to consolidate diverse SSL variants into a unified representation. SCOREQ~\cite{ragano2024scoreq} introduces a contrastive triplet loss objective that addresses generalization failures of L2-based regression and enhances predictive consistency across domains. IndicMOS~\cite{udupa2024indicmos} extends SSL-based modeling to multilingual contexts using a universal encoder for Indian languages. ARECHO~\cite{shi2025arecho}, built upon Uni-VERSA~\cite{shi2025uni}, proposes dynamic classifier chaining and confidence-aware decoding to jointly estimate correlated speech metrics such as PESQ, STOI, and MOS. Uni-VERSA itself offers a unified architecture for predicting naturalness, intelligibility, speaker similarity, and prosody, thereby enabling comprehensive evaluation across multiple perceptual axes. To support lightweight deployment, Distill-MOS~\cite{stahl2025distillation} adopts model pruning and distillation strategies, significantly reducing model size while maintaining high fidelity to human ratings. HAAQI-Net~\cite{wisnu2025haaqi} brings SSL to the domain of music evaluation in hearing aid contexts, integrating BLSTM attention with BEATs-based features and demonstrating robustness under varying sound pressure levels. UPPSQA~\cite{UPPSQA} addresses the generalization limitations of existing preference-score-based SQA methods in content-matched scenarios by leveraging a semantic-acoustic-driven MOS prediction model and training on diverse types of paired speech data. Altogether, these developments illustrate the scalability, flexibility, and label-efficiency of SSL-based evaluation, making them particularly well-suited to low-resource, multilingual, and reference-free quality assessment scenarios.

\vspace{5pt}
With the increasing maturity of learning based evaluation models, recent research trends reveal a gradual shift toward unified multi-metric architectures and integration with foundation models. On one hand, traditional MOS specific predictors are being extended to simultaneously estimate multiple perceptual metrics such as intelligibility, naturalness, and speaker similarity within a single framework. Representative examples include Uni-VERSA~\cite{shi2025uni} and ARECHO~\cite{shi2025arecho}, which demonstrate how classifier chaining, joint optimization, and confidence aware inference can offer consistent multidimensional predictions. This unified modeling strategy not only improves computational efficiency but also enhances interpretability and coherence across evaluation axes. On the other hand, the reliability and scalability of supervised MOS prediction still depend on the availability of large scale, high quality human annotations. Future improvements in this domain may rely on more structured and diverse crowdsourcing pipelines, as well as better alignment between annotation protocols and perceptual dimensions. Moreover, self supervised learning continues to evolve beyond conventional SSL encoders. Recent developments have begun to explore the use of large language models (LLMs) or audio language models (ALMs) for quality prediction, treating evaluation as a generation or reasoning task conditioned on audio input. While LLM or ALM based evaluation is currently in its early stages, its potential to model context, semantics, and perception holistically suggests a promising future trajectory. Looking ahead, we expect that the next generation of learning based evaluation methods will increasingly integrate unified multi metric modeling, improved data annotation protocols, and foundation model reasoning, leading toward more robust, explainable, and transferable quality assessment systems.

\subsection{LLM/ALM-based Evaluation}

The emergence of Large Language Models (LLMs) and Audio-Language Models (ALMs) has opened new possibilities for automatic evaluation of synthesized speech. In this context, as shown in Table~\ref{tab:llm_based_eval_speech}, current approaches can be broadly classified into two categories: (1) \textit{prompt-based evaluation}, which leverages the zero-shot capabilities of general-purpose LLMs through carefully constructed prompts, and (2) \textit{fine-tuning-based evaluation}, where audio-capable LLMs are adapted to speech assessment tasks via parameter-efficient fine-tuning.

\subsubsection*{Prompt-based methods}
Prompt-based methods provide a reference-free and training-free paradigm for speech quality evaluation. A notable example is GPT-Whisper~\cite{GPT-Whisper}, which integrates OpenAI’s Whisper ASR system with the multimodal GPT-4o. Specifically, the system transcribes input speech using Whisper and then prompts GPT-4o to assess the naturalness of the transcribed output. In contrast, intelligibility is evaluated by directly querying GPT-4o with the raw audio input, without relying on intermediate transcription. Despite its straightforward design, GPT-Whisper demonstrates a moderate correlation with human MOS score, and exhibits strong consistency with ASR metrics such as character error rate (CER). These results suggest that prompt-based LLM evaluation offers a viable solution for low-resource or reference-scarce scenarios.

\subsubsection*{Fine-tuning-based methods}
Fine-tuning-based methods adapt pre-trained audio-language models to address diverse speech quality assessment tasks.This approach is particularly motivated by the limited availability of large-scale, high-quality annotated datasets for perceptual evaluation. Since collecting such labels typically relies on costly human annotation or crowd-sourced MOS scoring, existing public resources remain scarce and are primarily represented by benchmarks such as BVCC~\cite{huang2022voicemos}, NISQA~\cite{mittag2021nisqa}, SOMOS~\cite{maniati2022somos}. These meta-evaluation benchmarks will be discussed in detail in later sections. To mitigate data limitations while maintaining adaptability, a representative design leverages auditory LLMs fine-tuned using parameter-efficient techniques such as LoRA on annotated speech quality datasets. These models are guided by task-specific prompts and support a wide range of evaluation functions, including mean opinion score (MOS) and speaker similarity (SIM) prediction, A/B preference testing, and multi-aspect natural language description generation. Experimental results of SALMONN-lora~\cite{lora} and Qwen-Lora~\cite{lora} demonstrate that such fine-tuned auditory LLMs can serve as versatile speech quality evaluators, achieving competitive performance relative to state-of-the-art task-specific models across multiple assessment scenarios. More recently, the Alignment with LLM Distillation (ALLD) method~\cite{alld} has been proposed to enhance the perceptual reasoning capabilities of ALMs. Existing ALMs often lack awareness of input speech quality, primarily because speech evaluation tasks are typically excluded from multitask training due to the scarcity of well-annotated datasets. To address this limitation, ALLD introduces a natural language-based evaluation corpus that extends beyond traditional MOS labels to include multidimensional quality attributes, degradation analyses, and descriptive A/B comparisons. Based on this resource, ALLD applies token-level distillation to align ALM outputs with expert annotations, enabling the model to perform a range of evaluation tasks. These include MOS prediction, pairwise quality judgments, and generating natural language justifications grounded in audio content. By integrating perceptual supervision into model training, ALLD advances the development of ALMs capable of producing human-aligned assessments and interpretable responses, thereby contributing to more perceptually aware and reliable multimodal agents.

Recent advances in LLM/ALM-based evaluation reflect several key developments:
\begin{itemize}
    \item \textbf{From fully supervised training to parameter-efficient generalization:} Earlier approaches often required extensive labeled data and full-model training for speech and audio quality prediction. In contrast, recent methods pursue more efficient paradigms, such as parameter-efficient fine-tuning (e.g., LoRA) and prompt-based zero-shot inference, enabling better scalability and domain transfer with minimal supervision.

    \item \textbf{From scalar prediction to interpretable evaluation:} Previous learning-based systems typically focused on predicting a single quality score, such as MOS. In contrast, modern ALM-based methods not only generate the MOS score, but also generate human-readable justifications and natural language descriptions, enabling richer and more interpretable evaluation outputs.

    \item \textbf{From task-specific to unified auditory evaluation:} Previous approaches often developed separate evaluation models for different audio types such as speech, non-speech sound, and music. Recent advancements in audio-language models enable unified evaluation across these auditory domains within a single architecture, allowing for improved scalability, consistency, and transferability. This shift leverages the multi-modal and multi-task capabilities of advanced ALMs to support diverse and holistic assessment scenarios.
\end{itemize}

\begin{table}[htbp]
\centering
\resizebox{\textwidth}{!}{
\begin{tabular}{lcccc}
\toprule
\textbf{Method} & \textbf{Model} & \textbf{Finetuning} & \textbf{Type} & \textbf{Dimension} \\
\midrule
GPT-Whisper~\cite{GPT-Whisper} & Whisper + GPT-4o & No & Prompt-based & Quality, Intelligibility \\
SALMONN-Lora~\cite{lora} & SALMONN (vic1.0\&1.5) & Yes (LoRA) & Fine-tuning-based & MOS, SIM, Descriptions \\
Qwen-1\&2-Lora~\cite{lora} & Qwen1\&2-Audio & Yes (LoRA) & Fine-tuning-based & MOS, SIM, Descriptions \\
ALLD~\cite{alld} & Qwen2-Audio + LLM Distillation & Yes (Token-level) & Fine-tuning-based & MOS, SIM, Explanations \\
\bottomrule
\end{tabular}
}
\caption{Taxonomy of LLM/ALM-based evaluation methods for automatic synthesized speech evaluation. This table summarizes recent approaches that leverage LLM/ALMs for speech quality assessment and MOS prediction. Methods are categorized by model architecture, whether fine-tuning is applied, evaluation type (prompt-based or fine-tuning-based), and the range of predicted perceptual dimensions, including Mean Opinion Score (MOS), intelligibility, speaker similarity (SIM), and natural language explanations or descriptions. The table highlights the increasing trend toward multi-dimensional, interpretable evaluation frameworks powered by generative and multimodal language models.}
\label{tab:llm_based_eval_speech}
\end{table}

\subsection{Benchmark-based Evaluation}

Benchmark-based evaluation has become an essential component in assessing speech generation systems, offering structured protocols grounded in curated datasets and standardized tasks. These benchmarks support reproducible comparisons across diverse evaluation dimensions, including comprehension, generation quality, safety, and interactive performance. Based on recent developments, we categorize benchmark-based evaluation into four major types which is shown in Table~\ref{tab:benchmark_eval_speech} :
(1) \textit{Understanding-oriented benchmarks}, which assess a model’s ability to interpret, reason over, and semantically ground audio input across domains such as speech, music, and environmental sound;
(2) \textit{Generation-oriented benchmarks}, which evaluate the quality, intelligibility, speaker consistency, and prosodic control of synthesized audio, often combining subjective and objective metrics;
(3) \textit{Interaction-oriented benchmarks}, which examine dialogue coherence, responsiveness, Instruction Following, and other characteristics essential for real-time voice-based interaction.
and (4) \textit{Safety-oriented benchmarks}, which assess the robustness and ethical alignment of audio-language models under adversarial, misleading, or sensitive input conditions. These include evaluations of jailbreak susceptibility, refusal behavior, multilingual safety alignment, and increasingly, detection of speech deepfakes and model misuse.

\subsubsection*{Understanding-oriented benchmarks}
Understanding-oriented benchmarks aim to evaluate a model’s capacity to comprehend, reason over, and generate responses grounded in speech, audio, or music inputs. This category includes several subtypes:

\textbf{QA-based Understanding.}
Several recent benchmarks adopt question-answering formats to evaluate a model's ability to extract structured and context-dependent knowledge directly from raw audio inputs. These tasks span diverse domains such as music, speech, environmental sounds, and animal vocalizations, and aim to probe both multimodal reasoning and auditory perception capabilities. MuChoMusic~\cite{weck2024muchomusic} targets musical understanding through 1,187 multiple-choice questions validated by human annotators across 644 tracks. It evaluates factual knowledge of music theory and history as well as stylistic and cultural interpretation. However, high scores from text-only models even with noise inputs reveal a heavy reliance on language priors, highlighting the need for more perceptually grounded evaluation. To address this limitation, RUListening~\cite{zang2025you} introduces the Perceptual Index, a metric that quantifies the dependence of each question on audio perception. By generating adversarial distractors and analyzing model uncertainty, it constructs QA items that require genuine auditory understanding and reveal significant performance gaps between audio-grounded and language-only models. SALMon~\cite{maimon2025salmon} evaluates fine-grained acoustic awareness in speech, focusing on aspects such as background noise, emotion, speaker identity, and room acoustics. Instead of using direct classification, it adopts a model-based scoring method that compares log-likelihoods of correct and incorrect samples, offering a scalable approach to measure sensitivity to acoustic cues. MAE~\cite{chen2024beyond} shifts focus to multi-audio scenarios, assembling 20 datasets across 11 tasks that involve speech and environmental sound. It evaluates the ability of models to process simultaneous or sequential audio streams, reflecting real-world auditory complexity. The proposed MALLM model demonstrates that synthetic multi-audio data can improve performance without extensive manual annotation. In the domain of bioacoustics, BEANS-Zero~\cite{robinson2024naturelm} provides a zero-shot benchmark for interpreting animal vocalizations. Covering classification and detection tasks across diverse species, it assesses generalization in settings with limited training data and emphasizes domain-specific audio-language alignment. These QA-based benchmarks are designed to systematically probe the capabilities of audio-language models in specific domains by formulating well-defined question-answering tasks. Rather than testing shallow pattern matching, they focus on whether models can extract structured, context-dependent information from audio inputs and ground their answers in perceptual and semantic understanding. 

\textbf{Reasoning-based Understanding.}
With the continued advancement of audio-language models (ALMs), the focus of speech processing research is gradually shifting from low-level perceptual tasks such as speaker identification, emotion classification, and speech transcription toward higher-level semantic and contextual understanding. This shift has led to the emergence of reasoning-oriented benchmarks that systematically evaluate a model’s ability to comprehend and infer from complex audio inputs, assessing not only recognition accuracy but also performance in multi-step reasoning, temporal sequencing, and multimodal integration. SAKURA \cite{yang2025sakura} presents a structured suite of tasks covering attributes such as speaker gender, language, and emotional state, emphasizing multi-hop reasoning; while models perform well on direct perception, they often struggle to integrate information across reasoning steps. JASCO \cite{wang2025they} targets joint reasoning over environmental sounds and human speech, showing that current models tend to rely disproportionately on one modality, revealing challenges in fusion. To evaluate logical consistency and sound understanding, \cite{kuan2025can} introduces diagnostic tasks on object existence, event ordering, and sound attribution, using before-and-after audio comparisons. By guiding models to produce intermediate auditory descriptions prior to answering, this work demonstrates improved prediction accuracy and interpretability. MMAR \cite{ma2025mmar} expands the reasoning evaluation scope with 1,000 QA items categorized into signal, perception, semantic, and cultural levels, each annotated with chain-of-thought rationales to probe multi-step inference. Results indicate that even advanced models face difficulties when reasoning requires domain knowledge or abstract understanding. Building on these insights, the Audio Entailment benchmark \cite{deshmukh2025audio} evaluates whether textual descriptions can be logically inferred from audio content, offering entailment, neutral, and contradiction labels across hypotheses generated from AudioCaps and Clotho. Current ALMs show limited performance under zero-shot and linear-probe conditions. Together, these benchmarks push ALM evaluation beyond surface-level recognition, clarifying model capabilities in multimodal integration, logical inference, and causality, and pointing to future directions for model development and task design.

\textbf{Multi-task Understanding.}
As ALMs evolve toward broader task generalization capabilities, the evaluation paradigm is also transitioning from early QA- and reasoning-based benchmarks focused on isolated skill testing to a more comprehensive framework encompassing multi-task processing, multimodal understanding, and instruction generalization. Dynamic-SUPERB Phase-2 \cite{huang2024dynamic} introduces 180 classification, regression, and generation tasks across speech, music, and environmental sound. Results show considerable performance variation across tasks, and no model has yet demonstrated stable, all-around proficiency, highlighting the challenges of task generalization and instruction comprehension. AudioBench \cite{wang2024audiobench} focuses on speech understanding, audio scene recognition, and paralinguistic analysis. It incorporates diverse instruction templates and audio-text open-ended tasks to test model robustness in long-form and instruction-driven scenarios, revealing persistent limitations in general audio reasoning. UltraEval-Audio \cite{ultraeval-audio2025} offers the first fully open-source benchmark framework supporting both speech understanding and generation. It consolidates 34 authoritative benchmarks across 10 languages, with automated and standardized evaluation workflows, greatly enhancing evaluation efficiency in multilingual and cross-task settings. FinAudio \cite{cao2025finaudio} targets the long-overlooked financial speech domain, defining three core tasks: short-form ASR, long-form ASR, and summarization. Results show that models still face significant challenges in processing long-form financial audio, and open-source models demonstrate particular advantages in privacy-sensitive applications. SAGI Benchmark \cite{bu2024roadmap} proposes a five-stage roadmap for advancing speech understanding capabilities, progressing from basic ASR to complex tasks involving abstract acoustic reasoning and paralinguistic signal interpretation, thereby identifying current limitations in general speech intelligence. These multi-task evaluation benchmarks signify a shift in the assessment of audio-language models from fragmented capability testing to a more comprehensive and systematic approach. They promote cross-domain, multimodal, and contextually complex evaluations, while providing a unified platform and critical support for developing next-generation ALMs with stronger generalization ability, instruction adaptability, and real-world robustness.

\subsubsection*{Generation-oriented benchmarks}
Generation-oriented benchmarks aim to systematically evaluate the perceptual and statistical quality of speech synthesis systems, covering a wide range of aspects from low-level acoustic features to high-level human perception. The evaluation subtype includes:

\textbf{Quality-based Evaluation.} DiscreteEval \cite{wang2024evaluating} proposes a quantitative evaluation framework that examines model performance across intelligibility, prosody, speaker consistency, and naturalness. Speech generation models based on discrete speech tokens outperform traditional TTS in naturalness and prosodic variation but still fall short in intelligibility and speaker consistency, and are prone to hallucinations or non-speech artifacts. Experiments also suggest that scaling up model size brings modest improvements in robustness.

\textbf{Multi-dimensional Evaluation.} EmergentTTS-Eval \cite{manku2025emergenttts} introduces a comprehensive evaluation suite targeting six complex synthesis scenarios, including emotional expression, paralinguistic cues, foreign words, syntactic complexity, difficult pronunciations, and questions. It contains 1,645 test samples automatically generated and expanded by large language models. This benchmark adopts a model-as-judge paradigm, where audio-language models assess speech outputs across dimensions such as prosody, emotion, intonation, and pronunciation accuracy. The model evaluations show strong alignment with human judgments. Applied to both open-source and proprietary TTS systems, EmergentTTS-Eval reveals nuanced performance differences in open-ended generation tasks, offering fine-grained insights for future model improvement.

\subsubsection*{Interaction-oriented benchmarks}
In current evaluation frameworks for speech interaction, researchers typically categorize assessment tasks based on the specific interactive capabilities of ALMs. This task taxonomy enables more precise characterization of model performance across complex human-computer spoken interaction scenarios. The evaluation tasks are generally divided into the following five subcategories:

\textbf{QA-based. }This category focuses on a model’s ability to understand and answer spoken questions, assessing whether the model can accurately extract query elements from audio inputs and generate semantically appropriate responses. Representative benchmarks include VoxEval \cite{cui2025voxeval} and Audiopedia \cite{penamakuri2025audiopedia}. VoxEval addresses the limitations of existing QA benchmarks in supporting end-to-end speech-based evaluation by introducing a fully spoken QA framework, which assesses ALMs’ knowledge understanding and complex reasoning abilities under diverse acoustic conditions. Complementing this, Audiopedia constructs three sub-tasks, namely single-audio QA, multi-audio QA, and retrieval-augmented QA, to comprehensively evaluate models’ capabilities in audio understanding and external knowledge reasoning, filling a gap left by conventional AQA tasks in knowledge-intensive scenarios.

\textbf{Spoken Dialogue-based.} This category emphasizes the coherence and interactivity of models in multi-turn speech-driven dialogue, requiring a robust understanding of context and maintenance of semantic consistency. Representative benchmarks include Full-Duplex-Bench \cite{Full-Duplex-Bench}, ContextDialog \cite{kim2025does}, and Talking Turns \cite{arora2025talking}. Full-Duplex-Bench addresses the coarse granularity and limited behavioral evaluation in traditional half-duplex models by systematically incorporating dimensions such as pause handling, interruption management, and turn-taking for real-time interaction evaluation. Building on this, ContextDialog targets models’ ability to retain and utilize spoken history across multi-turn dialogues, revealing the memory limitations of open-source models. Further extending this line of inquiry, Talking Turns introduces an evaluation protocol that assesses when models should speak, whether they interrupt, or leave inappropriate silences, thereby testing models' turn-taking prediction and conversational timing skills.

\textbf{Conversational Behavior-based.} This category extends spoken dialogue tasks by focusing on models’ sensitivity to and adaptation toward paralinguistic cues such as emotion, prosody, and speaking style. Representative benchmarks include StyleTalk \cite{lin2024advancing}, SD-Eval \cite{ao2024sd-eval}, VoxDialogue \cite{cheng2025voxdialogue} and Vox-Profile \cite{feng2025vox}. StyleTalk assesses the model’s ability to perceive speaking style and generate style-consistent responses under controlled semantics. Building on the need for richer paralinguistic understanding, SD-Eval provides a comprehensive benchmark incorporating emotion, accent, age, and background noise, using both objective and subjective evaluations to analyze how models adapt to complex paralinguistic environments. In parallel, VoxDialogue identifies twelve non-textual acoustic attributes, testing models’ ability to recognize and utilize cues such as rhythm, emphasis, and background sound to generate coherent and natural conversations. Finally, Vox-Profile proposes a multidimensional evaluation scheme combining static and dynamic vocal traits to measure how well ALMs generalize across diverse speaker profiles, support personalized speech recognition and generation, and maintain consistent performance in customized interaction scenarios.

\textbf{Instruction Following-based.} This category evaluates whether models can correctly interpret spoken commands and generate corresponding actions or speech responses. Representative benchmarks include S2S-Arena \cite{jiang2025s2s}, EvalSIFT \cite{pandey2025sift}, and Speech-IFEval \cite{Speech-ifeval}. S2S-Arena focuses on real-world speech-to-speech tasks, incorporating paralinguistic features to evaluate instruction comprehension and style consistency across tasks and domains. Expanding upon this, EvalSIFT builds on a large-scale multilingual speech-text instruction dataset and offers standardized test sets to assess models’ generalization across tasks and languages. To further diagnose model limitations, Speech-IFEval decouples speech perception from instruction-following, providing a diagnostic framework that reveals performance instability and prompt sensitivity in basic instruction execution.

Together, these subcategories reflect the evolution of speech interaction evaluation from single-turn response to multi-turn dialogue, from linguistic understanding to paralinguistic adaptation, and from general-purpose tasks to personalized interaction. This taxonomy offers a clear framework for systematically analyzing the capability boundaries of ALMs in real-world interaction settings.

\subsubsection*{Safety-oriented benchmarks}
This category focuses on evaluating the robustness, ethical alignment, and refusal behaviors of Audio Language Models (ALMs) under adversarial, deceptive, or potentially harmful audio conditions. Representative benchmarks are typically grouped into two subtypes: 

\textbf{Multi-dimensional safety-based. } AudioTrust~\cite{li2025audiotrust} proposes the first comprehensive trustworthiness benchmark for ALMs, covering six core dimensions—fairness, hallucination, safety, privacy, robustness, and authentication—through 18 real-world experimental setups and 9 audio-specific evaluation metrics, offering a scalable framework for ethical model deployment.

\textbf{Jailbreak and adversarial.} This subcategory focuses on evaluating Audio Language Models’ (ALMs) resilience to malicious audio inputs designed to bypass safety constraints or elicit harmful responses. AudioJailbreak~\cite{song2025audio} proposes AJailBench, the first benchmark that systematically evaluates jailbreak vulnerability through adversarial audio prompts and their perturbed variants, targeting semantically consistent yet policy-violating outputs. Building on this direction, AdvBench-Audio~\cite{kang2024advwave} introduces AdvWave, a dual-phase optimization framework that generates perceptually natural adversarial perturbations using gradient-based attacks, significantly improving attack success rates against state-of-the-art ALMs. JALMBench~\cite{peng2025jalmbench} further expands the landscape by providing a large-scale, unified benchmark covering over 50,000 adversarial audio samples, supporting various attack and defense strategies, and enabling standardized cross-model comparisons. Complementing these works, Multi-AudioJail~\cite{roh2025multilingual} reveals that multilingual and accented audio inputs pose additional security risks, showing that cross-lingual phonetic variation and acoustic perturbations can dramatically increase jailbreak success rates—especially in multimodal systems. 

These benchmarks collectively offer a comprehensive landscape for evaluating ALMs’ safety boundaries in real-world and high-risk audio understanding and interactions.

Despite recent progress, future benchmark-based evaluation of ALMs will increasingly emphasize:
\begin{itemize}
  \item \textbf{Unified and Multi-Dimensional Evaluation.} Existing benchmarks are fragmented across task types. Future efforts should promote integrated, scenario-based evaluations that jointly assess reasoning, safety, and interactivity.
  
  \item \textbf{Open-World and Low-Resource Generalization.} Most current datasets are curated and high-resource. Robust evaluation requires benchmarks targeting low-resource, zero-shot, and long-tail audio scenarios.

  \item \textbf{Real-Time Personalization and Long-Context Dialogue.} Current interaction benchmarks fall short in modeling speaker diversity, personalization, and memory over long-form conversations. Future benchmarks should capture dynamic adaptation to user profiles, evolving dialogue states, and sustained speech-driven interactions.

  \item \textbf{Beyond Jailbreak: Societal and Ethical Risks.} Safety evaluation remains narrowly focused. Broader assessments should include fairness, bias, deepfake misuse, and privacy concerns under realistic deployment conditions.
  
  \item \textbf{Personalized and Profile-Aware Generation Evaluation.} Future benchmarks should evaluate whether models can adapt speech generation to user-specific profiles, capturing speaker identity, emotion, intent, and historical context to enable more natural, consistent, and user-centric interactions.

  \item \textbf{Scalable and Transparent Evaluation Frameworks.} There is a growing need for reproducible, extensible evaluation pipelines that support automation, multi-language tasks, and unified scoring across benchmarks.
\end{itemize}

\begin{table}[htbp]
\centering
\resizebox{\textwidth}{!}{
\begin{tabular}{ccccc}
\toprule
\textbf{Benchmark} & \textbf{Category} & \textbf{Modality} & \textbf{Evaluation Subtype} & \textbf{Representative Metrics} \\
\midrule
AIR-Bench~\cite{yang2024airbench} & Understanding & Speech/Audio/Music & Reasoning-based & Task Success Rate \\
MuChoMusic~\cite{weck2024muchomusic} & Understanding & Music & QA-based & QA Accuracy \\
MMAU~\cite{sakshi2024mmau} & Understanding & Speech/Audio/Music & Multi-task & Task-specific Metrics\\
AudioBench~\cite{wang2024audiobench} & Understanding & Speech/Audio & Multi-task & Task-specific Metrics \\
Dynamic-SUPERB-P2~\cite{huang2024dynamic} & Understanding & Speech/Audio/Music & Multi-task  & Task-specific Metrics \\
MMAR~\cite{ma2025mmar} & Understanding & Speech/Audio/Music & Reasoning-based & Accuracy \\
UltraEval-Audio~\cite{ultraeval-audio2025} & Understanding & Speech/Audio/Music & Multi-task & Task-specific Metrics \\
SALMon~\cite{maimon2025salmon} & Understanding & Speech & QA-based & Accuracy \\
FinAudio~\cite{cao2025finaudio} & Understanding & Speech & Multi-task & Task-specific Metrics \\
Audio Entailment~\cite{deshmukh2025audio} & Understanding & Audio & Reasoning-based & Classification metrics \\
SAKURA~\cite{yang2025sakura} & Understanding & Speech/Audio & Reasoning-based & LLM-vetted Accuracy \\
JASCO~\cite{wang2025they} & Understanding & Speech/Audio/Music & Reasoning-based & Relevance Score \\
SAGI~\cite{bu2024roadmap} & Understanding & Speech/Audio/Music & Multi-task & Task-specific Metrics \\
MAE~\cite{chen2024beyond} & Understanding & Speech/Audio & QA-based & Task-specific Metrics \\
BEANS-Zero~\cite{robinson2024naturelm} & Understanding & Audio & QA-based & Task-specific Metrics \\
RUListening~\cite{zang2025you} & Understanding & Music & QA-based & Accuracy \\
SpeechCaps~\cite{huang2025speechcaps} & Understanding & Speech & QA-based & Accuracy \\
\cite{kuan2025can} & Understanding & Audio & Reasoning-based & Accuracy \\
\cite{gpt4o} & Understanding & Speech/Audio/Music & Reasoning-based & Micro-averaged Accuracy \\
\midrule
DiscreteEval~\cite{wang2024evaluating} & Generation & Speech & Quality-based & Task-specific Metrics \\
EmergentTTS-Eval~\cite{manku2025emergenttts} & Generation & Speech & Multi-dimensional-based & LLM Judges \\
\midrule
VoxEval~\cite{cui2025voxeval} & Interaction & Speech & QA-based & QA Accuracy \\
SD-Eval~\cite{ao2024sd-eval} & Interaction & Speech/Audio & Spoken Dialogue & LLM Judges \\
VoiceBench~\cite{chen2024voicebench} & Interaction & Speech/Audio & Conversational Behavior & LLM Judges \\
Full-Duplex-Bench~\cite{Full-Duplex-Bench} & Interaction & Speech & Conversational Behavior & LLM Judges, Latency \\
Vox-Profile~\cite{feng2025vox} & Interaction & Speech & Personalization \& Profiling & Accuracy \\
URO-Bench~\cite{yan2025uro} & Interaction & Speech & Conversational Behavior & MOS, WER, Latency \\
Audiopedia~\cite{penamakuri2025audiopedia} & Interaction & Audio & QA-based & QA Accuracy \\
StyleTalk~\cite{lin2024advancing} & Interaction & Speech & Conversational Behavior & Task-specific Metrics \\
VoxDialogue~\cite{cheng2025voxdialogue} & Interaction & Speech/Audio/Music & Spoken Dialogue & Task-specific Metrics \\
IFEval-Audio~\cite{gao2025ifeval} & Interaction & Audio/Speech & Instruction Following & Instruction Following Rate \\
Talking Turns~\cite{arora2025talking} & Interaction & Speech & Spoken Dialogue & Turn-Taking Metrics \\
EvalSIFT~\cite{pandey2025sift} & Interaction & Speech & Instruction Following & LLM Judges \\
Speech-IFEval~\cite{Speech-ifeval} & Interaction & Speech & Instruction Following & LLM Judges \\
ContextDialog~\cite{kim2025does} & Interaction & Speech & Spoken Dialogue & LLM Judges \\
ADU-Bench~\cite{gao2024benchmarking} & Interaction & Audio & Spoken Dialogue & LLM Judges \\
S2S-Arena~\cite{jiang2025s2s} & Interaction & Speech & Instruction Following & LLM Judges \\
\midrule
AudioTrust~\cite{li2025audiotrust} & Safety & Speech/Audio & Multi-dimensional Safety & Group Fairness Metrics \\
AdvBench-Audio~\cite{kang2024advwave} & Safety & Audio & Jailbreak \& Adversarial & Jailbreak Success Rate \\
AudioJailbreak~\cite{song2025audio} & Safety & Audio & Jailbreak \& Adversarial & Jailbreak Success Rate \\
JALMBench~\cite{peng2025jalmbench} & Safety & Audio & Jailbreak \& Adversarial  & LLM judges \\
Multi-AudioJail~\cite{roh2025multilingual} & Safety & Audio & Jailbreak \& Adversarial & Jailbreak Success Rate \\
\bottomrule
\end{tabular}
}
\caption{Refined taxonomy of benchmark-based evaluation for speech, audio or music. The "Evaluation Subtype" column categorizes benchmarks into finer-grained types based on evaluation intent, offering clearer distinctions within Understanding, Generation, Interaction, and Safety tasks.}
\label{tab:benchmark_eval_speech}
\end{table}

\subsection{Meta-Evaluation Benchmarks for Speech Generation}
\label{sec:speech_meta_evaluation_benchmarks}
Currently, most meta-evaluation benchmarks for audio modality generation focus on automatic MOS prediction for text-to-speech (TTS) systems, primarily evaluating the correlation between automatic scoring metrics and human subjective ratings. In the audio and music domains, due to the reliance on extensive expert annotations for aesthetic evaluation, relevant meta-evaluation benchmark data remains scarce. The prevailing meta-evaluation benchmarks shown in Table~\ref{tab:speech_meta_evaluation} can be categorized into three main types: (1) \textit{challenge-oriented}, (2) \textit{dataset-oriented}, and (3) \textit{toolkit-oriented}. It is important to note that toolkit-based meta-evaluation approaches differ from toolkits directly used for speech and audio quality assessment such as VERSA~\cite{shi2024versa} and Aquatk~\cite{vinay2023aquatk}. Their primary role is to serve as auxiliary toolkits for evaluating MOS prediction models, providing convenient meta-evaluation support for researchers.

\subsubsection*{Challenge-oriented Benchmarks}

Challenge-oriented benchmarks are typically organized as community competitions designed to test the generalization and robustness of MOS prediction models under realistic, complex, and challenging acoustic environments. Representative examples include VoiceMOS~\cite{huang2022voicemos,cooper2023voicemos,huang2024voicemos}, a multi-year evaluation series covering zero-shot prediction, cross-lingual robustness, and singing voice quality prediction. This series builds a dynamic evaluation framework based on large-scale human subjective rating corpora. Recently, the AudioMOS Challenge~\cite{audiomos2025} expanded the evaluation scope by introducing innovative tasks such as text-to-music MOS prediction, high-fidelity speech quality assessment, and audio aesthetic alignment, establishing a new benchmark for general-purpose audio generation quality evaluation. In addition, ConferencingSpeech 2022~\cite{yi2022conferencingspeech} focuses on remote conferencing scenarios, evaluating speech clarity and call quality in complex communication environments with multiple speakers.

\subsubsection*{Dataset-oriented Benchmarks}

Dataset-oriented benchmarks provide standardized corpora annotated with human and expert ratings, offering a solid data foundation for training and evaluating speech quality models. Their advantage lies in covering multiple dimensions across speech, singing, and music, supporting comprehensive analysis across diverse acoustic environments and tasks. SOMOS~\cite{maniati2022somos} is the first large-scale MOS dataset composed solely of neural TTS synthesized samples. It contains approximately twenty thousand utterances generated by two hundred distinct neural acoustic models. All samples use a uniform vocoder to ensure differences arise only from acoustic models, facilitating the training of automatic MOS predictors and quality assessment of modern speech synthesizers. NISQA~\cite{mittag2021nisqa} targets real-world communication scenarios, comprising diverse speech samples with simulated distortions such as packet loss, filtering, and coding artifacts, authentic background noises, as well as live telephone and VoIP conversations. All samples are annotated with multiple subjective ratings according to ITU-T P.808 and P.800 standards, ensuring label accuracy and reliability. This dataset supports the evaluation of models predicting speech quality under complex network and multi-device conditions. SingMOS~\cite{tang2024singmos} addresses cross-cultural music expression needs by providing a singing synthesis quality dataset for Chinese and Japanese. It covers outputs from various synthesis and conversion systems, enabling multi-style analysis and filling the gap in music-related speech evaluation. MOS-Bench~\cite{huang2024mosbench} integrates nineteen MOS datasets from text-to-speech, voice conversion, and speech enhancement domains, and combines with the SHEET toolkit to build a unified scoring process and interpretable evaluation ecosystem. Tencent~\cite{yi2022conferencingspeech} focuses on meeting speech quality, offering large-scale real meeting audio samples and detailed quality annotations, with particular attention to network jitter and packet loss effects on speech quality. PSTN~\cite{mittag2021deep} collects speech samples from traditional public switched telephone networks, covering various encoding formats and transmission impairments, serving as a benchmark for telephone speech quality evaluation. IUMOS~\cite{dong2020pyramid} targets speech quality assessment in noisy environments, including recordings from factories, traffic, and public places, supporting the development of robust speech evaluation models. Furthermore, MusicEval~\cite{liu2025musiceval}, as the first meta-evaluation benchmark for generative music, provides a reliable validation platform for text-to-music generation systems through expert annotation and rigorous experimental design.

\subsubsection*{Toolkit-oriented Benchmarks}

Toolkit-oriented benchmarks provide standardized infrastructure at the system implementation level, aiming to support automated evaluation, result reproducibility, and interpretable analysis for speech generation systems. SHEET~\cite{huang2024mosbench} offers a comprehensive evaluation pipeline solution that not only supports standard MOS model assessment but also innovatively integrates score gap analysis, latent space visualization, and generalization diagnostics, covering evaluation from fundamental metrics to in-depth analysis.

The three categories of benchmarks form a closely connected framework. Dataset-oriented benchmarks supply essential data for challenge-oriented evaluations, while toolkit-oriented benchmarks provide fair and efficient pipeline support for the implementation of both. Meta-evaluation benchmarks will become increasingly important in the future, not only because automatic evaluation metrics require closer alignment with human preferences and intentions, but also because assessing the value of automatic evaluation models plays a critical role in advancing the quality and development of speech generation systems. Future trends will inevitably move towards interpretable, multi-dimensional meta-evaluation frameworks that are closely aligned with human judgment.

\begin{table}[htbp]
\centering
\scalebox{0.9}{
\begin{tabular}{cccc}
\toprule
\textbf{Benchmark} & \textbf{Type} & \textbf{Focus} & \textbf{Data} \\
\midrule
VoiceMOS~\cite{huang2022voicemos,cooper2023voicemos,huang2024voicemos} & Challenge & MOS Prediction & Speech/Singing \\
SingMOS~\cite{tang2024singmos} & Challenge & MOS Prediction & Singing \\
AudioMOS~\cite{audiomos2025} & Challenge & MOS \& Aesthetics Prediction & Audio \\
ConferencingSpeech~\cite{yi2022conferencingspeech} & Challenge & MOS Prediction & Speech \\
QualiSpeech~\cite{wang2025qualispeech} & Corpus & Description \& Explanation & Speech \\
SongEval~\cite{yao2025songeval} & Corpus & Aesthetics Prediction & Singing\\
SOMOS~\cite{maniati2022somos} & Corpus & MOS Prediction & Speech \\
NISQA~\cite{mittag2021nisqa} & Corpus & Speech Quality & Speech \\
MOS-Bench~\cite{huang2024mosbench} & Corpus & MOS Prediction & Speech \\
Tencent~\cite{yi2022conferencingspeech} & Corpus & Speech Quality & Speech \\
PSTN~\cite{mittag2020dnn} & Corpus & Speech Quality & Speech \\
IUMOS~\cite{dong2020pyramid} & Corpus & Speech Quality & Speech \\
MusicEval~\cite{liu2025musiceval} & Corpus & Music Quality & Music \\
SHEET~\cite{huang2024mosbench} & Toolkit & MOS Prediction & Speech \\
\bottomrule
\end{tabular}
}
\caption{A summary of representative meta-evaluation benchmarks for automatic speech assessment. Benchmarks are categorized by their type, either challenge oriented community competitions or dataset oriented curated corpora; their focus, such as mean opinion score MOS prediction, perceptual quality modeling, or natural language explanation; and their supported data modality, including speech, singing voice, and general audio. These benchmarks provide standardized resources for evaluating the performance and generalization ability of automatic speech quality assessment models across domains, languages, and acoustic conditions.}
\label{tab:speech_meta_evaluation}
\end{table}

\subsection{Comparing Automatic MOS Evaluations for Speech Generation\label{sec:comparision_mos_eval}}

Automatic mean opinion score (MOS) prediction plays a crucial role in evaluating the perceptual quality of synthesized speech, serving as a scalable and cost-effective alternative to traditional human evaluations in modern text-to-speech (TTS) systems. These methods train models on annotated datasets to simulate subjective human ratings, enabling large-scale and consistent quality assessment. We compare two core approaches reflecting the evolution of this field: learning-based methods and ALM-based methods. Learning-based approaches learn MOS prediction through fully supervised or self-supervised techniques, whereas ALM-based approaches leverage large-scale pretrained multimodal models fine-tuned on task-specific MOS data. To systematically assess these models, we conduct experiments on multiple benchmark datasets covering both general and domain-specific scenarios. To ensure consistency and comparability, we adopt a unified utterance-level evaluation protocol. General benchmarks include test sets from NISQA~\cite{mittag2021nisqa}, VoiceMOS~\cite{huang2022voicemos}, and SOMOS~\cite{maniati2022somos}, aligned with the settings of MOS-Bench~\cite{huang2024mosbench}. For domain generalization evaluation, we select the entire test set from SingMOS~\cite{tang2024singmos} and the development set from Tencent~\cite{yi2022conferencingspeech}. We employ three standard metrics to measure model performance: linear correlation coefficient (LCC) and Spearman’s rank correlation coefficient (SRCC) to evaluate the linear and monotonic correlations between predicted and human ratings, and mean squared error (MSE) to quantify absolute prediction error. Higher LCC and SRCC values and lower MSE indicate stronger alignment between model predictions and human perception. The evaluated models include supervised learning-based methods like MOSNet \cite{lo2019mosnet} and LDNet \cite{huang2022ldnet}, self-supervised models such as UTMOSv2 \cite{baba2024t05}, SCOREQ \cite{ragano2024scoreq}, RAMP \cite{wang2023ramp}, and Modified SSL-MOS \cite{huang2024mosbench}, as well as ALM-based methods like SALMONN-Lora and Qwen2-Audio-Lora \cite{lora}, which utilize pretrained audio-language backbones with task-specific Lora fine-tuning.

Our comparative analysis in Table \ref{tab:general-purpose-speech-eval} and Table \ref{tab:generalized-speech-eval} evaluates Learning-based and ALM-based approaches for automatic mean opinion score (MOS) prediction across multiple benchmarks. On general purpose datasets such as NISQA, VoiceMOS-BVCC, and SOMOS, ALM-based models particularly SALMONN-Lora consistently outperform Learning-based methods. For example, SALMONN-Lora achieves an LCC of 0.861, SRCC of 0.859, and MSE of 0.347 on NISQA, indicating strong alignment with human perception and low prediction error. Similarly, on SOMOS, it attains an LCC of 0.644 and MSE of 0.196, outperforming many baselines. In contrast, Learning-based models like UTMOSv2 demonstrate competitive results on clean or in domain data, such as the VoiceMOS-BVCC test, where it achieves an LCC of 0.945 and SRCC of 0.949. However, their performance deteriorates significantly under domain shifts. On domain specific benchmarks like SingMOS (singing voice synthesis) and TencentDev (real world conferencing speech), many Learning-based models show weak or even negative correlations with human ratings. For instance, on SingMOS, SALMONN-Lora’s correlation drops to an LCC of 0.372, SRCC of 0.347 and MSE increases to 1.928, whereas Learning-based RAMP achieves a better LCC of 0.505. Meanwhile, SALMONN-Lora demonstrates superior robustness on Tencent-Dev, reaching an LCC of 0.758 and MSE of 0.394, substantially outperforming Learning-based alternatives.

These findings suggest that ALM-based methods benefit from large scale pretraining and contextual representation learning, which enhance generalization and robustness across diverse acoustic and semantic conditions. Meanwhile, Learning-based models can excel in specialized domains when carefully adapted. Overall, integrating multimodal pretraining with domain aware fine tuning appears essential for developing reliable and generalizable MOS prediction systems, positioning ALMs as promising foundations for future perceptual speech evaluation. ALM-based approaches are still at an early stage and hold significant potential for further development. Future trends may include expanding to more diverse and complex meta evaluation datasets while maintaining stability, optimizing ALMs specifically for speech evaluation through techniques such as knowledge distillation from ALLD~\cite{alld}, improved prompt engineering, fine tuning strategies, and incorporating explainability-driven training data. These advancements are likely to further enhance the accuracy, robustness, and interpretability of ALM-based MOS prediction models.

\begin{table*}[htbp]
    \centering
    \resizebox{\textwidth}{!}{
    \small
    \begin{tabular}{ll|
        ccc|ccc|ccc}
        \toprule
        \multirow{2}{*}{\textbf{Category}} & \multirow{2}{*}{\textbf{Model}} &
        \multicolumn{3}{c|}{\textbf{NISQA\textsubscript{AVG}}} &
        \multicolumn{3}{c|}{\textbf{VoiceMOS-BVCC\textsubscript{Test}}} &
        \multicolumn{3}{c}{\textbf{SOMOS\textsubscript{Test}}} \\
        & &
        \textbf{LCC$\uparrow$} & \textbf{SRCC$\uparrow$} & \textbf{MSE$\downarrow$} &
        \textbf{LCC$\uparrow$} & \textbf{SRCC$\uparrow$} & \textbf{MSE$\downarrow$} &
        \textbf{LCC$\uparrow$} & \textbf{SRCC$\uparrow$} & \textbf{MSE$\downarrow$} \\
        \midrule
        Learning-based & MOSNET~\cite{lo2019mosnet} & 0.413 & 0.394 & 0.975 & 0.501 & 0.530 & 0.993 & - & - & - \\
        Learning-based & LDNET~\cite{huang2022ldnet} & 0.290 & 0.345 & 0.936 & 0.606 & 0.596 & 0.880 & - & - & - \\
        Learning-based & UTMOSv2~\cite{baba2024t05} & 0.629 & 0.614 & 0.696 & \textbf{0.945} & \textbf{0.949} & 0.439 & 0.438 & 0.413 & 0.306 \\
        Learning-based & SCOREQ~\cite{ragano2024scoreq} & 0.711 & 0.694 & 0.732 & 0.893 & 0.892 & 0.240 & 0.449 & 0.436 & 0.849 \\
        Learning-based & RAMP~\cite{wang2023ramp} & - & - & - & 0.904 & 0.903 & 0.177 & 0.395 & 0.387 & 0.958 \\
        Learning-based & Modified SSL-MOS~\cite{huang2024mosbench} & 0.613 & 0.614 &  0.860 & 0.895 & 0.891 & 0.253 & 0.492 & 0.482 & 0.662 \\
        \midrule
        ALM-based & SALMONN(vic1.5)-Lora~\cite{lora} & \textbf{0.861} & \textbf{0.859} & \textbf{0.347} & 0.826 & 0.833 & \textbf{0.282} & \textbf{0.644} & \textbf{0.636} & \textbf{0.196} \\
        ALM-based & Qwen2-Audio-Lora~\cite{lora} & 0.768 & 0.780 & 0.643 & 0.681 & 0.678 & 0.493 & 0.583 & 0.572 & 0.216 \\
        \bottomrule
    \end{tabular}
    }
\caption{\label{tab:mos_eval} Comparison of automatic MOS prediction methods across \textbf{NISQA}, \textbf{VoiceMOS-BVCC}, and \textbf{SOMOS}. These benchmarks focus on speech quality evaluation.}
\label{tab:general-purpose-speech-eval}
\end{table*}

\begin{table*}[htbp]
    \centering
    \resizebox{0.85\textwidth}{!}{
    \small
    \begin{tabular}{ll|
        ccc|ccc}
        \toprule
        \multirow{2}{*}{\textbf{Category}} & \multirow{2}{*}{\textbf{Model}} &
        \multicolumn{3}{c|}{\textbf{SingMOS\textsubscript{Test}}} &
        \multicolumn{3}{c}{\textbf{Tencent\textsubscript{Dev}}} \\
        & &
        \textbf{LCC$\uparrow$} & \textbf{SRCC$\uparrow$} & \textbf{MSE$\downarrow$} &
        \textbf{LCC$\uparrow$} & \textbf{SRCC$\uparrow$} & \textbf{MSE$\downarrow$} \\
        \midrule
        Learning-based & MOSNET~\cite{lo2019mosnet} & 0.327 & 0.158 & 1.379 & 0.375 & 0.417 & 0.972 \\
        Learning-based & LDNET~\cite{huang2022ldnet} & 0.424 & 0.200 & \textbf{1.158} & -0.285 & -0.288 & 1.166 \\
        Learning-based & UTMOSv2~\cite{baba2024t05} & 0.506 & 0.452 & 4.449 & 0.435 & 0.395 & 1.575 \\
        Learning-based & SCOREQ~\cite{ragano2024scoreq} & 0.493 & \textbf{0.480} & 3.348 & 0.442 & 0.311 & 2.111 \\
        Learning-based & RAMP~\cite{wang2023ramp} & \textbf{0.505} & \textbf{0.480} & 3.489 & 0.325 & 0.240 & 1.669 \\
        Learning-based & Modified SSL-MOS~\cite{huang2024mosbench} & 0.429 & 0.415 & 5.740 & 0.338 & 0.297 & 2.553 \\
        \midrule
        ALM-based & SALMONN(vic1.5)-Lora~\cite{lora} & 0.372 & 0.347 & 1.928 & \textbf{0.758} & \textbf{0.767} & \textbf{0.394 }\\
        ALM-based & Qwen2-Audio-Lora~\cite{lora} & - & - & - & - & - & - \\
        \bottomrule
    \end{tabular}
    }
\caption{\label{tab:generalization_eval} Evaluation of model generalization on \textbf{SingMOS} and \textbf{Tencent} datasets, which reflect more diverse and open-domain speech scenarios.}
\label{tab:generalized-speech-eval}
\end{table*}

\subsection{Challenges and Future Trends} 
\label{sec:speech_conclusion_challenge_future}

Despite significant progress, current automatic evaluation methods for speech synthesis still face challenges such as limited generalization, insufficient interpretability, and weak robustness under diverse conditions. Moreover, the scarcity of high-quality and diverse annotated data, coupled with the high cost of obtaining annotations from listeners and experts, especially constrains the construction of high-quality meta-evaluation benchmarks in emerging scenarios such as generalized audio, emotional speech, accented speech, and speech disorders. These limitations further restrict the applicability and generalizability of automatic evaluation models. Additionally, sensitivity to prompt design and inadequate calibration of prediction confidence hinder their practical utility and stability.

With the advent of large-scale audio-language models (ALMs) tailored for speech understanding, future developments in automatic evaluation will focus on improving universality, interpretability, and multidimensional coverage. ALM-based methods are capable not only of producing scalar quality scores but also of generating diagnostic feedback in natural language, significantly enhancing the transparency and explainability of evaluation results. ALMs demonstrate strong potential in cross-lingual and cross-domain generalization, laying a solid foundation for building robust and universal speech quality assessment systems.

This survey has discussed future trends in each section related to speech and audio generation evaluation methods. Here, we particularly emphasize that the multimodal perception capabilities of ALMs open new avenues for cross-modal consistency evaluation and facilitate integration between audio evaluation and other modalities. For instance, in talking-face synthesis, ALMs can assess the emotional alignment between speech and facial expressions; in music generation, they can evaluate the coherence among lyrics, melody, and visual rhythm; in immersive 3D audio scenarios, evaluation must encompass source localization and spatial auditory consistency. These multimodal challenges highlight the necessity of developing comprehensive evaluation frameworks equipped with multimodal understanding and reasoning capabilities. Consequently, optimizing ALMs for high-quality fine-tuning and learning on currently limited meta-evaluation datasets emerges as a critical trend.

Looking ahead, automatic evaluation of audio generation is expected to transcend traditional distortion-based metrics and evolve toward unified, multidimensional, and generalizable frameworks that integrate semantic understanding, emotional expression, cross-modal alignment, and explainable reasoning. Such evaluation systems will not only drive innovation in generative model technologies but also provide strong support for intelligent multimodal interaction and content creation.
\section{Conclusions and Future Work}

This paper presents a systematic review of automatic evaluation methods for generative models across text, visual, and audio modalities, introducing a unified taxonomy that classifies existing approaches into five categories: heuristic-based, embedding-based, learning-based and LLM/VLM/ALM-based evaluation. Through detailed comparative analysis, we elucidate the strengths and limitations of each paradigm, providing a foundation for advancing evaluation methodologies.
Future work will extend this framework to other modalities and tasks while addressing critical challenges such as evaluation bias, generalization across domains, and scalability in increasingly complex generative systems.


\appendix

\end{document}